%% file: RAE-PPG_arxiv.tex
\documentclass{article}

\usepackage[T1]{fontenc}
\usepackage{iclr2027_conference,times}
\usepackage{amsmath,amssymb}
\usepackage{booktabs}
\usepackage{tabularx,array}
\usepackage{float}
\usepackage{wrapfig}
\usepackage{placeins}
\usepackage{graphicx}
\usepackage{hyperref}
\usepackage{url}
\usepackage[normalem]{ulem}

\newcommand{\runinheading}[1]{\textbf{#1}}
\newcommand{\runinbreak}{\par\noindent}
\newcommand{\topcaption}[1]{\caption{#1}}
\AddToHook{env/table/begin}{\setlength{\belowcaptionskip}{\baselineskip}}
\AddToHook{env/table*/begin}{\setlength{\belowcaptionskip}{\baselineskip}}
\makeatletter
\let\rae@native@makecaption\@makecaption
\def\rae@tabletype{table}
\long\def\@makecaption#1#2{%
  \ifx\@captype\rae@tabletype
    \vskip\abovecaptionskip
    \noindent #1: #2\par
    \vskip\belowcaptionskip
  \else
    \rae@native@makecaption{#1}{#2}%
  \fi}
\makeatother
\newcommand{\bestscore}[1]{{\fontencoding{T1}\fontfamily{ptm}\fontseries{b}\selectfont #1}}
\newcommand{\secondbest}[1]{{\renewcommand{\ULthickness}{0.75pt}\setlength{\ULdepth}{1.35pt}\uline{#1}}}

\title{RAE-PPG: Duration-Grounded\\Retain-and-Extend Pretraining\\for PPG Foundation Models}
\hypersetup{pdftitle={RAE-PPG: Duration-Grounded Retain-and-Extend Pretraining for PPG Foundation Models},
  pdfauthor={Suyeong Lee, Hochang Lee, Seokyong Sheem, Daekyum Kim}}
\author{Suyeong Lee\thanks{Equal contribution.}\quad Hochang Lee\textsuperscript{\ensuremath{\ast}}\quad Seokyong Sheem\quad Daekyum Kim\thanks{Corresponding author.}\\
\normalfont Korea University\\
\normalfont\texttt{\{tsw1615, cvcvcv99, sheemsy, daekyum\}@korea.ac.kr}}
\iclrfinalcopy

\begin{document}
\defcitealias{taskforce1996hrv}{Task Force, 1996}

\maketitle
% A preprint has no conference-status running header or header rule.
\fancyhead{}
\renewcommand{\headrulewidth}{0pt}

\begin{abstract}
\input{main4/sections/abstract_main4}
\end{abstract}

\input{main4/sections/intro_main4}
\input{main4/sections/related_main4}
\input{main4/sections/method_main4}
\input{main4/sections/experiments_main4}
% Continue Discussion after the Results text and floats without a forced blank page.
\input{main4/sections/discussion_main4}
% Continue the preprint statements after Discussion without a forced mostly blank page.
\FloatBarrier

% Reproducibility statement (2026-09-20), placed before the AI use disclosure as in the ICLR template.
\section*{Reproducibility Statement}
The accompanying RAEPPG code package provides the core implementation of the proposed method:
pretrained weights, signal preprocessing, signal-feature target extraction, the encoder, staged pretraining,
embedding extraction and frozen linear-probe evaluation. It includes numerical configurations,
training-split target statistics, dependency versions, synthetic verification tests and a README with
setup and execution instructions. Appendices~\ref{app:pretraining-corpus}--\ref{app:implementation}
specify the pretraining data, model, targets and optimization; Appendix~\ref{app:cross-duration-evaluation}
documents downstream cohorts and evaluation, and Appendices~\ref{app:physiological-evidence},
\ref{app:mechanistic-protocol} and~\ref{app:matched-retrain} describe the additional analyses.
Source recordings remain available from their respective providers, and no participant-level data
are redistributed.

% PI-supplied AI use disclosure, 2026-09-17.
\section*{AI Use Statement}
Generative AI tools were used for literature discovery, language editing and polishing, feedback on author-developed research methodology, and assistance with software implementation and code organization. The core research questions, methodological design, experimental decisions, and scientific claims were developed and determined by the authors.

\section*{Acknowledgments}
This work was supported by the Korean ARPA-H Project through the Korea Health Industry Development Institute (KHIDI), funded by the Ministry of Health \& Welfare, Republic of Korea (No. RS-2025-25455839).
This research was conducted with support from the AI Data Center service provided by the Artificial Intelligence Industry Cluster Agency (AICA), Republic of Korea.

% References follow Acknowledgments; only the appendix starts a new page.
% The template explicitly identifies References as a possible font-size exception.
\begingroup
\small
\bibliography{references_main4}
\bibliographystyle{iclr2027_conference}
\endgroup

\clearpage
\appendix
\section*{Appendix}
% Use the same top-caption alignment in the Appendix.
% Appendix tables are placed where they are declared ([H]) unless listed as floats,
% so the appendix does not leave large blank areas. Main-text placement is unchanged.
\input{main4/sections/appendix_method_main4_format}
\input{main4/sections/appendix_experiments_main4}

\input{main4/sections/appendix_matched_retrain_main4}

\end{document}

%% file: main4/sections/abstract_main4.tex
Signal features derived from photoplethysmography (PPG) require different signal durations to characterize. Existing PPG foundation models treat duration as a pretraining or evaluation condition rather than using the different durations required by PPG features to organize self-supervision. We hypothesize that self-supervision should expand with signal duration, allowing a single encoder to progressively acquire additional features while preserving and reusing earlier learning. We introduce Retain-and-Extend PPG (RAE-PPG), which trains a single Transformer encoder successively on 10~s, 30~s, and 240~s inputs, adding supervision for signal features supported by each longer observation. The encoder is partitioned into duration-specific parameter groups, allowing later stages to reuse earlier groups while updating only the group assigned to the current stage. Selected earlier targets are reused to supervise later stages, encouraging the corresponding features to remain accessible in longer-input representations. Direct decoding from the final encoder shows that earlier features remain recoverable from longer-input representations, while later-stage features show higher mean decoding performance at their introduction durations. Controlled comparisons further show that prior-stage learning provides a better basis for learning newly introduced features at both transitions. Across 18 tasks from eight datasets, the final frozen encoder achieves the best observed score on 12 tasks compared with five existing PPG foundation models.

%% file: main4/sections/intro_main4.tex
% PI-supplied Introduction, 2026-09-24; citation markup, final punctuation, and PI-requested first-mention full name adapted.
\section{Introduction}
\label{sec:introduction}

Photoplethysmography (PPG) is a non-invasive and widely accessible method for monitoring cardiovascular, respiratory, and autonomic physiology~\citep{allen2007photoplethysmography,sheem2026must}. PPG signals contain physiology-relevant features spanning local pulse morphology, fiducial timing, and temporal variability. Local morphology and fiducial timing can be characterized from individual pulses or a few cardiac cycles~\citep{elgendi2012fingertip}, whereas beat-interval variability and slower waveform dynamics require observations over tens of seconds to minutes~\citep{mejiamejia2023duration,kiselev2024ppgv}. Signal duration therefore determines which physiology-relevant features can be reliably characterized from PPG.

Existing PPG foundation models have been pretrained on fixed-length inputs ranging from seconds to minutes, using morphology-aware, contrastive, and generative self-supervised objectives~\citep{pillai2025papagei,saha2025pulseppg,guo2026sigmappg}. Recent work has also incorporated physiologically grounded supervision into PPG representation learning using PPG morphology indices and auxiliary physiological metrics, including those derived from ECG~\citep{pillai2025papagei,geenjaar2026robustppg}. However, such supervision is still typically defined within a fixed observation duration, such as 10~s~\citep{pillai2025papagei,geenjaar2026robustppg}. This limits the range of physiology-relevant features that can guide representation learning: short inputs can support local pulse features but not features that require longer observations~\citep{elgendi2012fingertip,mejiamejia2023duration}. Such longer-timescale features may capture task-relevant information for downstream tasks requiring longer signal durations. These observations motivate learning from physiology-relevant features at the signal durations needed to characterize them. The key question is how to use such duration-appropriate supervision to learn representations that remain informative across varying input lengths.

\input{main4/figures/rae_ppg_pipeline_main4}

We address this question with Retain-and-Extend PPG (RAE-PPG), a single Transformer encoder trained successively on 10~s, 30~s, and 240~s inputs (Figure~\ref{fig:rae-pipeline}). Our focus is the successive setting in which learning acquired at one signal duration becomes prior learning for the next as additional feature supervision is introduced. RAE-PPG follows a retain-and-extend strategy: as signal duration increases, it retains earlier learning for reuse while extending supervision with additional PPG-derived targets supported by the longer observation. Longer PPG recordings contain the local pulse structure present in shorter recordings while supporting features that require longer observation. However, successive training creates a risk that learning newly available information alters what the encoder has already learned from shorter inputs. To preserve earlier learning, the encoder is partitioned into duration-specific parameter groups, allowing later stages to use groups learned at shorter durations while updating only the group assigned to the current stage. Furthermore, selected earlier targets continue to supervise later stages so that features learned at shorter signal durations remain accessible in longer-input representations.

Across 18 downstream tasks from eight datasets spanning 10~s, 30~s, and 240~s inputs, the final frozen RAE-PPG encoder achieves the best observed score on 12 tasks compared with five existing PPG foundation models. Beyond downstream transfer, direct feature decoding verifies the intended retain-and-extend behavior: features learned at shorter durations remain recoverable from longer-input representations, while features introduced at later stages become more decodable on average at their introduction durations. Ablations further clarify how the proposed components contribute to this behavior. Protected parameter organization preserves shorter-input computations, while retained-target supervision improves access to earlier features from longer inputs. Controlled comparisons between reusing prior-stage parameters and resetting them to their initial values show that prior-stage learning improves the acquisition of newly introduced features. Together, these findings support a physiology-guided framework for successive duration extension, in which a single encoder preserves and reuses earlier learning while incorporating additional feature supervision as longer observations become available.

%% file: main4/figures/rae_ppg_pipeline_main4.tex
% PI-supplied vector master, final_20260924; embedded fonts, no raster downsampling.
% Declared on page 1 for placement at the top of page 2; aspect ratio preserved.
\begin{figure}[!t]
  \centering
  \includegraphics[width=\textwidth]{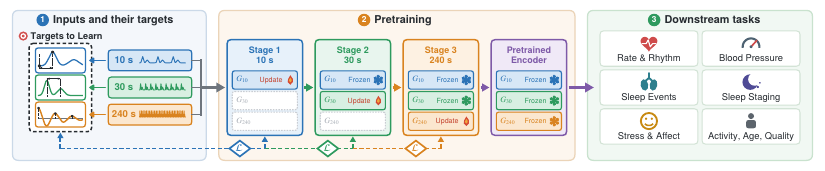}
  \caption{RAE-PPG overview: duration-grounded targets, retain-and-extend pretraining, and frozen-encoder downstream evaluation.}
  \label{fig:rae-pipeline}
\end{figure}

%% file: main4/sections/related_main4.tex
% main2-only Related Work: PI-supplied text, 2026-09-11.
% 2026-09-17: feature-duration sentences revised at the user's instruction; see main4_rebuild_20260917/literature_audit/LITERATURE_AUDIT_REPORT_KO.md sections 1-C and 3.
% Run-in headings use the official ICLR paragraph style.

\section{Related Work}
\label{sec:related}

\noindent\runinheading{Self-supervision in PPG foundation models.} PPG foundation models have used participant-aware contrastive learning, waveform-derived priors, signal-quality objectives, cross-modal alignment with ECG, and reconstruction or prediction of signal content~\citep{abbaspourazad2024wearable,pillai2025papagei,ding2024siamquality,nie2025anyppg,chen2025gptppg,guo2026sigmappg,thukral2026wavelet,lee2026himae}.
A close precedent, PaPaGei-S, uses established PPG morphology indices from 10~s signals for contrastive learning and auxiliary prediction~\citep{pillai2025papagei}, whereas \citet{geenjaar2026robustppg} use ECG- and respiratory-derived metrics to guide contrastive learning over 10~s PPG windows.
These studies demonstrate that physiologically grounded supervision can support PPG representation learning without downstream labels. However, both approaches operate on fixed 10~s PPG windows rather than progressively organizing signal-feature supervision across signal durations.

\noindent\runinheading{Signal duration in PPG representation learning.} The signal duration needed to characterize a physiology-relevant PPG feature depends on the feature.
Local waveform morphology and fiducial timing are defined on individual pulses or consecutive pulse pairs~\citep{elgendi2012fingertip}, whereas variability and spectral indices are typically evaluated over longer segments: prior studies have reported analysis durations of roughly 30--60~s for RMSSD, 60--90~s for pNN50, and 1--2~min for low-frequency power of beat-interval series~(\citetalias{taskforce1996hrv}; \citealp{baek2015ultrashort,mejiamejia2023duration}).
\citet{baek2015ultrashort} compared variability indices from ultra-short segments (10--270~s) with those from 5~min recordings, and \citet{mejiamejia2023duration} evaluated PRV indices computed from simulated PPG signals of different durations.
These studies indicate that suitable analysis duration differs across variability indices and study settings.

PPG foundation-model studies also report duration-dependent downstream performance.
Pulse-PPG shows non-monotonic performance across models pretrained with 1~min, 2~min, and 4~min PPG signals~\citep{saha2025pulseppg}, while SIGMA-PPG reports task-dependent performance with 30~s, 60~s, 120~s, and 240~s inputs~\citep{guo2026sigmappg}.
However, these approaches vary signal duration as a pretraining or evaluation condition rather than progressively organizing signal-feature supervision across signal durations.
HiMAE studies a related but different problem by learning representations at multiple internal temporal resolutions while keeping the physical signal duration fixed~\citep{lee2026himae}.
In RAE-PPG, increasing signal duration instead organizes the staged introduction of additional signal-feature supervision within a single encoder, adding physiology-relevant signal features as longer inputs support their prescribed analysis intervals.

\noindent\runinheading{Progressive and continual learning.} Sequential training can alter previously learned representations, motivating methods that protect earlier parameters or reuse earlier features~\citep{ramasesh2021anatomy,elsayed2024addressing}.
Progressive Neural Networks freeze earlier task-specific columns and reuse their features through lateral connections when new tasks are learned~\citep{rusu2016progressive}. Unlike these task-sequential settings, RAE-PPG applies preservation and reuse across increasing signal durations, introducing additional signal features as longer inputs support their prescribed analysis intervals.

%% file: main4/sections/method_main4.tex
% Method prose inherited from main3; RAE-PPG denotes RAE-PPG2-R1 (2026-09-17).
% 2026-09-17: feature-duration sentences revised at the user's instruction; see main4_rebuild_20260917/literature_audit/LITERATURE_AUDIT_REPORT_KO.md sections 1-C and 3.
% 2026-09-17: 12-of-14 sentence bound to the final-extractor re-run research/results/raeppg2_r1_q1_span_repeatability_20260917_v1 (user instruction).
% Scope: MAIN2_METHOD_REPLACEMENT_AND_APPENDIX_20260912.md.

\section{Method}
\label{sec:method}

RAE-PPG addresses two coupled requirements: introducing supervision as longer inputs support additional signal features, and preserving earlier parameters for reuse during subsequent learning.
For \(\mathcal{D}=\{10,30,240\}\)~s, a single encoder maps a preprocessed, root-wise z-scored single-channel PPG input \(x^{(r)}\in\mathbb{R}^{rf_s}\), sampled at \(f_s=64\)~Hz, to
\begin{equation}
H^{(r)}=E_{\theta}\!\left(x^{(r)}\right)\in\mathbb{R}^{K\times d},
\qquad K=64,\quad d=384,\quad r\in\mathcal{D}.
\label{eq:encoder}
\end{equation}
A local feature extractor maps the PPG waveform to a feature vector at each sample.
We pool these local features within \(K\) consecutive intervals, determined by a predefined rule based on signal quality, to form tokens (\S~\ref{app:local-feature-extractor}).
A Transformer processes the tokens, and its full output sequence \(H^{(r)}\) is used as the learned representation for downstream tasks. The resulting encoder has 10.76M parameters (\S~\ref{app:encoder}).

\subsection{Duration-grounded signal-feature supervision}
\label{sec:method-supervision}

To specify what the encoder learns, we use the signal features that classical PPG analysis derives before physiological interpretation~\citep{allen2007photoplethysmography,elgendi2012fingertip,mejiamejia2022signalprocessing}.
Computed directly from unlabeled PPG over prescribed analysis intervals, these measurements provide duration-grounded targets without external physiological annotations.
For a physical interval \(W\), the two pathways are
\begin{equation}
\begin{aligned}
\text{Classical PPG:}\quad
&x_{\mathrm{rec}}
\xrightarrow{\phi_f(\cdot;W)}
y_f(W)
\xrightarrow{\mathrm{interpretation}}
I_{\mathrm{phys}},\\
\text{RAE-PPG pretraining:}\quad
&x^{(r)}
\xrightarrow{E_\theta}
H^{(r)}
\xrightarrow{\operatorname{Read}_W,\,g_f}
\hat y_f^{(r)}(W),
\qquad
\hat y_f^{(r)}(W)\approx \tilde y_f(W).
\end{aligned}
\label{eq:classical-rae-bridge}
\end{equation}
Here, \(x_{\mathrm{rec}}\) is the continuous recording and \(\phi_f\) performs fixed, target-specific preprocessing and extraction (\S~\ref{app:targets}); \(\operatorname{Read}_W\) is an interval-specific token readout, \(g_f\) is a feature-prediction head, a single affine map without a hidden layer or output activation, and \(I_{\mathrm{phys}}\) denotes physiological interpretation.
Targets summarize \(W\); filtering and beat detection use surrounding recording context.
For each signal feature \(f\), we normalize the target \(y_f\) by subtracting its median and dividing by a scale derived from its interquartile range (IQR), yielding \(\tilde y_f\).
The median and IQR are computed using only the training data.

\noindent\runinheading{Target set and duration grounding.} Signal features require different analysis intervals.
We therefore introduce duration-defining targets only when inputs contain their prescribed intervals.
Table~\ref{tab:feature-families} summarizes sixteen PPG-derived pretraining targets across four physiology-relevant feature families and separate signal-quality supervision.
Morphology and vascular-timing targets characterize waveforms rather than directly measure vascular properties~\citep{hickey2016morphology}. Beat-interval variability targets are PPG-derived PRV measures and are not assumed to be interchangeable with ECG-derived HRV~\citep{mejiamejia2020prv}.

% Keep the expanded explanation on nine main-text pages without changing type or page geometry.
\begingroup
\begin{table}[!ht]
\topcaption{Physiology-relevant signal-feature families used by RAE-PPG.}
\label{tab:feature-families}
\centering
\scriptsize
\renewcommand{\arraystretch}{0.95}
\setlength{\tabcolsep}{2pt}
\begin{tabularx}{\textwidth}{@{}
>{\raggedright\arraybackslash}p{0.24\textwidth}
>{\raggedright\arraybackslash}p{0.34\textwidth}
>{\raggedright\arraybackslash}X@{}}
\toprule
Feature family & Signal-processing role & Realization in RAE-PPG \\
\midrule
Rate
& Pulse cadence
& Local pulse rate at 10~s; full-window mean rate at
240~s\textsuperscript{\ensuremath{\dagger}} \\
\specialrule{0.2pt}{1pt}{1pt}
Morphology / vascular timing
& Within-beat contour, fiducial timing, and reflected-wave structure
& Local morphology at 10~s; reflected-wave timing at 30~s \\
\specialrule{0.2pt}{1pt}{1pt}
Beat-interval variability
& Successive interval changes and dispersion
& PRV-RMSSD at 30~s; PRV-pNN50 and interval dispersion at 240~s \\
\specialrule{0.2pt}{1pt}{1pt}
Waveform statistics
& Baseline-sensitive autocorrelation and low-frequency modulation
& Uncentered ACF index and low-frequency waveform variability at 240~s \\
\specialrule{0.2pt}{1pt}{1pt}
Signal quality\textsuperscript{\ensuremath{\dagger}}
& Reliability of periodic pulse structure
& Spectral purity at 10~s \\
\bottomrule
\end{tabularx}
\vspace{2pt}
\begin{minipage}{0.98\textwidth}
\scriptsize
\textsuperscript{\ensuremath{\dagger}}Auxiliary targets provide cadence or
quality supervision; their placement does not define when the corresponding
quantity first becomes measurable.
\end{minipage}
\end{table}
\endgroup

We translate these duration requirements into a progression of signal-feature supervision for a single encoder. The 10~s stage provides local pulse and quality supervision, 30~s accommodates the prescribed 20~s normalized reflected-delay and 30~s PRV-RMSSD targets, and 240~s covers the remaining prescribed intervals up to 240~s (Table~\ref{tab:app-targets}).
For a duration-defining target with prescribed interval length \(\ell_f\), the introduction stage is \(r_f=\min\{r\in\mathcal D:\ell_f\le r\}\).
At the stage using 240~s inputs, we introduce targets computed over 60~s or 120~s intervals within each input.
The stages introduce nine, two, and five targets, respectively, including the auxiliary targets distinguished in Table~\ref{tab:feature-families}. These placements reflect the selected target definitions rather than universal physiological thresholds. We use three stages because two successive extensions are the minimum needed to test whether learning acquired at one extension can serve as prior learning for the next; 60~s and 120~s targets are therefore grouped into the 240~s stage rather than defining additional stages.
\S~\ref{app:targets} details target-set selection and construction, availability and numerical checks, and signal-level characterization.

\noindent\runinheading{Interval-aligned prediction.} To align each prediction with its target, the analysis interval \(W_{f,j}=[a_{f,j},b_{f,j})\) for feature \(f\) and occurrence \(j\) also serves as the query interval.
For pulse-level targets, it spans an individual pulse or pulse pair.
With token \(h_k^{(r)}\) assigned to interval \(\tau_k^{(r)}\), we compute
\begin{equation}
\begin{gathered}
y_{f,j}
=\phi_f\!\left(x_{\mathrm{rec}};W_{f,j}\right),
\qquad
\hat y_{f,j}^{(r)}
=g_f\!\left(z^{(r)}(W_{f,j})\right),\\
z^{(r)}(W)
=\operatorname{Read}_W\!\left(H^{(r)}\right)
=\sum_{k=1}^{K}\alpha_k^{(r)}(W)h_k^{(r)},\qquad
\alpha_k^{(r)}(W)
=\frac{|W\cap\tau_k^{(r)}|}
{\sum_{q=1}^{K}|W\cap\tau_q^{(r)}|}.
\end{gathered}
\label{eq:interval-readout}
\end{equation}
The query interval is not a learned query token; it selects readout support without restricting the encoder's receptive field, as tokens are contextualized over the complete input.

\noindent\runinheading{Feature-prediction loss.} Within each input, we compute the Huber loss between the predicted and training-normalized target values for each feature \(f\) at each assigned analysis interval~\citep{huber1964robust}.
We sum these losses and divide by the total number of assigned intervals to obtain \(\mathcal{L}_f^{(r)}(x)\).
Intervals with unavailable targets or invalid readouts contribute zero loss but remain included in the denominator (\S~\ref{app:loss}).

\subsection{Retain-and-extend pretraining}
\label{sec:method-rae}
\label{sec:method-encoder}

\input{main4/figures/rae_ppg_stages_fig2}

Longer inputs support additional features but still contain the pulse structure observed in shorter inputs.
We therefore separate read access from parameter updates to reuse information learned at shorter signal durations without overwriting the corresponding parameters.
Pretraining proceeds from 10 to 30 to 240~s, with each successive pretraining stage initialized from the preceding checkpoint.

\noindent\runinheading{Nested read and protected write.} We partition the encoder into duration-specific parameter groups \(G_{10}\), \(G_{30}\), and \(G_{240}\), each containing four attention heads and 512 feed-forward units in each of six pre-norm Transformer layers (Figure~\ref{fig:rae-stages}).
We assign the parameters of the local feature extractor, token projection, and layer normalization, together with shared bias terms, to \(G_{10}\).
These parameters are used for all input durations but are updated only during the 10~s stage.
Each input duration has a separate embedding, updated only during its corresponding stage (\S~\ref{app:allocation}).
For signal duration \(r\), \(\mathcal A_r\) denotes the groups read and \(\mathcal U_r\) those updated at its pretraining stage: \(\mathcal A_{10}=G_{10}\) and \(\mathcal U_{10}=G_{10}\); \(\mathcal A_{30}=G_{10}\cup G_{30}\) and \(\mathcal U_{30}=G_{30}\); \(\mathcal A_{240}=G_{10}\cup G_{30}\cup G_{240}\) and \(\mathcal U_{240}=G_{240}\).
Nested read activates the heads and feed-forward units in \(\mathcal A_r\), allowing shorter-duration groups and the current-duration group to contribute to the same token representation.
Protected write updates only \(\mathcal U_r\), preserving all other encoder parameters and optimizer states.
Because groups assigned to longer signal durations are inactive for shorter inputs, the shorter-duration input-specific computations remain unchanged during subsequent pretraining.

\noindent\runinheading{Retained-target supervision.} Preserving the computation for a shorter signal duration does not ensure that its features remain decodable from longer-input representations.
Newly active groups contribute to the shared residual stream, so frozen earlier groups can receive different intermediate activations when processing longer inputs.
We therefore retain selected targets introduced at shorter signal durations while learning newly introduced features.
Each retained target keeps its signal-processing definition, analysis interval, training-split normalization, and frozen prediction head.
Gradients pass through the head and token readout to update only the current-duration group.

\noindent\runinheading{Stage objectives.} Let \(\mathcal F_r\) denote targets introduced at signal duration \(r\) and \(\mathcal R_{r'\to r}\subseteq\mathcal F_{r'}\) those retained from the immediately preceding pretraining stage.
Weighted sums of their feature losses (Section~\ref{sec:method-supervision}) define \(\mathcal L_{\mathrm{native}}^{(r)}\) and \(\mathcal L_{\mathrm{retain}}^{(r'\to r)}\), respectively: \(\mathcal L^{(10)}=\mathcal L_{\mathrm{native}}^{(10)}\), \(\mathcal L^{(30)}=\mathcal L_{\mathrm{native}}^{(30)}+\lambda_{10\to30}\mathcal L_{\mathrm{retain}}^{(10\to30)}\), and \(\mathcal L^{(240)}=\mathcal L_{\mathrm{native}}^{(240)}+\lambda_{30\to240}\mathcal L_{\mathrm{retain}}^{(30\to240)}\).
The coefficients \(\lambda_{r'\to r}\) scale retained-target supervision.
We retain all nine 10~s targets at 30~s and both 30~s targets at 240~s, with no direct retention term from 10~s to 240~s.
New prediction heads are trained only at their introduction stage.
Stage-specific loss coefficients and retained-query placements are given in \S~\ref{app:loss}--\ref{app:retention}.
After pretraining, prediction heads are removed.
One final checkpoint produces \(H^{(r)}\) using \(\mathcal A_r\) for all three signal durations.

%% file: main4/figures/rae_ppg_stages_fig2.tex
% PI-supplied vector master, final_20260924; preserve legibility at full text width.
% Use the official template's native caption spacing, with no added vertical gap.
\begin{figure}[!t]
  \centering
  \includegraphics[width=1.000\textwidth]{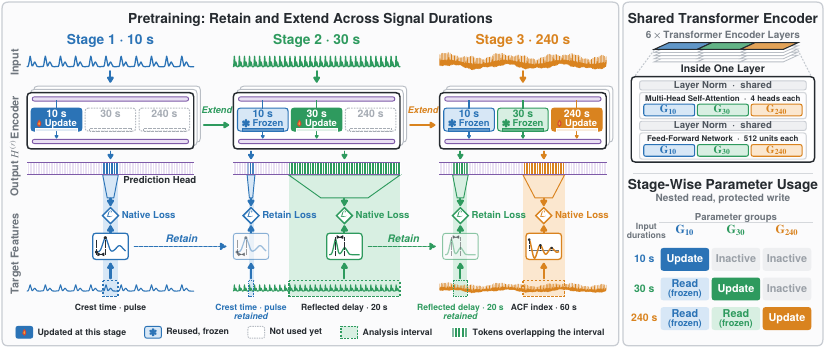}
  \caption{RAE-PPG architecture and pretraining. Later stages reuse frozen groups and learn new targets while retaining selected earlier targets.}
  \label{fig:rae-stages}
\end{figure}

%% file: main4/sections/experiments_main4.tex
\section{Experiments}
\label{sec:experiments}

\noindent\runinheading{Pretraining.} We pretrained RAE-PPG on the MIMIC-III Waveform Database Matched Subset~\citep{moody2020mimicmatched,johnson2016mimic} and VitalDB~\citep{lee2022vitaldb}, using approximately 11.79, 7.26, and 1.34 million eligible input segments at the 10~s, 30~s, and 240~s stages, respectively. Encoder inputs were low-pass filtered at 12~Hz, resampled to 64~Hz, and z-scored using valid waveform samples. All recordings from each MIMIC-III participant or VitalDB surgical case were assigned to only one of the training, validation, or test sets. Preprocessing and optimization details are provided in \S~\ref{app:method}.

\noindent\runinheading{Downstream transfer across tasks and durations.} We evaluate the final frozen RAE-PPG encoder on 18 tasks from eight datasets and compare it with PaPaGei-P/S~\citep{pillai2025papagei}, Pulse-PPG~\citep{saha2025pulseppg}, SIGMA-PPG~\citep{guo2026sigmappg}, and AnyPPG~\citep{nie2025anyppg}. RAE-PPG uses mean-pooled token representations, while each baseline uses a checkpoint-specific interface, with duration and normalization adaptations detailed in \S~\ref{app:foundation-interfaces}. For each task, we assign one input duration using a prior-study-informed feature--task rationale together with the recording and label setting, choosing a duration by which the relevant feature families have been introduced in RAE-PPG. Seven tasks use 10~s, five use 30~s, and six use 240~s. For each task, all encoders are evaluated on the same labeled samples, input duration, and participant- or case-level splits. Following PaPaGei~\citep{pillai2025papagei}, probe selection and fitting are restricted to the training split, while the original validation and test splits are combined for final held-out evaluation. Regularization is selected by grouped four-fold cross-validation within the training split; standardization is fitted within each training fold and refitted on the full training split before held-out evaluation. Classification tasks use L2-regularized logistic regression and regression tasks use ridge regression; we report AUROC, macro-F1, or MAE according to the task (\S~\ref{app:downstream-details} and~\ref{app:linear-evaluation}). For 13 tasks with PPG recordings long enough to construct all three input durations, we also evaluate matched 10~s, 30~s, and 240~s inputs centered at the same recording point and paired with the same label. We use the same final frozen encoder with its duration-specific read sets and fit a separate linear probe at each duration using the same participant- or case-level splits. With a reference duration defined for each task, Retention tests whether performance at every longer duration remains within a prespecified margin of the reference, whereas Extend tests improvement from the immediately shorter duration to the reference. Decisions use paired confidence intervals with Holm correction (\S~\ref{app:cross-duration-scores}--\ref{app:retention-margin-sensitivity}); grouped out-of-fold evaluation is reported in \S~\ref{app:oof-cross-duration}.

\noindent\runinheading{Signal-feature retention and extension.} We test whether earlier features remain decodable from longer-input representations and whether later-stage features are more decodable at their introduction duration than from the preceding shorter input. Using recordings held out from pretraining, we apply the same final frozen encoder at every duration and fit separate ridge probes for each feature and interval. For feature retention, we assess continued decodability using the shorter stage's pooling interval; downstream Retention tests task-performance preservation within a prespecified margin. Extension compares the same later-stage targets decoded from mean-pooled representations at the preceding shorter duration and their introduction duration. Target values, analysis intervals, and validity masks are matched within each comparison. We quantify feature decodability as the fraction of Huber error removed relative to a fixed constant baseline, \(\mathrm{gain}=1-E_{\mathrm{probe}}/E_0\). The baseline predicts the median of each feature estimated from the pretraining training split, making gain 0 at baseline performance, 1 at zero probe error, and negative when probe error exceeds the baseline. This normalization allows decodability to be compared across features with different scales (\S~\ref{app:physiological-protocol}).

\noindent\runinheading{Downstream prediction from signal features.} For each downstream task, we identify PPG feature families that prior studies associate with that task. We first test whether features from these families predict the task labels in our downstream data. We then test whether the corresponding feature information remains accessible in the frozen representation. The Feature-to-Task Bridge evaluates 22 core relations used to motivate the task-duration assignments, together with 11 additional prior-study relations. For each relation, task labels are predicted either from feature values computed directly from PPG using the pretraining definitions or from the corresponding feature values decoded from frozen representations. The linear feature probes used for the latter are trained without downstream labels. Both analyses are compared with 32 participant-reassignment controls that jointly reassign feature-family vectors and their availability masks across participants while keeping the downstream samples and labels fixed; the full feature-decoding and downstream-prediction procedure is repeated for each control. Control construction and sensitivity analyses are detailed in \S~\ref{app:bridge-protocol}. A relation is positive when the one-sided 95\% lower confidence bound for improvement over the median control exceeds zero (\S~\ref{app:bridge-protocol}).

\noindent\runinheading{Encoder organization and retained-target supervision.} To test how the two design components contribute to retain-and-extend behavior, we train four variants: nested read with protected write and a fully shared encoder, each with retained-target supervision enabled or disabled. All variants use the same data, native targets, initialization, stage order, optimization schedule, and training budget. We first test whether shorter-duration learning is preserved during later pretraining by comparing representations, feature decodability, and downstream predictions across checkpoints with the 10~s or 30~s input fixed. We then compare the final encoders to test whether retained-target supervision keeps earlier features accessible from longer-input representations while measuring newly introduced feature decodability and downstream transfer (\S~\ref{app:mechanistic-protocol}).

\noindent\runinheading{Reuse of prior-stage learning.} Finally, we test whether prior-stage learning is reused when acquiring features introduced at the next stage. At each transition, 10~s to 30~s and 30~s to 240~s, two matched training continuations start from the same parent checkpoint. LEARNED retains the learned previous-stage parameters, whereas FULL-PRIOR INITIAL restores them to their initial values, including the local feature extractor, token projection and normalization layers, and earlier attention/FFN groups. In both conditions, these parameters remain active in the forward pass but are frozen, and retained-target supervision is disabled. All other training conditions are matched. After each continuation, independent ridge probes predict the newly introduced features from frozen mean-pooled representations. We report feature gain and the percentage reduction in probe error achieved by LEARNED relative to FULL-PRIOR INITIAL (\S~\ref{app:shorter-group-reuse}).

\section{Results}
\label{sec:results}

\subsection{Downstream transfer}
\label{sec:transfer-results}
\label{sec:functional-trajectory-results}
\input{main4/tables/rae_ppg_transfer_combined_main4}

\noindent\runinheading{Transfer across tasks.} RAE-PPG showed broad transfer across the 18 evaluated tasks rather than strong performance confined to a particular task type or input duration. The final frozen encoder achieved the best observed score on 12 of 18 tasks and ranked first or second on 17, giving the lowest overall mean rank among the six models, 1.44 versus 2.39 for AnyPPG (Table~\ref{tab:transfer-all}). Top-two performance covered six of seven tasks at 10~s and every task evaluated at 30~s and 240~s, with the lowest mean rank for both regression (1.63) and classification (1.30). The learned representation therefore remained broadly useful across tasks evaluated at all three durations, rather than deriving its aggregate advantage from a narrow subset of settings. \S~\ref{app:loso-evaluation}--\ref{app:token-count} and~\ref{app:matched-retrain} report robustness checks and comparison-model retraining.

\runinbreak
\runinheading{Task-dependent effects of input duration.} Increasing input duration did not consistently improve downstream performance. Retention was confirmed for 2 of 8 applicable tasks and Extend for 6 of 11. Both sleep-staging tasks met Retention, while all three MESA event-burden tasks improved from 30~s to 240~s. WESAD arousal instead reached its highest AUROC at 30~s (Table~\ref{tab:cross-duration-ci} and \S~\ref{app:cross-duration-scores}--\ref{app:retention-margin-sensitivity}, with grouped OOF results in \S~\ref{app:oof-cross-duration}). Additional observation time therefore changed downstream utility selectively, benefiting some tasks while offering little benefit or reducing performance for others.

\subsection{Signal-feature information across durations}
\label{sec:physiological-retain-extend-results}
\label{sec:feature-task-bridge-results}
\input{main4/tables/rae_ppg_evidence_wrap_main4}

\runinheading{Earlier-feature retention in longer inputs.} Despite these task-dependent changes, longer-input representations continued to carry information about features learned at earlier stages. The nine 10~s features achieved a mean gain of 0.830 when decoded from 30~s representations, only 0.008 below their native-duration gain. The two 30~s features achieved a mean gain of 0.666 at 240~s, compared with 0.898 at their native 30~s duration (Table~\ref{tab:physiological-evidence}A and \S~\ref{app:physiological-protocol}). Earlier-feature accessibility was therefore nearly unchanged from 10~s to 30~s and remained recoverable, although reduced, from 30~s to 240~s. Thus, moving to a longer input did not simply replace shorter-duration information. Earlier-stage feature information remained available in the longer-input representations.
\runinbreak
\runinheading{Acquisition of later-stage features.} The features introduced at the 30~s and 240~s stages were recoverable from the final encoder representations, showing that the encoder learned the additional signal features targeted at each stage. Mean gain reached 0.898 for the two 30~s features and 0.755 for the five 240~s features (Table~\ref{tab:physiological-evidence}A). When the same later-stage target values were predicted from representations of the preceding shorter inputs, mean gain was lower: 0.611 from 10~s representations for the 30~s targets and 0.579 from 30~s representations for the 240~s targets. Providing the input duration at which these features were introduced increased gain by 28.7 and 17.6 percentage points, respectively. Both 30~s features showed higher gain with 30~s than with 10~s representations, and four of the five 240~s features showed higher gain with 240~s than with 30~s representations. Together with the retention results, this shows that each duration extension broadens the feature information available in the representation: information from the preceding stage remains accessible while additional features supported by the longer input are learned.

\runinheading{Downstream prediction from signal features.} The signal features recovered from the frozen representation also carried information useful for downstream prediction. For 13 of the 22 core task--feature-family relations used in the primary duration rationale, both the directly computed PPG features and the corresponding decoded features predicted the downstream labels better than participant-reassigned controls. The same pattern was observed for five of the 11 additional prior-study relations, yielding 18 of 33 relations in which both pathways showed downstream predictive value (Table~\ref{tab:physiological-evidence}B and \S~\ref{app:bridge-protocol}). This agreement was most frequent for morphology/vascular-timing relations (8/9), followed by rate (8/16) and beat-interval variability (2/8). Thus, for many relations, the same task--feature-family relationship observed with features computed directly from PPG was also observed when those features were estimated from the RAE-PPG representation. This shows that the representation preserves information about these signal features that remains useful for downstream prediction, rather than merely making the pretraining targets decodable.

\subsection{Preservation and reuse across durations}
\label{sec:mechanistic-results}
\label{sec:shorter-group-reuse-results}
\input{main4/tables/rae_ppg_ablation_wrap_main4}

\runinheading{Preservation during later pretraining.} Shorter-input representations remained unchanged throughout later pretraining only when earlier parameter groups were protected from subsequent updates. Both nested/protected variants produced identical shorter-input representations and predictions across later stages, whereas earlier-feature decodability decreased in both fully shared variants. In the fully shared variant without retained-target supervision, mean gain fell by 16.4, 18.7, and 38.1 percentage points for the 10~s to 30~s, 10~s to 240~s, and 30~s to 240~s comparisons, respectively. Retained-target supervision reduced these losses in the fully shared encoder but did not eliminate them (\S~\ref{app:ablation-transfer}). Protected write therefore preserves the computations used for shorter inputs, which retained-target supervision alone does not guarantee in a fully shared encoder.
\runinbreak
\runinheading{Retained-target supervision and longer-input accessibility.} Within the protected organization, re\-tained-target supervision had a different effect. It improved earlier-feature accessibility when the encoder processed a longer input. Mean gain increased from 0.710 to 0.830 for 10~s features read from 30~s representations and from 0.534 to 0.666 for 30~s features read from 240~s representations, corresponding to 12.1 and 13.2 additional percentage points of baseline error removed (Table~\ref{tab:ablation-design-main} and \S~\ref{app:fourarm-feature-accessibility}). By comparison, native-duration decodability of the newly introduced features changed by only 0.60 and 0.65 percentage points. Retained-target supervision therefore primarily improved access to previously learned feature information from longer inputs, complementing the preservation of shorter-input computations provided by protected write.

\noindent\runinheading{Preservation--adaptation trade-off.} The full design also showed a different balance of downstream performance across durations. RAE-PPG achieved the lowest mean rank among the four ablation variants, 2.03 versus 2.56 for the fully shared encoder with retained-target supervision (Table~\ref{tab:ablation-design-main}). With retained-target supervision enabled in both encoders, RAE-PPG led on six of seven 10~s tasks and all five 30~s tasks, whereas the fully shared encoder led on five of six 240~s tasks and achieved higher native 240~s feature decodability (\S~\ref{app:ablation-transfer}--\ref{app:fourarm-feature-accessibility}). The ablation therefore exposes a preservation--adaptation trade-off. Protected training showed stronger transfer across earlier and intermediate durations, while fully shared training showed stronger adaptation to the terminal duration.

\noindent\runinheading{Reuse of prior-stage learning.} Earlier-stage learning was not only preserved but also useful for acquiring newly introduced signal features. Keeping the learned inherited pathway reduced the remaining baseline-normalized Huber error by 34.7\% (95\% CI: 29.2--39.8\%) from 10~s to 30~s and 35.2\% (32.4--38.1\%) from 30~s to 240~s relative to resetting the inherited pathway to its initial values. All seven feature-level confidence intervals favored LEARNED (\S~\ref{app:shorter-group-reuse}). The inherited pathway was frozen in both conditions, so it could not acquire additional information during either continuation. The consistently better acquisition of new features with the learned pathway therefore shows that prior-stage learning was reused to support learning at the next duration.

%% file: main4/tables/rae_ppg_transfer_combined_main4.tex
% Consolidated regression/classification table; numerical rows preserved verbatim.
\begin{table}[!ht]
\topcaption{Frozen-encoder downstream comparison (95\% CIs in brackets).}
\label{tab:transfer-all}
\begingroup
\makeatletter
\edef\@currentlabel{\thetable A}\label{tab:transfer-main}
\edef\@currentlabel{\thetable B}\label{tab:transfer-classification}
\makeatother
\endgroup
\centering
\setlength{\parskip}{0pt}
\fontsize{6.5}{7.5}\selectfont
\setlength{\tabcolsep}{0.75pt}
\renewcommand{\arraystretch}{1.0}
\setlength{\aboverulesep}{1pt}\setlength{\belowrulesep}{1pt}
\newcommand{\scoreci}[3]{\mbox{\fontsize{6}{7}\selectfont #1\,[#2--#3]}}
\renewcommand{\tabularxcolumn}[1]{m{#1}}
\begin{tabularx}{\textwidth}{@{}>{\raggedright\arraybackslash}X*{6}{>{\centering\arraybackslash}m{55pt}}@{}}
\toprule
Downstream task & PaPaGei-P & PaPaGei-S & Pulse-PPG & SIGMA-PPG & AnyPPG & \bestscore{RAE-PPG} \\
\midrule
\multicolumn{7}{@{}l}{\textbf{A. Regression: MAE $\downarrow$}} \\
DaLiA: HR\textsuperscript{10} &
\scoreci{13.9397}{12.25}{15.86} & \scoreci{16.2506}{13.25}{19.37} & \scoreci{10.4389}{8.34}{12.71} & \scoreci{14.8701}{12.28}{18.98} & \scoreci{\bestscore{8.1192}}{5.54}{11.30} & \scoreci{\secondbest{8.6522}}{6.31}{11.18} \\
BUT PPG: HR\textsuperscript{10} &
\scoreci{10.9192}{8.05}{14.40} & \scoreci{10.8344}{8.06}{14.22} & \scoreci{10.5904}{8.00}{13.76} & \scoreci{10.6122}{7.91}{13.87} & \scoreci{\bestscore{9.6609}}{7.24}{12.62} & \scoreci{\secondbest{10.5144}}{8.21}{13.37} \\
PTT-PPG: SBP\textsuperscript{10} &
\scoreci{9.9489}{5.53}{15.25} & \scoreci{10.4675}{5.81}{15.99} & \scoreci{\secondbest{9.8620}}{5.36}{15.32} & \scoreci{10.3279}{5.64}{15.98} & \scoreci{10.1314}{5.63}{15.52} & \scoreci{\bestscore{9.5291}}{5.24}{14.87} \\
PTT-PPG: DBP\textsuperscript{10} &
\scoreci{\secondbest{5.6742}}{3.94}{7.48} & \scoreci{5.9042}{4.02}{7.90} & \scoreci{5.7877}{4.10}{7.55} & \scoreci{6.0150}{4.19}{7.91} & \scoreci{5.8514}{4.03}{7.76} & \scoreci{\bestscore{5.4489}}{3.78}{7.21} \\
DREAMT: Age\textsuperscript{10} &
\scoreci{\bestscore{10.8791}}{8.66}{13.41} & \scoreci{11.9138}{9.55}{14.51} & \scoreci{11.0811}{8.63}{13.91} & \scoreci{12.1837}{9.58}{14.75} & \scoreci{\secondbest{10.9798}}{8.37}{13.81} & \scoreci{11.8329}{9.13}{14.75} \\
MESA$^*$: A/H\textsuperscript{240} &
\scoreci{0.0887}{0.086}{0.091} & \scoreci{0.0897}{0.087}{0.092} & \scoreci{0.0893}{0.087}{0.092} & \scoreci{0.0885}{0.086}{0.091} & \scoreci{\secondbest{0.0867}}{0.084}{0.089} & \scoreci{\bestscore{0.0834}}{0.081}{0.086} \\
MESA$^*$: Arousals\textsuperscript{240} &
\scoreci{0.0464}{0.045}{0.047} & \scoreci{0.0474}{0.046}{0.049} & \scoreci{0.0474}{0.046}{0.048} & \scoreci{0.0460}{0.045}{0.047} & \scoreci{\secondbest{0.0437}}{0.043}{0.045} & \scoreci{\bestscore{0.0432}}{0.042}{0.044} \\
MESA$^*$: SpO$_2$\textsuperscript{240} &
\scoreci{\secondbest{0.2008}}{0.198}{0.204} & \scoreci{0.2085}{0.205}{0.211} & \scoreci{0.2021}{0.199}{0.205} & \scoreci{0.2078}{0.204}{0.211} & \scoreci{0.2022}{0.199}{0.206} & \scoreci{\bestscore{0.1768}}{0.174}{0.180} \\
\cmidrule(lr){1-7}
Mean rank $\downarrow$ & 3.25 & 5.63 & 3.38 & 4.63 & \secondbest{2.50} & \bestscore{1.63} \\
\midrule
\multicolumn{7}{@{}l}{\textbf{B. Classification: AUROC/macro-F1 $\uparrow$}} \\
DeepBeat: AF\textsuperscript{10} &
\scoreci{0.6694}{0.637}{0.719} & \scoreci{0.6413}{0.609}{0.687} & \scoreci{0.6877}{0.632}{0.719} & \scoreci{0.7357}{0.680}{0.779} & \scoreci{\bestscore{0.8443}}{0.768}{0.890} & \scoreci{\secondbest{0.7774}}{0.716}{0.809} \\
DeepBeat: Quality\textsuperscript{10\ensuremath{\dagger}} &
\scoreci{0.5981}{0.554}{0.629} & \scoreci{0.5969}{0.543}{0.640} & \scoreci{\secondbest{0.6502}}{0.608}{0.679} & \scoreci{0.6350}{0.596}{0.660} & \scoreci{0.6427}{0.604}{0.672} & \scoreci{\bestscore{0.6718}}{0.636}{0.690} \\
WESAD: Stress\textsuperscript{30} &
\scoreci{0.8320}{0.751}{0.922} & \scoreci{0.8435}{0.765}{0.922} & \scoreci{0.8334}{0.721}{0.931} & \scoreci{0.8859}{0.821}{0.944} & \scoreci{\secondbest{0.9335}}{0.892}{0.971} & \scoreci{\bestscore{0.9514}}{0.918}{0.979} \\
WESAD: Arousal\textsuperscript{30} &
\scoreci{0.7106}{0.587}{0.841} & \scoreci{0.7198}{0.570}{0.871} & \scoreci{0.7308}{0.579}{0.888} & \scoreci{0.7381}{0.631}{0.847} & \scoreci{\secondbest{0.7729}}{0.683}{0.908} & \scoreci{\bestscore{0.8571}}{0.758}{0.959} \\
WESAD: Valence\textsuperscript{30} &
\scoreci{0.6253}{0.553}{0.715} & \scoreci{0.5613}{0.427}{0.695} & \scoreci{0.5280}{0.486}{0.627} & \scoreci{0.6547}{0.517}{0.783} & \scoreci{\secondbest{0.6587}}{0.581}{0.760} & \scoreci{\bestscore{0.6973}}{0.602}{0.841} \\
DaLiA: Activity\textsuperscript{30\ensuremath{\dagger}} &
\scoreci{0.2636}{0.199}{0.315} & \scoreci{0.1429}{0.108}{0.176} & \scoreci{0.2919}{0.190}{0.363} & \scoreci{0.2740}{0.203}{0.349} & \scoreci{\secondbest{0.3135}}{0.233}{0.389} & \scoreci{\bestscore{0.4078}}{0.312}{0.507} \\
DREAMT: Sleep\textsuperscript{30\ensuremath{\dagger}} &
\scoreci{0.2353}{0.215}{0.257} & \scoreci{0.2234}{0.203}{0.247} & \scoreci{0.2464}{0.225}{0.264} & \scoreci{0.2490}{0.226}{0.272} & \scoreci{\bestscore{0.2746}}{0.242}{0.309} & \scoreci{\secondbest{0.2687}}{0.236}{0.307} \\
MESA$^*$: Sleep\textsuperscript{240\ensuremath{\dagger}} &
\scoreci{0.4295}{0.421}{0.438} & \scoreci{0.4310}{0.423}{0.439} & \scoreci{0.4246}{0.417}{0.433} & \scoreci{0.4469}{0.439}{0.455} & \scoreci{\bestscore{0.4743}}{0.463}{0.485} & \scoreci{\secondbest{0.4735}}{0.464}{0.483} \\
SDB: AHI code\textsuperscript{240} &
\scoreci{0.5426}{0.391}{0.705} & \scoreci{0.5238}{0.370}{0.683} & \scoreci{0.5414}{0.381}{0.714} & \scoreci{\secondbest{0.5977}}{0.442}{0.759} & \scoreci{0.5476}{0.374}{0.718} & \scoreci{\bestscore{0.6779}}{0.525}{0.818} \\
DREAMT: A/H\textsuperscript{240} &
\scoreci{0.5060}{0.424}{0.589} & \scoreci{0.5145}{0.420}{0.608} & \scoreci{0.4932}{0.398}{0.583} & \scoreci{\secondbest{0.5192}}{0.430}{0.604} & \scoreci{0.4887}{0.402}{0.586} & \scoreci{\bestscore{0.5225}}{0.435}{0.614} \\
\cmidrule(lr){1-7}
Mean rank $\downarrow$ & 4.90 & 5.10 & 4.40 & 3.00 & \secondbest{2.30} & \bestscore{1.30} \\
\midrule
\textbf{Overall mean rank} $\downarrow$ & 4.17 & 5.33 & 3.94 & 3.72 & \secondbest{2.39} & \bestscore{1.44} \\
\bottomrule
\end{tabularx}
\par\smallskip
\noindent\parbox{\textwidth}{\scriptsize Bold/underline: best/second-best. Superscripts: input duration (s); \textsuperscript{\ensuremath{\dagger}}macro-F1 (otherwise AUROC). MESA regression targets are event burdens. $^*$MESA used in PaPaGei-P/S and AnyPPG pretraining. Details: \S~\ref{app:downstream-details}--\ref{app:linear-evaluation}; abbreviations: end of B.1\textquotesingle s first paragraph (p.~\pageref{app:downstream-details}).}\par
\end{table}

%% file: main4/tables/rae_ppg_evidence_wrap_main4.tex
\begin{wraptable}{r}{0.47\textwidth}
% Position the complete caption/table block at the adjacent first text line.
\vspace{-2\baselineskip}
\setlength{\belowcaptionskip}{\baselineskip}
\topcaption{Signal-feature retention, extension, and downstream prediction.}
\label{tab:physiological-evidence}
\centering
% Local table enlargement; native caption spacing retained.
\setlength{\aboverulesep}{0.5pt}
\setlength{\belowrulesep}{0.5pt}
\fontsize{8}{9}\selectfont
\setlength{\parskip}{0pt}
\setlength{\tabcolsep}{2pt}
\renewcommand{\arraystretch}{1.00}
\begin{tabularx}{\linewidth}{@{}>{\raggedright\arraybackslash}X>{\raggedright\arraybackslash}p{0.45\linewidth}@{}}
\toprule
\textbf{A. Retention / extension} & Gain or $\Delta$ gain [95\% CI] \\
\midrule
retention $10\to30$~s \mbox{($n=9$)} & 0.830 [0.815, 0.846] \\
extension $10\to30$~s \mbox{($n=2$)} & $0.611\rightarrow0.898$;\newline $+0.287$ [0.267, 0.307] \\
retention $30\to240$~s \mbox{($n=2$)} & 0.666 [0.645, 0.687] \\
extension $30\to240$~s \mbox{($n=5$)} & $0.579\rightarrow0.755$;\newline $+0.176$ [0.163, 0.189] \\
\bottomrule
\end{tabularx}
\par\smallskip
\setlength{\tabcolsep}{1.5pt}
\begin{tabularx}{\linewidth}{@{}>{\raggedright\arraybackslash}Xrrrr@{}}
\toprule
\textbf{B. Feature family} & Relations & PPG & Decoded & Both \\
\midrule
Rate & 16 & 9 & 8 & 8 \\
Morphology / vascular timing & 9 & 9 & 8 & 8 \\
Beat-interval variability & 8 & 2 & 3 & 2 \\
\midrule
All 33 relations & 33 & 20 & 19 & 18 \\
\bottomrule
\end{tabularx}
\par\smallskip
\noindent\parbox{\linewidth}{\fontsize{8}{9}\selectfont A: $n$ counts targets. retention: aligned earlier features; extension: new later-stage features. Intersection-union test: $p_{\mathrm{IUT}}<.001$. B: positive relations; Both: intersection. Families may share tasks.}
\end{wraptable}

%% file: main4/tables/rae_ppg_ablation_wrap_main4.tex
\begin{wraptable}{R}{0.47\textwidth}
% Position the complete caption/table block at the adjacent first text line.
\vspace{-2\baselineskip}
\setlength{\belowcaptionskip}{\baselineskip}
\topcaption{Ablations of parameter organization and retained-target supervision.}
\label{tab:ablation-design-main}
\centering
% Local table enlargement; native caption spacing retained.
\setlength{\aboverulesep}{0.5pt}
\setlength{\belowrulesep}{0.5pt}
\setlength{\parskip}{0pt}
\fontsize{8}{9}\selectfont
\setlength{\tabcolsep}{1.5pt}
\renewcommand{\arraystretch}{1.0}
\begin{tabularx}{\linewidth}{@{}X*{4}{>{\centering\arraybackslash}p{0.135\linewidth}}@{}}
\toprule
 & \multicolumn{2}{c}{Nested/protected} & \multicolumn{2}{c}{Fully shared} \\
\cmidrule(lr){2-3}\cmidrule(l){4-5}
Evaluation & A: ON & D: OFF & B: ON & C: OFF \\
\midrule
Mean rank\newline (18 tasks) $\downarrow$ & 2.03 & 2.58 & 2.56 & 2.83 \\
\midrule
\multicolumn{5}{@{}p{\linewidth}@{}}{\textit{Earlier features: aligned readout, gain $\uparrow$}} \\
$10\to30$~s ($n=9$) & 0.830 & 0.710 & 0.663 & 0.635 \\
$30\to240$~s ($n=2$) & 0.666 & 0.534 & 0.747 & 0.530 \\
\midrule
\multicolumn{5}{@{}p{\linewidth}@{}}{\textit{New features: native-duration mean pooling, gain $\uparrow$}} \\
30~s ($n=2$) & 0.898 & 0.892 & 0.606 & 0.533 \\
240~s ($n=5$) & 0.755 & 0.749 & 0.802 & 0.787 \\
\bottomrule
\end{tabularx}
\par\smallskip
\noindent\parbox{\linewidth}{\fontsize{8}{9}\selectfont Gains are feature-equal means. A = RAE-PPG. ON/OFF: retained-target supervision. Arrows: feature duration to input duration. $n$ counts targets.
A--D label the conditions as in Table~\ref{tab:ablation-terminal}.}
\end{wraptable}

%% file: main4/sections/discussion_main4.tex
\section{Discussion}
\label{sec:discussion}

\noindent\runinheading{Summary.} RAE-PPG organizes self-supervision around signal duration, introducing features as longer observations support them while retaining earlier learning. Across 18 tasks, the frozen encoder showed broad transfer across three durations, although longer inputs did not uniformly improve downstream performance. Feature decoding showed earlier features remained accessible while later-stage features became more decodable on average at their introduction durations. Ablations separated protected write from retained-target supervision, and matched continuations showed learned prior-stage parameters supported newly introduced features, demonstrating preservation and reuse across successive duration extensions. \runinheading{Limitations and future work.} RAE-PPG introduces each feature only when the input duration is long enough to support its prescribed analysis interval. The exact stage durations, however, are design choices. The present 10~s, 30~s, and 240~s configuration establishes retain-and-extend behavior over two successive extensions. Future work can study alternative stage partitions and longer observation horizons, including different numbers and spacings of stages and additional terminal durations, to determine how preservation, reuse, and adaptation behave as features requiring longer analysis intervals are progressively introduced.

%% file: main4/sections/appendix_method_main4_format.tex
% Appendix material deferred from Section~\ref{sec:method}.
% PI-requested concise target reporting (2026-09-06).
% Sources: MODEL_SPEC.md, LOSS_SPEC.md and the R1 target lineage
% REGISTRY_V31_PHYSICAL_CONTEXT_QUALITY_Q1 (extraction, admission and PRV gates).
% 2026-09-17: interval grounding, ACF and LF sentences bound to research/results/raeppg2_r1_q1_span_repeatability_20260917_v1.
% 2026-09-18: A.3 restructured at the user's instruction (selection timing and floors, detector benchmark scope,
% landmark rules checked against targets/dewt_r1 code, adjacent-window definition, scope paragraph, uncentered ACF index).

\section{Method details}
\label{app:method}

\input{main4/sections/appendix_corpus_main4}

\subsection{Input, local feature extractor, and tokenization}
\label{app:local-feature-extractor}
Inputs are single-channel PPG at $64$~Hz (low-pass $12$~Hz view), giving $640$, $1{,}920$, and $15{,}360$ samples for the $10$, $30$, and $240$~s roots.
Before applying the local feature extractor, the input pipeline $z$-scores each complete root
using the mean and population standard deviation of its finite,
quality-mask-valid samples. Invalid samples are set to zero, and no clipping
is applied.
Signal-feature targets are computed from the unnormalized waveform views of
Appendix~\ref{app:targets}; root normalization changes only the encoder input.
Roots with invalid normalization statistics are not admitted to pretraining.
Appendix~\ref{app:foundation-interfaces} specifies downstream preprocessing and handling of invalid roots.
No augmentation is applied and dropout is $0$.

The local feature extractor computes $\mathrm{LN}(\mathrm{Linear}_{1\to128}(x)) + 0.5\,\mathrm{LN}(\mathrm{Conv}_{3}\!\circ\!\mathrm{GELU}\!\circ\!\mathrm{Conv}_{3}(x))$, producing a $128$-dimensional feature vector at each sample.
Both convolutions use kernel size $3$, stride $1$, padding $1$, and biases;
layer-normalization $\epsilon$ is $10^{-5}$.

Token boundaries are obtained by inverting cumulative quality-weighted mass at $K=64$ equal levels; Appendix~\ref{app:token-count} reports pretraining with $K=128$ and $K=256$.
Valid samples have weight $1$, and invalid samples have weight $10^{-4}$.
Each token is an overlap-weighted mean of the local feature vectors inside its boundary pair.

The token projector is $\mathrm{LN}(\mathrm{Linear}_{128\to384}(\text{token}) + \mathrm{Linear}_{32\to384}(\text{pos}) + e_{\text{root}})$.
The vector $e_{\text{root}}\in\mathbb{R}^{384}$ is the learned embedding for the input duration.
For token boundaries $a,b$ in sample coordinates, the position vector contains
$a/N$, $b/N$, $c=(a+b)/(2N)$, $w=(b-a)/N$, $\log w$, the token index divided
by $63$, the duration index divided by $2$, and $\log N/10$. Sine/cosine pairs $\sin(2\pi ft),\cos(2\pi ft)$
use frequencies $f=1,2,4,8,16,32$ for $t=c$ and $f=1,2,4$ for
each normalized boundary $t=a/N,b/N$,
giving 32 coordinates.

\subsection{Encoder}
\label{app:encoder}
The encoder uses multi-head self-attention~\citep{vaswani2017attention} in six pre-norm Transformer layers with $d_{\text{model}}=384$, $12$ heads of dimension $32$, feed-forward width $1{,}536$ (ReLU), no positional attention bias, and a final layer normalization.
Table~\ref{tab:architecture-spec} reports module sizes and parameter ownership; the exported encoder excludes the feature heads.

\input{main4/tables/rae_ppg_architecture_spec_main4}

\subsection{Signal-feature targets}
\label{app:targets}
\label{app:formulas}

Table~\ref{tab:app-targets} specifies the definitions, analysis intervals, and sources of the sixteen pretraining targets. Table~\ref{tab:matched-event-ecg} reports the matched-event ECG comparison for pulse-interval SD, and Table~\ref{tab:target-evidence} summarizes adjacent-window rank consistency and held-out availability of the final targets.

\paragraph{Target-set selection.}
We use a compact rather than exhaustive set of PPG-derived targets. Guided by prior PPG signal-analysis studies, we considered scalar features with explicit definitions whose values can be computed from the PPG waveform or PPG-derived events over analysis intervals no longer than 240~s. We excluded or reserved candidates that required longer intervals or non-PPG information, closely duplicated a selected feature, or lacked a sufficiently specified extraction procedure. The set contains fourteen duration-defining targets spanning rate, morphology/vascular timing, beat-interval variability, and waveform statistics, plus two auxiliary targets for signal quality and full-window pulse cadence. Feature identities and extractor implementations were selected separately. ECG reference beats were used to assess interval detection and PPG--ECG correspondence, not to define pretraining target values.

\paragraph{Preprocessing and pulse detection.}
Pretraining targets are computed from the same MIMIC-III and VitalDB recordings as the encoder inputs, before root-wise normalization. Each recording is split into continuous blocks of finite samples lasting at least 30~s; gaps are not bridged. At the source rate (125~Hz for MIMIC-III and 500~Hz for VitalDB), fourth-order Butterworth filters are applied forward and backward with odd-extension padding to obtain a 12~Hz low-pass measurement view and a $0.5$--$12$~Hz band-pass detection view. Both views are resampled to 64~Hz by linear interpolation on sample times, without an additional anti-aliasing filter. Per-sample masks flag flatlines (at least $0.5$~s with $|\Delta x|\le10^{-6}$), saturation (samples at or beyond the block's $0.5$th or $99.5$th percentile), non-finite or interpolated samples, and filter edges. At each block boundary, the edge mask covers 2~s or $5\%$ of the block duration, whichever is shorter. The source-support ratio is stored alongside these masks.

Peaks are detected once per continuous block on the detection view using a two-event-related-moving-average detector adapted from \citet{elgendi2013peaks}, with $W_1=111$~ms, $W_2=667$~ms, and $\beta=0.02$. For gap rescue, the local reference interval is the median of up to four neighboring inter-peak gaps on each side, excluding the current gap. When an inter-peak gap is at least $1.5$ times this reference, up to three candidates from a secondary peak detector can be inserted. Each candidate must reach at least $40\%$ of the median positive pulse height of the initially detected peaks, remain more than $0.35$ times the local reference interval from the bounding peaks and at least that distance from other inserted peaks, and lie within half the local reference interval of its rhythm-predicted position. Each onset is the leftmost minimum of the detection view strictly between two consecutive peaks. Landmark times are stored in continuous-record coordinates and reused when forming input windows. The 240~s mean-rate target uses a separate implementation of the same detector and accepts intervals of $0.25$--$2$~s. For PRV targets, peak times are additionally refined by parabolic interpolation while preserving the original interval-run adjacency.

\paragraph{Analysis intervals and input durations.}
An analysis interval is the pulse, pulse pair, or waveform segment used to compute one target value. Input duration is the length of the signal supplied to the encoder. Filtering and beat detection use the surrounding continuous recording, which can extend beyond the target's analysis interval. A duration-defining target is introduced at the earliest selected stage whose input contains its prescribed interval (Section~\ref{sec:method-supervision}). When retained at a later stage, it keeps the same analysis interval and training-fitted normalization rather than being recomputed over the entire longer input. Auxiliary targets use the placements in Table~\ref{tab:feature-families}.

\paragraph{Choice of stage number and durations.}
Three stages provide two successive extensions, allowing learning acquired at the intermediate stage to be reused at the next stage. We use 10~s for pulse-local and quality supervision, consistent with morphology-based PPG pretraining~\citep{pillai2025papagei}. The 30~s stage accommodates the prescribed 20~s reflected-delay and 30~s RMSSD intervals. The 240~s stage accommodates the longest selected interval, that of pulse-interval SD, while also introducing the 60~s and 120~s features without changing their analysis intervals. These stage durations are design choices that group target introductions rather than assign a separate stage to every analysis interval.

\paragraph{Pulse landmarks.}
Let $p_i$ and $o_i$ denote peak and onset times in seconds, with
$I_i=p_{i+1}-p_i$ and pulse $[o_i,o_{i+1})$ of duration $T_i=o_{i+1}-o_i$.
For PRV targets, $I_i$ uses the parabolically refined peak times; the
peak-interval rate uses the detector's sample-aligned times.
Pulse-level features use the measurement view $x$. The crest $s_i$ is the
leftmost maximum of $x$ strictly inside the pulse. Slope transit time divides
the crest amplitude $x(s_i)-x(o_i)$ by the largest three-point central
derivative between $o_i$ and $s_i$. The APG is the five-point cubic
Savitzky--Golay second derivative of $x$, with coefficients
$f_s^2(2,-1,-2,-1,2)/7$, where $f_s=64$~Hz, giving waveform units per
second squared. Its $a$ wave is the leftmost maximum between
$o_i$ and the detection-view peak $p_i$ and must be positive. Starting after the $a$ wave, $b$, $c$, and $d$ are identified sequentially as the first strict local minimum, maximum, and minimum, respectively, all before $o_i+0.6\,T_i$. The signed amplitudes are $a_i,b_i,d_i$, and $q_i^b$ is the
$b$-wave time. A ratio is unavailable when $a_i$ does not exceed a
scale-relative numerical floor, and a missing $b$ or $d$ wave removes only the
features that use it. These automatic selections use the APG wave names of \citet{takazawa1998apg}.

PW50 and the split-area ratio share a linear baseline from $x(o_i)$ to the value
at $o_{i+1}$ extrapolated from the last two samples of the pulse. The crossings
$t_i^{\uparrow50}$ and $t_i^{\downarrow50}$ are the last upward crossing
before $s_i$ and the first downward crossing after $s_i$ of half the
baseline-subtracted crest height, located by linear interpolation; any sample
exactly at that level or a missing crossing makes the pulse unavailable.
Among strict local minima after $p_i$ and before $o_i+0.8\,T_i$, we select the one whose subsequent strict local maximum within this range occurs earliest. The selected minimum must follow the steepest post-crest descent and rebound by at least $0.02$ times the pulse excursion; otherwise, the pulse is unavailable. Only when no such
minimum exists, a fallback uses the negative-to-nonnegative APG zero crossing in
the same range with the most negative first derivative. Areas
$A_i^{\rm sys},A_i^{\rm dia}$ are trapezoidal signed integrals of the
baseline-subtracted waveform before and after the split point; the ratio is
unavailable when $A_i^{\rm sys}$ does not exceed a numerical floor, and the areas
describe waveform shape rather than blood volume. \citet{ahn2017aging} split the
pulse area at the dicrotic notch; the fallback branch, baseline, and
notch-acceptance rule are our implementation choices. The reflected point
$q_i^{\rm rp}$ is the first strict local maximum that follows a strict local
minimum after $s_i$ and before $o_i+0.8\,T_i$; without one, it is the largest local maximum of the five-point first derivative after that derivative has become negative within the same range, taking the earliest in a tie.

\paragraph{Spectral purity.}
For spectral purity, $P$ is Welch power, $f_1$ maximizes power over
$0.5$--$3$~Hz, and $B_k=[kf_1-0.1k,kf_1+0.1k]\cap[0.5,8]$.
The estimator uses $512$-sample Hann segments with $50\%$ overlap, mean removal
and an $8192$-point FFT. At 64~Hz, each 10~s query contributes one 512-sample (8~s) segment to this Welch estimate.
The spectrum uses the stored samples without masking or interpolation; the value is unavailable only when a sample in the 10~s query is non-finite, the query is flat, or total power is not positive.
Spectral purity is an auxiliary target, not a common exclusion threshold for the other targets.

\paragraph{Uncentered ACF index.}
For the uncentered ACF index, let $u_0,\ldots,u_{n-1}$ be the query samples after
interpolation to $128$~Hz and a $257$-tap, $4$~Hz low-pass filter with reflected
edges. We compute
$R_\ell=\bigl(\sum_{t=0}^{n-\ell-1}u_tu_{t+\ell}\bigr)/\bigl(\sum_{t=0}^{n-1}u_t^2\bigr)$
without mean removal, as in the equation of \citet{valiaho2021acf}. The 60~s
query must be gap-free with source support of at least $0.5$ and at least 55~s of
unflagged samples; flagged samples remain in both sums, which run over all sample
pairs of the query. We use $\Lambda=\{32,\ldots,512\}$ samples ($0.25$--$4$~s), an implementation choice spanning two pulse cycles at 30~bpm.
For fixed query support, the index is invariant to uniform amplitude scaling but sensitive to additive offsets. Across 11,968 windows of 60~s from the 1,536 groups in the adjacent-window analysis, its Spearman correlation was $0.99$ with the baseline-power fraction, $\bar{u}^2/\overline{u^2}$, and $0.02$ with the mean-removed index. We therefore describe it as a baseline-sensitive waveform index. Decoding this target from root-normalized inputs does not establish recovery of the original waveform baseline.

\paragraph{LF fraction.}
The LF fraction uses the LF ($0.04$--$0.15$~Hz) and total-power ($0$--$0.4$~Hz) bands of \citet{kiselev2024ppgv}. That study used Welch spectra in 2~min sliding windows of longer recordings; we compute one linearly detrended, symmetric-Hann periodogram per 120~s query. The query must be gap-free, with at least $11/12$ unflagged samples, unflagged endpoints, and no flagged run longer than 1~s. Flagged samples remain in the primary estimate. We also recompute the fraction after linearly interpolating those samples and mark the query unavailable if the absolute change exceeds $0.05$.

\paragraph{Native-query placement.}
In 30~s inputs, the 20~s reflected-delay queries start at 0 and 10~s. In 240~s inputs, 60~s queries start at $0,30,\ldots,180$~s, and 120~s queries start at $2,60,120$~s. Native queries covering the full input start at zero. Event-level targets retain their individual pulse or pulse-pair support. Retained-query placements are specified in Appendix~\ref{app:retention}.

\paragraph{Quality control.}
A pulse-level value requires every sample of its pulse to be finite and valid, outside gaps and filter edges, with source support of at least $0.5$ and no flatline, saturation, or interpolation flag. Each target also applies the fiducial, count, and duration requirements specified in Table~\ref{tab:app-targets}. Missing fiducials and numerically degenerate denominators yield unavailable targets.

\paragraph{PRV interval screening.}
For each query window, every admitted interval is compared with the median of up to the five nearest admitted intervals on each side within the same contiguous admitted run. These medians are computed before gating and exclude the interval being assessed. With at least three such neighbors, an interval is kept only if its duration lies within $[0.6,1.6]$ times that median; intervals with fewer neighbors are kept ungated. Gating is query-specific, so overlapping queries can retain different interval sets. With the continuous-record preprocessing fixed, the same physical query has identical interval membership, gating, availability, and target values across enclosing input durations. Successive differences use only pairs that were adjacent in the original pre-gating run and whose two intervals both survive gating. Excluded intervals therefore break adjacency rather than reconnect the surviving intervals. Let $\mathcal C$ contain these valid adjacent pairs, $D=|\mathcal C|$, and $N_I$ denote the number of retained intervals. Pulse-interval SD uses the retained interval set without successive differencing.

The minimum interval counts, valid-pair counts, and merged interval coverage in Table~\ref{tab:app-targets} are availability requirements within the prescribed windows. Targets summarize the retained intervals; the entire window need not be valid. Our pNN50 divides by $D$, the number of valid adjacent pairs, rather than the total interval count used in the ECG-HRV definition~\citepalias{taskforce1996hrv}.

\paragraph{PRV duration references.}
\citet{baek2015ultrashort} reported study-specific minimum durations of 30~s for RMSSD, 60~s for pNN50, and 240~s for SDNN against 5~min references obtained from resting fingertip PPG. In noise-free simulated PPG, \citet{mejiamejia2023duration} reported 30--60~s for RMSSD and 90~s for pNN50. \citetalias{taskforce1996hrv} specifies approximately 2~min for low-frequency power of beat-interval series, a different quantity from the PPG waveform LF fraction. The coverage floors in Table~\ref{tab:app-targets} are operational availability requirements within the prescribed analysis windows.

Across the training corpus, $99.94\%$ of roots passed input normalization and $93.3\%$ of native target rows had available values. Unavailable targets contribute zero loss without imputation (Appendix~\ref{app:loss}).

\paragraph{Selection of extractor realizations.}
Development-stage candidates varied in landmark search, derivative filtering, and baseline handling. Candidate sets and selection rules were fixed before candidate outputs were inspected, with features sharing a landmark selected jointly. Selection used numerical checks, invariance across enclosing input durations for identical physical queries, and availability on a 1,024-group TRAIN panel (512 groups per dataset). Selected development-stage realizations were confirmed once on 512 additional VAL groups (TRAIN for the LF fraction), without re-ranking. The two auxiliary targets were outside this selection. Selection used no downstream labels, task metrics, representation probes, pretraining losses, or TEST data. Subsequent extraction revisions are reflected in the final definitions above; Table~\ref{tab:target-evidence} reports checks of the final implementations, not the development-stage candidates. Candidate-specific settings and development-only results are documented in the accompanying development record.

\input{main4/tables/rae_ppg_target_provenance_main4_format.tex}

\paragraph{Numerical checks.}
The final PRV implementation was separately verified to give identical target values and availability for identical physical query windows across enclosing input durations. The variability, ACF, and LF values used in the adjacent-window analysis matched the production target values bit for bit in 55,296 windows.

\paragraph{PRV screening check.}
In 87 groups with usable ECG-derived beat references, the gate removed $45$--$55\%$ of mismatched intervals in MIMIC-III and $14$--$21\%$ in VitalDB, while removing at most $0.09\%$ of matched intervals. This check evaluates interval screening; the matched-event comparison below evaluates pulse-interval SD values.

\paragraph{Adjacent-window rank consistency.}
We assessed whether relative feature rankings were consistent across adjacent, non-overlapping windows, using the final extractors on a TRAIN analysis panel of 1,024 groups and a TRAIN replication panel of 512 groups. Each participant or case contributed one 480~s block.
Each block was divided into adjacent non-overlapping windows of a given span, and each window received one value: the median of its pulse values for pulse and pulse-pair targets, the mean of its pulse values for reflected delay, and otherwise the target computed over the window. The spans searched were 10, 20, 30, 60, 120 and 240~s for pulse and pulse-pair targets; 20, 30, 60, 120 and 240~s for reflected delay; 30, 60, 120 and 240~s for RMSSD; 60, 120 and 240~s for pNN50 and the ACF index; 120 and 240~s for the LF fraction; and 240~s for interval SD. A 480~s block holds a single adjacent pair of 240~s windows, so longer spans could not be tested.

For each window length, we computed Spearman's $\rho$ across complete adjacent-window pairs and transformed it to the prespecified rank-consistency score $2\rho/(1+\rho)$ using the Spearman--Brown formula. Confidence intervals used 1,000 participant/case-cluster bootstrap draws. This score is not an estimate of single-window measurement reliability; a score of $0.70$ corresponds to $\rho\approx0.54$. A window length met the criterion when the lower 95\% bound was at least $0.70$ overall and in each dataset, with at least 384 groups overall and 128 per dataset contributing complete pairs. A target passed if the shortest length meeting the criterion in both panels did not exceed its assigned input duration.

Twelve of the fourteen evaluated targets met the criterion at the first tested window length (Table~\ref{tab:target-evidence}). The LF fraction and pulse-interval SD did not; rank consistency was a diagnostic, not a target-inclusion rule. For the LF fraction, the minimum lower bounds remained below $0.70$ at 240~s: $0.686$ and $0.565$ in the analysis and replication panels, with 797 and 394 groups contributing complete pairs, respectively. Table~\ref{tab:target-evidence} reports the results at the first tested window length.

\input{main4/tables/rae_ppg_matched_event_ecg_main4}

\input{main4/tables/rae_ppg_target_evidence_main4}

\paragraph{Matched-event comparison of pulse-interval SD.}
We compared PPG-derived and ECG-derived interval SD computed from identical matched interval sets. A fixed hash rule selected 128 MIMIC-III participants and 64 VitalDB cases from TRAIN groups with an eligible 240~s root, independently of target values. ECG-reference qualification retained 51 and 28 groups, respectively; final matched-comparable counts after production-availability and event-matching requirements are reported in Table~\ref{tab:matched-event-ecg}. Groups were not replaced based on comparison outcomes.
ECG reference beats were constructed from WFDB XQRS and GQRS detections matched one-to-one within $\pm50$~ms. A record-specific PPG--ECG lag was estimated from a separate 300--360~s calibration interval and held fixed for the 0--240~s scoring interval; after lag correction, PPG peaks were matched one-to-one to ECG consensus beats within $\pm150$~ms. Pulse-interval SD was then recomputed from identical matched interval sets in the two modalities. Median absolute differences were $2.39$~ms in MIMIC-III and $1.21$~ms in VitalDB (Table~\ref{tab:matched-event-ecg}).

\paragraph{Within-participant variation of the LF fraction.}
We analyzed four consecutive 120~s LF-fraction values from the 480~s analysis and replication blocks. The analysis included 1,153 participants with all four values available. Across dataset/panel combinations, within-participant SD, calculated as the square root of the mean within-participant variance, ranged from $0.107$ to $0.125$. The SD of participant-mean LF fractions ranged from $0.130$ to $0.143$, with within/between variance ratios of $0.645$--$0.837$. Substantial within-participant variation persisted under stricter signal-quality restrictions, although rank consistency improved in some high-quality subsets. Correlation did not decrease consistently with temporal separation across panels. For available LF windows, the 90th percentile of the change after interpolating flagged samples was $0.012$; excluding the $0$-Hz bin from total power preserved the ranking (Spearman $\rho=0.9996$). These analyses do not separate genuine temporal variation from finite-window estimation variability. We use the LF fraction as a window-level low-frequency waveform statistic.

\paragraph{Target decodability.}
All sixteen target summaries had positive decoding gains at their introduction durations on participants or cases held out from pretraining, with each reported 95\% interval above zero (Table~\ref{tab:physiology-all}). These included gains of $0.544$ for the LF fraction and $0.761$ for pulse-interval SD.

\FloatBarrier
\subsection{Loss, normalization, and validity}
\label{app:loss}
Available targets are normalized as
$(y-\operatorname{median})/\max(\operatorname{IQR}/1.349,10^{-4})$,
using statistics fitted on deduplicated available native TRAIN occurrences at each feature's introduction duration.
For sample $u$, input-duration identifier $c$ (denoted by $r$ in the main text), feature $f$, and role
$\eta\in\{\mathrm{native},\mathrm{retained}\}$, let $\Omega^{\mathrm{struct}}_{u,c,f,\eta}$
denote the configured target occurrences before availability and readout masks are applied. The feature loss is
\begin{equation}
\mathcal{L}_{u,c,f,\eta}
=\frac{1}{\max(1,|\Omega^{\mathrm{struct}}_{u,c,f,\eta}|)}
\sum_{j\in\Omega^{\mathrm{struct}}_{u,c,f,\eta}}
z_fa_jv_j\,\rho_1(\hat y_j-\tilde y_j).
\label{eq:loss}
\end{equation}
Here, $z_f$, $a_j$, and $v_j$ indicate a selected target coordinate, target availability, and a valid interval-aligned readout, respectively.
The function $\rho_1$ is the Huber loss with threshold 1 in the normalized coordinates shared by $\hat y_j$ and $\tilde y_j$.
The denominator counts all configured occurrences, including those with unavailable targets or invalid readouts, whose loss contributions are zero.
Per-example feature losses are averaged over the batch and combined using feature weights that do not vary with target availability.

Native weights are stage-specific loss-balancing coefficients: each target receives $1/9$ at 10~s and $1/5$ at both 30~s and 240~s.
The two native targets at 30~s therefore have a total coefficient of $0.4$.
These weights remain unchanged when retained-target supervision is disabled.

\subsection{Retained queries}
\label{app:retention}
At the 30~s stage, all nine targets introduced at 10~s are retained.
Peak-interval rate uses fully contained peak-pair queries, and the seven pulse-morphology targets use fully contained onset-to-onset pulse queries.
Spectral-purity queries span 10~s and start at 0, 10, and 20~s. Each retained feature has coefficient $\lambda_{\mathrm{ret}}/9$.

The 240~s stage retains only normalized reflected delay and PRV-RMSSD.
Reflected-delay queries span 20~s and start at $0,10,\ldots,220$~s (23 placements);
RMSSD queries span 30~s and start at $0,30,\ldots,210$~s (eight placements).
Their coefficients are $\lambda_{\mathrm{ret}}/4$ and $\lambda_{\mathrm{ret}}/6$, respectively.
No direct 10~s-to-240~s retention term is used.

Retained queries reuse the target values and normalization defined at each feature's introduction duration.
Earlier prediction-head rows remain fixed, while gradients pass through the heads and interval-aligned readout to the current-duration encoder group.
We use $\lambda_{\mathrm{ret}}=1.25$.

\subsection{Nested-read, protected-write allocation}
\label{app:allocation}
\label{app:nsa}
In each of the six layers, attention heads $\{0\text{--}3\}$, $\{4\text{--}7\}$, and $\{8\text{--}11\}$,
together with feed-forward units $\{0\text{--}511\}$, $\{512\text{--}1023\}$, and $\{1024\text{--}1535\}$,
form $G_{10}$, $G_{30}$, and $G_{240}$, respectively.
Each attention head owns its query/key/value projection rows and biases and its output-projection columns.
Each feed-forward unit owns its first-layer row and bias and its second-layer column.
Layer norms, shared biases, the final norm, the local feature extractor, and token projection belong to $G_{10}$.
Each input duration has an embedding row updated only at its corresponding stage.
Active prediction-head rows are updated only at their introduction stage; inactive rows are never updated.

During the forward pass at input duration $r$, read masks zero the contributions of attention output-projection columns and feed-forward-unit activations outside the readable set.

Write masks apply to parameter entries, both AdamW moment estimates, and per-entry step counters.
The global gradient norm is computed over writable entries and clipped at $3.0$.
Protected parameter and optimizer-state entries remain bit-identical across later stages.
For a fixed shorter input, the local feature extractor, its input-duration embedding, and all readable parameters remain unchanged;
with zero dropout, its computation is preserved during later pretraining.

\subsection{Training schedule}
\label{app:schedule}
All stages use masked AdamW~\citep{loshchilov2019adamw} with $\beta=(0.9,0.95)$ and $\epsilon=10^{-8}$.
Table~\ref{tab:app-schedule} gives the stage-specific settings.
Each stage saves its terminal checkpoint and sampler/optimizer state; the next stage inherits the parameters and optimizer state.
The final checkpoint after $1{,}323{,}524$ updates is used at all three input durations.
\begin{table}[!t]
\topcaption{Stage-specific training settings and exposure.}
\label{tab:app-schedule}
\centering
\small
\begin{tabular}{lccc}
\toprule
Setting & 10~s & 30~s & 240~s \\
\midrule
Initialization & Random & 10~s checkpoint & 30~s checkpoint \\
Optimizer updates & 857{,}664 & 340{,}309 & 125{,}551 \\
Micro-batch $\times$ accumulation & $64\times1$ & $64\times1$ & $8\times4$ \\
Training roots & 11{,}785{,}831 & 7{,}257{,}646 & 1{,}338{,}827 \\
Processed roots & 54.9 M & 21.8 M & 4.02 M \\
Nominal epochs & 4.66 & 3.0 & 3.0 \\
Weight decay & 0.0132 & 0.0160 & 0.0101 \\
Warm-up optimizer updates & 21{,}442 & 8{,}344 & 3{,}066 \\
Training time (GPU-h) & 15.3 & 10.0 & 3.5 \\
\bottomrule
\end{tabular}
\par\noindent\raggedright
Micro-batches count input roots. Training roots are normalization-valid inputs in each stage's pool; processed roots include repeated presentations.
Training time includes checkpoint writes and in-loop diagnostics, but excludes data preloading and separate evaluations.
\end{table}

Let $T$ be the number of optimizer updates in a stage and $t=0,\ldots,T-1$ its zero-based update index.
We use $\eta_{\max}=3\times10^{-4}$ and $\eta_{\min}=0.1\eta_{\max}$.
Each stage has a separate learning-rate schedule starting at $\eta_{\min}$.
At 10~s, the rate is evaluated at $f(t)=t/(T-1)$:

\begin{equation}
\eta_{10}(t)=
\begin{cases}
\eta_{\min}+(\eta_{\max}-\eta_{\min})\dfrac{f(t)}{0.025}, & f(t)<0.025,\\[6pt]
\eta_{\min}+\dfrac{\eta_{\max}-\eta_{\min}}{2}
\left[1+\cos\left(\pi\dfrac{f(t)-0.025}{0.975}\right)\right], & f(t)\ge 0.025.
\end{cases}
\label{eq:lr-short}
\end{equation}

The 30~s and 240~s stages use a discrete reference schedule with $P$ positions, of which the first $W$ are warm-up.
At optimizer update $t$, this schedule is evaluated at $j(t)=\min\!\left(P-1,\left\lfloor tP/T\right\rfloor\right)$:
\begin{equation}
\eta_{\mathrm{ref}}(j)=
\begin{cases}
\eta_{\min}+(\eta_{\max}-\eta_{\min})\dfrac{j}{W-1}, & j<W,\\[6pt]
\eta_{\min}+\dfrac{\eta_{\max}-\eta_{\min}}{2}
\left[1+\cos\left(\pi\dfrac{j-W}{P-W-1}\right)\right], & j\ge W,
\end{cases}
\qquad
\eta(t)=\eta_{\mathrm{ref}}\!\left(j(t)\right).
\label{eq:lr-long}
\end{equation}

The schedule parameters are $(T,P,W)=(340{,}309,\allowbreak\,100{,}496,\allowbreak\,2{,}464)$ at 30~s and $(125{,}551,\allowbreak\,18{,}841,\allowbreak\,460)$ at 240~s.
The floor in $j(t)$ repeats reference rates across consecutive optimizer updates. All three stages end at $\eta_{\min}$.

\subsection{Implementation and data access}
\label{app:implementation}
Training used one NVIDIA B200 GPU and one process per stage, totaling $28.8$ GPU-hours (Table~\ref{tab:app-schedule}). Model parameters are float32, automatic mixed
precision is not used, and deterministic algorithms are enforced. The code
leaves PyTorch's TF32 defaults unchanged, so matrix multiplications run in
full float32 while cuDNN convolutions may use TF32. The recorded pretraining environment specifies Python 3.10.19 and PyTorch 2.9.1 with CUDA 12.8 and cuDNN 9.10.2.
The terminal-transfer evaluation records NumPy 2.2.6, SciPy 1.15.3, and scikit-learn 1.7.2.

Dataset releases are identified in Appendices~\ref{app:pretraining-corpus} and~\ref{app:downstream-details} and are used under their respective access conditions;
MESA access is through the National Sleep Research Resource.
Code and weights will be released publicly upon publication. The accompanying package is described in the Reproducibility Statement.

%% file: main4/sections/appendix_corpus_main4.tex
\subsection{Pretraining corpus and partitioning}
\label{app:pretraining-corpus}
Pretraining uses the PLETH channel from MIMIC-III Waveform Matched Subset v1.0
(125~Hz ICU recordings)~\citep{moody2020mimicmatched,johnson2016mimic} and
VitalDB (500~Hz surgical recordings)~\citep{lee2022vitaldb}.
All records from a MIMIC participant or a VitalDB case share one partition.
Partitions are assigned before screening and crop generation using the first
48 bits of a SHA-1 hash of each participant or case key with a fixed prefix,
mapped to TRAIN/VAL/TEST at $0.8/0.1/0.1$.
Table~\ref{tab:pretraining-corpus} summarizes the processed pretraining pool. Continuous blocks are
variable-length intervals between recording gaps. Blocks pass the stored
quality screen when valid-sample fraction is at least $0.8$, flatline fraction
at most $0.25$, and saturation fraction at most $0.05$.

\begin{table}[!htbp]
\topcaption{Pretraining corpus after block-level screening. Counts before and
after screening are shown for participants (MIMIC-III) or cases (VitalDB).
Records are source segment files for MIMIC-III and case recordings for VitalDB.
Hours are interval unions within each source record, before crop selection;
they are not deduplicated across separate records.}
\label{tab:pretraining-corpus}
\centering\footnotesize
\setlength{\tabcolsep}{4pt}
\begin{tabular}{llrrrr}
\toprule
Source & Partition & Groups: before $\to$ after & Records & Blocks & Source hours \\
\midrule
MIMIC-III & TRAIN & 5,614 $\to$ 5,544 & 174,858 & 3,101,986 & 421,887 \\
MIMIC-III & VAL & 729 $\to$ 724 & 21,560 & 465,577 & 66,397 \\
MIMIC-III & TEST & 737 $\to$ 736 & 22,447 & 413,427 & 56,047 \\
VitalDB & TRAIN & 4,930 $\to$ 4,609 & 4,609 & 7,216 & 13,841 \\
VitalDB & VAL & 604 $\to$ 566 & 566 & 898 & 1,698 \\
VitalDB & TEST & 622 $\to$ 582 & 582 & 867 & 1,774 \\
\bottomrule
\end{tabular}
\end{table}

For a root of $n$
samples and a block containing $N$ samples, let $m=\lfloor N/n\rfloor$ and
$k=\min(4,m)$. We select offsets
$n\lfloor j(m-1)/(k-1)\rfloor$, $j=0,\ldots,k-1$, using offset zero when
$k=1$ and no crop when $k=0$. Thus each block contributes at most four
non-overlapping crops per duration; crops at different durations may overlap.
After root-normalization screening, MIMIC-III contributes
$11{,}757{,}100/7{,}229{,}324/1{,}312{,}393$ roots and VitalDB
$28{,}731/28{,}322/26{,}434$ roots at $10/30/240$~s.
The 240~s pools cover $5{,}480$ participants and $4{,}608$ cases; the shorter
pools cover all retained TRAIN groups in Table~\ref{tab:pretraining-corpus}.
Each stage permutes its combined root pool without replacement, with no
additional dataset or participant balancing. VitalDB accounts for $0.24\%$, $0.39\%$ and $1.97\%$ of eligible roots at
$10$, $30$ and $240$~s, respectively.
Table~\ref{tab:app-schedule} reports stage-level root counts and training exposure.

Target-candidate development and the final-extractor consistency analysis use
TRAIN panels; the one-time confirmation of selected realizations used $512$ VAL
participants/cases ($512$ further TRAIN participants/cases for the LF fraction).
The $1{,}308$-group decoding panel in Appendix~\ref{app:physiological-protocol}
contains every TEST participant or case with an eligible $240$~s root, and the
reuse analysis in Appendix~\ref{app:shorter-group-reuse} uses a hash-selected
VAL panel. Feature normalization is fitted
only on unique available native occurrences in TRAIN.

%% file: main4/tables/rae_ppg_architecture_spec_main4.tex
\begin{table}[!htb]
\topcaption{Architecture and parameter ownership. $B$ is batch size, $N$ the
input length in samples, and $Q_h$ and $d_h$ the query count and output width of head $h$.
Counts include biases and normalization parameters.}
\label{tab:architecture-spec}
\centering\fontsize{8}{9}\selectfont
\setlength{\tabcolsep}{3pt}\renewcommand{\arraystretch}{1}
\begin{tabularx}{\linewidth}{@{}l>{\raggedright\arraybackslash}Xr>{\raggedright\arraybackslash}p{.18\linewidth}@{}}
\toprule
Module & Operations and output shape & Parameters & Writable stage \\
\midrule
Local feature extractor & Linear and local two-convolution branch; $B\times N\times128$ & 50,560 & 10~s \\
Token projection & Content + position + duration embedding, LN; $B\times64\times384$ & 64,128 & 10~s; duration embeddings at their corresponding stages \\
Transformer & Six attention/FFN blocks; $B\times64\times384$ & 10,646,784 & Current attention/FFN group; shared norms/biases at 10~s \\
Final LN & $B\times64\times384$ & 768 & 10~s \\
Feature heads & Eight affine maps $384\to d_h$; $B\times Q_h\times d_h$ & 7,315 & Selected rows at their introduction stage \\
\midrule
Exported encoder & Signal tokens; downstream mean is $B\times384$ & 10,762,240 & Frozen for evaluation \\
Complete model & Encoder and feature heads & 10,769,555 & Stage-dependent masks \\
\bottomrule
\end{tabularx}
\end{table}

Each prediction head is a single
affine map, with no hidden layer or output activation. The eight head widths
are $1,7,1,2,3,2,1,2$; sixteen of the nineteen configured output coordinates
are supervised. Heads use PyTorch's default initialization, and the three
unsupervised coordinates keep their initial values.
Retained queries reuse their original head rows.
The encoder has $3{,}672{,}832$, $3{,}544{,}704$, and $3{,}544{,}704$ parameter
entries owned by the 10~s, 30~s, and 240~s groups. Including active head rows,
the corresponding writable counts are $3{,}676{,}297$, $3{,}545{,}474$, and
$3{,}546{,}629$.

Read masks are applied to dense attention/FFN computations. Local feature-extraction cost depends on input length.

%% file: main4/tables/rae_ppg_target_provenance_main4_format.tex
% Compact definitions for the pretraining targets.
\begin{table}[!htb]
\topcaption{The sixteen pretraining targets. \emph{Pulse} denotes
$[o_i,o_{i+1})$; ratios are dimensionless unless a unit is shown.
$P(B)$ sums spectral power in band $B$. Source citations identify the underlying signal-feature definitions; the extraction settings and support requirements are specified in this section. \emph{Adapted from} indicates a changed definition or extraction procedure. PRV denotes variability of PPG-derived pulse intervals.}
\label{tab:app-targets}
\centering
\scriptsize
\renewcommand{\arraystretch}{1}
\setlength{\tabcolsep}{4pt}
\begin{tabularx}{\textwidth}{@{}>{\raggedright\arraybackslash}p{0.22\textwidth}>{\raggedright\arraybackslash}p{0.10\textwidth}>{\raggedright\arraybackslash}X>{\raggedright\arraybackslash}p{0.18\textwidth}@{}}
\toprule
Feature (unit) & Support & Definition & Source \\
\midrule
\multicolumn{4}{@{}l}{\textit{Introduced at 10~s}} \\
Peak-interval rate (bpm) & Peak pair & $60/I_i$; $0.3\leq I_i\leq2$~s & Adapted from \citet{goda2024pyppg} \\
Slope transit time (s) & Pulse & $[x(s_i)-x(o_i)]/\max_{[o_i,s_i]}\dot x$ & \citet{natarajan2022upstroke} \\
APG $b/a$ & Pulse & Signed amplitude ratio $b_i/a_i$ & \citet{takazawa1998apg} \\
APG $d/a$ & Pulse & Signed amplitude ratio $d_i/a_i$ & \citet{takazawa1998apg} \\
Onset-to-APG-$b$ (s) & Pulse & $q_i^b-o_i$ & Adapted from \citet{natarajan2022upstroke} \\
Normalized crest time & Pulse & $(s_i-o_i)/T_i$ & Adapted from \citet{goda2024pyppg} \\
Normalized PW50 & Pulse & $(t_i^{\downarrow50}-t_i^{\uparrow50})/T_i$ & Adapted from \citet{goda2024pyppg}; width-at-50\% precedent: \citet{khalid2018bp} \\
Pulse split-area ratio & Pulse & $A_i^{\rm dia}/A_i^{\rm sys}$ & Adapted from \citet{ahn2017aging} \\
Spectral purity & 10~s & $\sum_{k=1}^3 P(B_k)/P([0.5,8])$ over one 8~s segment & Harmonic-band variant of \citet{elgendi2016sqi} \\
\midrule
\multicolumn{4}{@{}l}{\textit{Introduced at 30~s}} \\
Normalized reflected delay & 20~s & $\operatorname{mean}_i[(q_i^{\rm rp}-s_i)/T_i]$; $\geq10$ valid pulses & Adapted from \citet{hickey2016morphology} \\
PRV-RMSSD (ms) & 30~s & $1000\sqrt{D^{-1}\sum_{(a,b)\in\mathcal C}(I_b-I_a)^2}$; $N_I\geq20$, coverage $\geq15$~s, $D\geq10$ & Adapted from Task Force (\citeyear{taskforce1996hrv}); \citet{mejiamejia2020prv} \\
\midrule
\multicolumn{4}{@{}l}{\textit{Introduced at 240~s}} \\
PRV-pNN50 (\%) & 60~s & $100D^{-1}\sum_{(a,b)\in\mathcal C}\mathbf1[|I_b-I_a|>0.05]$; $N_I\geq40$, coverage $\geq30$~s, $D\geq10$ & Adapted from Task Force (\citeyear{taskforce1996hrv}); \citet{mejiamejia2020prv} \\
Uncentered ACF index & 60~s & $|\Lambda|^{-1}\sum_{\ell\in\Lambda}|R_\ell|$, uncentered $R_\ell$; gap-free, $\geq55$ unflagged seconds & Adapted from \citet{valiaho2021acf} \\
PPGV LF fraction & 120~s & $P([5/120,18/120])/P([0,48/120])$; $\geq7040$ unflagged samples & Adapted from \citet{kiselev2024ppgv} \\
Pulse-interval SD (ms) & 240~s & $1000\sqrt{(N_I-1)^{-1}\sum_i(I_i-\bar I)^2}$; $N_I\geq120$, interval coverage $\geq120$~s & Adapted from SDNN, Task Force (\citeyear{taskforce1996hrv}); \citet{mejiamejia2020prv} \\
Mean pulse rate (bpm) & 240~s & $60/\operatorname{mean}(I_i)$; $\geq30$ intervals, $\geq60$~s coverage & Peak detector: \citet{elgendi2013peaks} \\
\bottomrule
\end{tabularx}
\end{table}

%% file: main4/tables/rae_ppg_matched_event_ecg_main4.tex
% CONFIRMED RESULT: matched_distribution/summary.json, SDPI rows only.
% Full three-feature audit is preserved in main4/supplement/target_followup_ecg_summary.json.
\begin{table}[!htb]
\topcaption{Matched-event PPG--ECG comparison of pulse-interval SD in 240~s windows. Bias is the mean PPG-minus-ECG difference; brackets show ECG-reference quartiles. P90 denotes the 90th percentile of absolute differences.}
\label{tab:matched-event-ecg}
\centering
\setlength{\parskip}{0pt}
\scriptsize
\setlength{\tabcolsep}{3pt}
\renewcommand{\arraystretch}{1}
\begin{tabularx}{\textwidth}{@{}X>{\centering\arraybackslash}p{0.08\textwidth}>{\centering\arraybackslash}p{0.25\textwidth}>{\centering\arraybackslash}p{0.14\textwidth}>{\centering\arraybackslash}p{0.24\textwidth}@{}}
\toprule
Dataset & Final groups & ECG reference median [Q1--Q3] & Bias (PPG $-$ ECG) & Median / P90 absolute difference \\
\midrule
MIMIC-III & 50 & $23.53\;[12.13, 37.40]$~ms & $+3.23$~ms & $2.39$ / $11.91$~ms \\
VitalDB & 19 & $25.99\;[18.73, 44.14]$~ms & $+2.84$~ms & $1.21$ / $14.74$~ms \\
\bottomrule
\end{tabularx}
\par\smallskip
\noindent\parbox{\textwidth}{\scriptsize Groups are the reference-scorable, matched-comparable subsets of the fixed TRAIN panel and are equally weighted. The ECG reference is an automatic two-detector consensus, not adjudicated normal-to-normal intervals; the comparison describes correspondence conditional on matched events.}
\end{table}

%% file: main4/tables/rae_ppg_target_evidence_main4.tex
% CONFIRMED RESULT: selection pilots (dewt_r1_fx7a_{s,m}_raw_pilot_20260806_v4,
% dewt_r1_fx7a_l_raw_pilot_20260806_v5, dewt_r1_l_four_target_bounded_pilot_20260807_v1,
% dewt_r1_l_ppgv_lf_fraction_20260808_r1), final-extractor repeatability on the M4 panels
% (raeppg2_r1_q1_span_repeatability_20260917_v1; development-stage M4 in
% dewt_r1_m4_scientific_execution_20260808_v1_verification_v1_1) and R1/Q1 held-out
% availability (raeppg2_r1_p2_20260915_v1). Rendered by render_main3_format_tables.py.
\begin{table}[!htb]
\topcaption{Adjacent-window rank consistency and held-out availability of the final targets. Rank-check windows summarize pulse and pulse-pair values by their median and reflected-delay values by their mean; other targets are computed over the indicated window. Each bound is the minimum of the overall and per-dataset lower 95\% bounds of the Spearman--Brown score, shown for the analysis and replication panels at the first tested window length. Held-out availability is the percentage of 87,362 held-out trajectories with an available final-target summary. Dashes indicate auxiliary targets excluded from the rank-consistency analysis.}
\label{tab:target-evidence}
\centering
\fontsize{8}{9}\selectfont
\renewcommand{\arraystretch}{1}
\setlength{\tabcolsep}{4pt}
\begin{tabularx}{\textwidth}{@{}>{\raggedright\arraybackslash}X>{\centering\arraybackslash}p{0.18\textwidth}>{\centering\arraybackslash}p{0.29\textwidth}>{\raggedleft\arraybackslash}p{0.16\textwidth}@{}}
\toprule
Feature & Rank-check window (s) & Lower 95\% bound (analysis / replication) & Held-out availability (\%) \\
\midrule
\multicolumn{4}{@{}l}{\textit{Introduced at 10~s}} \\
Peak-interval rate & 10 & 0.932 / 0.919 & 97.8 \\
Slope transit time & 10 & 0.935 / 0.929 & 97.5 \\
APG $b/a$ & 10 & 0.878 / 0.873 & 96.4 \\
APG $d/a$ & 10 & 0.848 / 0.849 & 88.4 \\
Onset-to-APG-$b$ & 10 & 0.794 / 0.782 & 96.4 \\
Normalized crest time & 10 & 0.878 / 0.867 & 97.5 \\
Normalized PW50 & 10 & 0.918 / 0.920 & 97.5 \\
Split-area ratio & 10 & 0.834 / 0.813 & 95.7 \\
Spectral purity & -- & -- & 99.9 \\
\midrule
\multicolumn{4}{@{}l}{\textit{Introduced at 30~s}} \\
Normalized reflected delay & 20 & 0.900 / 0.895 & 90.6 \\
PRV-RMSSD & 30 & 0.851 / 0.823 & 86.7 \\
\midrule
\multicolumn{4}{@{}l}{\textit{Introduced at 240~s}} \\
PRV-pNN50 & 60 & 0.821 / 0.750 & 97.1 \\
Uncentered ACF index & 60 & 0.914 / 0.881 & 99.9 \\
PPGV LF fraction & 120 & 0.652 / 0.560 & 99.2 \\
Pulse-interval SD & 240 & 0.697 / 0.698 & 93.9 \\
Mean pulse rate & -- & -- & 99.9 \\
\bottomrule
\end{tabularx}
\end{table}

%% file: main4/sections/appendix_experiments_main4.tex
\section{Downstream data and evaluation}
\label{app:cross-duration-evaluation}

Table~\ref{tab:evaluation-ledger} summarizes the evaluation protocols used throughout the paper.

\input{main4/tables/rae_ppg_evaluation_ledger_main4}

\subsection{Datasets, tasks and evaluation cohorts}
\label{app:downstream-details}

We evaluate PPG-DaLiA~\citep{reiss2019dalia}, BUT PPG~\citep{nemcova2021butppg}, version 2.0.0~\citep{nemcova2024butppgdata}, DeepBeat~\citep{torressoto2020deepbeat}, PTT-PPG~\citep{mehrgardt2022pttppg}, WESAD~\citep{schmidt2018wesad}, MESA~\citep{chen2015mesasleep}, SDB~\citep{garde2014sdb}, and DREAMT v2.2.0~\citep{wang2024dreamt,wang2026dreamtdata}.
PulseDB~\citep{wang2023pulsedb} was excluded because it is derived from the MIMIC-III and VitalDB pretraining corpora.
Tables~\ref{tab:downstream-cohorts} and~\ref{tab:downstream-labels} specify the primary evaluation cohorts, input durations, and acquisition and label-construction rules.
In Table~\ref{tab:transfer-all}, DaLiA denotes PPG-DaLiA; HR, SBP/DBP, A/H, and AF denote heart rate, systolic/diastolic blood pressure, apnea/hypopnea, and atrial fibrillation, respectively.

\phantomsection
\noindent\runinheading{Task-duration selection.}
\label{app:task-duration-rationale}
Task durations were fixed using the recording and label context together with the feature families introduced by each duration.
These are evaluation settings, not estimates of the minimum or optimal duration for each task.
Tables~\ref{tab:bridge-evidence-map}--\ref{tab:bridge-relation-matrix} document the source findings and identify the core anchors and contextual relations used in this rationale.

\input{main4/tables/rae_ppg_downstream_cohorts_main4_format}

\input{main4/tables/rae_ppg_downstream_labels_main4}

\subsection{Foundation-model interfaces}
\label{app:foundation-interfaces}

The comparison models were selected for their released encoder checkpoints and reproducible inference interfaces.

All six encoders are frozen. RAE-PPG uses its final checkpoint with the read set matching the input duration and averages all 64 signal tokens to obtain a 384-dimensional representation; pretraining heads are excluded.
Table~\ref{tab:foundation-interfaces} specifies model-specific preprocessing and readouts, and Table~\ref{tab:baseline-artifacts} identifies the comparison checkpoints~\citep{pillai2025papagei,saha2025pulseppg,guo2026sigmappg,nie2025anyppg}.

RAE-PPG processes each evaluation window separately at its source rate using the 12~Hz zero-phase low-pass filter, linear-interpolation resampling to 64~Hz, and masks described in Appendix~\ref{app:targets}.
Edge masks cover 0.5, 1.5, and 2~s at each end of 10~s, 30~s, and 240~s inputs, respectively.
The resulting input is $z$-scored over mask-valid samples as specified in Appendix~\ref{app:local-feature-extractor}.
Invalid normalization statistics produce an all-zero input without dropping the evaluation example.

DeepBeat uses the stored comparison inputs: the 32~Hz segment is upsampled to 64~Hz by polyphase resampling, and its central 10~s is filtered with a fourth-order, zero-phase, 12~Hz Butterworth low-pass filter and min--max scaled to $[0,1]$.
All samples are marked valid before root-wise $z$-scoring, without edge or saturation masks.

\input{main4/tables/rae_ppg_foundation_interfaces_main4_format}

Pulse-PPG uses per-window $z$-scoring because its pretraining participant-normalization statistics are unavailable.
Checkpoints are loaded with strict key matching. The SIGMA-PPG snapshot retains the released student-encoder tensors unchanged, removes the \texttt{student.} prefix, and omits pretraining-only heads and optimizer state.

\input{main4/tables/rae_ppg_baseline_artifacts_main4}

Primary transfer and both cross-duration evaluations use deterministic float32 representation extraction, with TF32 disabled for CUDA matrix multiplication and cuDNN.
Pretraining precision settings are specified separately in Appendix~\ref{app:implementation}.

The released PaPaGei-P/S and AnyPPG checkpoints were pretrained on MESA, but exact participant overlap is unavailable~\citep{pillai2025papagei,nie2025anyppg}.
RAE-PPG pretraining excludes MESA. Appendix~\ref{app:matched-retrain} reports comparison-model retraining on eligible RAE-PPG source recordings.

\subsection{Linear-probe evaluation}
\label{app:linear-evaluation}

Probe selection and fitting use TRAIN only, with VAL+TEST reserved for evaluation, following the split principle of PaPaGei~\citep{pillai2025papagei}.
All crops and repeated trials from a participant or case remain in the same partition.
These fixed partitions are shared across encoders, feature-to-task pathways, and stage-checkpoint comparisons.
Regularization is selected using four grouped inner folds: GroupKFold for regression and StratifiedGroupKFold for classification.
StandardScaler is fitted on each inner-training fold. After selecting regularization by the mean inner-validation score, the scaler and probe are refitted on all TRAIN examples.

\noindent\runinheading{Probe search space.}
Classification uses $L_2$-regularized logistic regression with \texttt{lbfgs} and \texttt{max\_iter=10000}; regression uses ridge regression with an intercept.
Primary transfer and fixed-partition cross-duration evaluation use a float64 SVD ridge solution, whereas LOSO and grouped-OOF evaluation use LSQR.
All encoders use the nine-value grid $\{10^{-4},10^{-3},\ldots,10^4\}$ for $C$ or $\alpha$.
Selection maximizes binary AUROC or multiclass macro-F1, or minimizes regression MAE.

\noindent\runinheading{Metrics and uncertainty.}
Binary classification is evaluated by AUROC, multiclass classification by macro-F1, and regression by MAE.
SDB uses participant-mean window embeddings before probe fitting.
DREAMT age instead fits the probe to window-level embeddings and averages predictions within participants for final scoring.
The 95\% intervals use 1,000 participant/case-cluster bootstrap draws of VAL+TEST, retaining all examples from each sampled group and sharing draws across models within a task.
Encoders and probes remain fixed. The same procedure evaluates the four final ablation checkpoints.
Ranks in Table~\ref{tab:transfer-all} use unrounded task scores.

\subsection{Leave-one-subject-out evaluation}
\label{app:loso-evaluation}

LOSO uses each task's primary input duration. Each outer fold holds out one participant and trains on the other 14 participants for DaLiA/WESAD or 21 for PTT-PPG.
Four grouped inner folds select regularization using the search and scaling procedure in Appendix~\ref{app:linear-evaluation}.
Scores pool held-out predictions from all outer folds. WESAD stress uses its primary cohort of 1,086 windows.
For each task, the comparator is the external model with the best pooled score.
Paired intervals use 1,000 shared participant-bootstrap draws, with the fitted models and observed-best comparator fixed.

\input{main4/tables/rae_ppg_loso_wearable_main4}

RAE-PPG has the better point score on five of seven tasks; the paired interval excludes zero only for DaLiA heart rate, favoring AnyPPG.

\subsection{Rank stability}
\label{app:rank-summary}

For each task, we rank the six models within each of the 1,000 shared participant/case-bootstrap draws.
Table~\ref{tab:bootstrap-ranks} reports mean ranks and first-place counts, averaging ranks equally across the 18 tasks.
RAE-PPG has the lowest bootstrap mean rank and the highest mean first-place count.

\input{main4/tables/rae_ppg_bootstrap_ranks_main4}

\subsection{Robustness to the pretraining seed}
\label{app:seed-robustness}

We performed five additional complete 10~s, 30~s, and 240~s pretraining runs, varying only the random seed governing parameter initialization and sample order.
Data, targets, hyperparameters, and update counts were fixed.
\input{main4/tables/rae_ppg_seed_robustness_main4}

Each final encoder was evaluated using Appendix~\ref{app:linear-evaluation}, with identical examples, splits, and probe grids; baseline scores were fixed to Table~\ref{tab:transfer-all}.
Across all six runs, RAE-PPG outperformed the best baseline on the same ten tasks and underperformed it on the same five; the ordering varied across seeds for the remaining three tasks (Table~\ref{tab:seed-robustness}).

\subsection{Token-count sensitivity}
\label{app:token-count}

We compared $K=64,128,256$ tokens, with one complete 10~s, 30~s, and 240~s pretraining run per setting.
Data, targets, initialization seed, objective, parameter count, optimizer, and update budgets were fixed, without hyperparameter retuning.
Final encoders were evaluated using Appendix~\ref{app:linear-evaluation} against the same five baseline checkpoints (Table~\ref{tab:token-intervals}).
Increasing $K$ did not improve the aggregate ranking in these runs, and $K=64$ remains the default.

\begin{table}[H]
\centering
\caption{Token-count sensitivity on the 18 primary tasks. Each setting uses one complete pretraining run and the same five fixed comparison checkpoints.}
\label{tab:token-intervals}
\small
\begin{tabular}{rrr}
\toprule
Tokens $K$ & Best-score tasks /18 & Mean rank $\downarrow$ \\
\midrule
64 & 12 & 1.44 \\
128 & 12 & 1.44 \\
256 & 12 & 1.44 \\
\bottomrule
\end{tabular}
\end{table}

\subsection{Fixed-partition cross-duration evaluation}
\label{app:cross-duration-scores}

The 13-task analysis uses fixed TRAIN/VAL+TEST participant partitions and matched labels and window centers at 10~s, 30~s, and 240~s.
Separate probes are fitted for each task and duration using Appendix~\ref{app:linear-evaluation}; all durations use the same final frozen encoder with duration-specific read sets.
WESAD stress uses the 107 windows with complete 240~s support: 64 TRAIN windows from nine participants and 43 evaluation windows from six participants, rather than the primary cohort of 1,086 windows.
Reference durations are fixed per task; MESA sleep staging uses a 30~s reference and the central 30~s label.

\input{main4/tables/rae_ppg_cross_duration_ci_main4}

\subsection{Retention-margin sensitivity}
\label{app:retention-margin-sensitivity}

The primary operational Retention margin $\delta$ is 0.02 for AUROC/macro-F1, 0.190~bpm for DaLiA heart rate, and 0.257~years for DREAMT age.
We scale each task's margin by $0.5$, $1$, $1.5$, or $2.5$, using the same 1,000 paired draws and a separate Holm-8 correction at each setting.
Retention is confirmed on one, two, three, and three of the eight applicable tasks, respectively.
MESA sleep staging passes at every setting, DREAMT sleep staging from $\delta$, and DREAMT age from $1.5\delta$.
Extend is unchanged at six of eleven because its criterion does not use this margin.

\subsection{Grouped out-of-fold cross-duration evaluation}
\label{app:oof-cross-duration}

We evaluated the final frozen RAE-PPG encoder on the same 13 matched-duration cohorts as in Appendix~\ref{app:cross-duration-scores}, using five participant-grouped outer folds and four inner folds.
Participants, labels, window centers, and folds were matched across input durations.
Scaling and probe selection follow the fold-local procedure in Appendix~\ref{app:linear-evaluation}.
Scores pool the outer-fold predictions so that every participant contributes held-out predictions; DREAMT age is scored after averaging predictions within participants.
Bootstrap resampling, reference durations, operational margins, and Retention/Extend criteria follow Appendix~\ref{app:cross-duration-scores}.
Table~\ref{tab:oof-cross-duration-ci} reports the RAE-PPG results.

\input{main4/tables/rae_ppg_cross_duration_oof_supp_main4}

% Output every Appendix B table before Appendix C begins.
\FloatBarrier

\section{Signal-feature analyses}
\label{app:physiological-evidence}

\subsection{Signal-feature retention and extension}
\label{app:physiological-protocol}

Feature decoding uses 87,362 aligned trajectories from 1,308 participants or cases held out from pretraining, without downstream labels.
All input durations use the same final frozen encoder.
A separate ridge probe is fitted for each feature and representation view using five participant/case-grouped outer folds and four inner folds.
Inner validation selects $\alpha$ from $\{10^{-1},10^0,\ldots,10^6\}$ by Huber error.
Standardization and probe fitting are performed within each training fold, with equal weight per participant or case.

\noindent\runinheading{Targets and readouts.}
Each decoding target averages available native feature occurrences within the centered 10~s, 30~s, or 240~s interval for its introduction stage, after pretraining-TRAIN normalization (Appendix~\ref{app:loss}).
For each feature, all compared views use the same trajectories with available targets.
Feature retention decodes earlier features from longer inputs using overlap-weighted token readouts aligned to the earlier interval.
Feature extension predicts the same later-stage targets from the means of all signal tokens in shorter and longer inputs.
In the shorter-input condition, the target is not recomputed over a shorter window.
Table~\ref{tab:physiology-all} specifies the input durations and readouts.

\noindent\runinheading{Decoding gain.}
Outer-fold predictions are pooled.
For each feature $f$, the probe error $E_{\mathrm{probe},f}$ and constant-zero prediction error $E_{0,f}$ use unit-threshold Huber loss in normalized target coordinates, with the same valid targets in both errors.
Each error is averaged within participants/cases, then equally across available groups within each dataset, and finally equally across MIMIC-III and VitalDB.
Gain is defined as $\mathrm{gain}_f=1-E_{\mathrm{probe},f}/E_{0,f}$; positive gain indicates improvement over the zero predictor.
Table~\ref{tab:physiological-evidence}A averages gains equally across the relevant features.
Feature retention reports gain at the longer input duration, while feature extension reports the longer-minus-shorter gain difference.

\noindent\runinheading{Constant-predictor controls.}
Positive gain over zero alone does not exclude improvement from a learned intercept, because pretraining normalization need not center the window-level decoding targets.
We therefore compared the saved probe predictions with three controls fitted separately for each feature using only the corresponding outer training fold: a participant/case-weighted mean, a constant minimizing the same dataset- and group-balanced Huber error used for evaluation, and separate Huber-minimizing constants for each dataset.
The mean matches an intercept-only ridge predictor; the latter two controls minimize Huber error without using representations.
All controls use the original target masks, outer folds, evaluation weights, and 5,000 bootstrap draws, with fitted predictors fixed.
Across all 45 feature/readout combinations, probe error was lower than under each of the three controls; all 135 paired error-reduction comparisons had positive individual 95\% confidence-interval lower bounds.
Against the pooled Huber constant, mean gains for retained 10~s features at 30~s and retained 30~s features at 240~s were 0.812 [0.795, 0.828] and 0.663 [0.642, 0.685], respectively.
The original zero-reference gains above remain unchanged.
These comparisons support feature information beyond a learned constant or dataset identity alone.

\noindent\runinheading{Uncertainty and tests.}
Percentile 95\% intervals use 5,000 participant/case bootstrap draws stratified by dataset and shared across features and views, with fitted probes fixed.
Each draw recomputes the group- and dataset-balanced errors and feature-average gains over available targets.
Four prespecified one-sided tests assess accessibility of 10~s features at 30~s, extension of 30~s features from 10~s to 30~s inputs, accessibility of 30~s features at 240~s, and extension of 240~s features from 30~s to 240~s inputs.
Null-centered component tests use Holm correction~\citep{holm1979multiple}.
The intersection--union test takes the largest raw component $p$-value; all four components are positive ($p_{\mathrm{IUT}}<.001$).
Benjamini--Hochberg correction is applied across the 18 secondary feature-level tests~\citep{benjamini1995fdr}.

\noindent\runinheading{Paired decoding comparisons.}
Relative to the corresponding introduction-duration readouts, mean gain differences were $-0.008$ [$-0.013$, $-0.004$] for 10~s features at 30~s, $-0.194$ [$-0.205$, $-0.183$] for 10~s features at 240~s, and $-0.232$ [$-0.253$, $-0.211$] for 30~s features at 240~s.
Brackets denote paired 95\% intervals.
Both features introduced at 30~s and four of the five features introduced at 240~s showed positive extension after correction across the 18 secondary tests.
The PRV-pNN50 extension difference was $-0.007$ [$-0.035$, $0.018$].

\input{main4/tables/rae_ppg_physiology_all_main4}

\subsection{Downstream prediction from signal features}
\label{app:bridge-protocol}

The Feature-to-Task Bridge tests downstream prediction using either features computed directly from PPG or the same features decoded from the frozen encoder.

\noindent\runinheading{Task--feature relations.}
The fixed analysis set contains 33 task--family relations: 22 core anchors used in the task-duration rationale and 11 additional relations. Core-anchor status denotes inclusion in the prespecified task-duration rationale, not the strength or statistical significance of the reported source finding.
At each task's input duration, only features introduced by that duration are eligible; a family with no eligible features is marked ineligible.
Table~\ref{tab:bridge-relation-matrix}A reports all 90 task--family combinations.
Under the original source classification, 71 are eligible: the 33 main relations, 18 contextual relations, and 20 unmapped relations.
The remaining 19 are duration-ineligible.
Table~\ref{tab:bridge-evidence-map} records the source measurements and findings for the main relations; core anchors are marked by an asterisk.

Source labels distinguish related measurements (R), indirect support (I), general task or waveform context (C), and relations not mapped by the cited sources (U).
The labels were assigned without using the present Bridge outcomes.
Alternative classifications from the subsequent source re-audit are examined below.

\noindent\runinheading{Contextual sources.}
Signal-quality context comes from a relative-power SQI evaluated on 60~s recordings; it differs from spectral purity and did not separate quality classes after correction~\citep{elgendi2016sqi}.
Other contextual sources concern end-to-end sleep staging from raw PPG~\citep{korkalainen2020sleep}, arousal-related PPG amplitude and spectral-mean-frequency responses~\citep{luukinen2024arousal}, and cyclic pulse-interval variation~\citep{hayano2020apnea}.
\citet{hayano2020apnea} reported an AHI association for the hourly frequency of detected cyclic variation, but not for conventional variability indices.
These sources provide task or event context rather than tests of the present waveform-statistics targets or 240~s burden endpoints.

\noindent\runinheading{Feature construction.}
The Bridge uses the sixteen target definitions in Appendix~\ref{app:targets}.
Each of the 109,929 windows used in this analysis is linked to its source recording.
Features are extracted from the finite source block containing the window, using the pretraining filters, resampling, extraction kernels, quality gates, and TRAIN-fitted target normalizers.
The 30~s minimum-block requirement applies only to the pretraining corpora: the Bridge uses each complete 10~s BUT PPG record and 25~s DeepBeat segment, with DeepBeat targets evaluated on the central 10~s crop.
Processing respects the available record boundaries, without concatenating or extending records.
Available normalized occurrence values are averaged within each input to form one scalar per eligible feature; SDB values are further averaged within participants.
Encoder inputs follow Appendix~\ref{app:foundation-interfaces}.

\noindent\runinheading{Probe fitting.}
Both pathways use the primary-transfer participants, examples, labels, and metrics.
Task-probe selection follows Appendix~\ref{app:linear-evaluation}.
The decoded pathway fits feature probes with $\alpha=1$ using feature targets, without downstream labels.
Within each grouped task-selection split, cross-fitting within the training subset generates decoded features for task-probe training.
Feature probes fitted on that entire training subset generate the task-validation features.
After selection, the final task probe is fitted on cross-fitted TRAIN features, and feature probes fitted on all TRAIN data provide the VAL+TEST features.

Within each fitting subset, a feature requires at least eight valid rows from at least three participant groups and positive variance.
Both pathways and all controls use the same eligible coordinates.
Scaling and mean imputation are fitted only on the corresponding fitting subset.
Evaluation rows are retained, and validity indicators are not supplied as predictors.

\noindent\runinheading{Participant-reassignment controls.}
Each pathway has 32 controls.
Within each fitting or prediction partition, feature-family vectors and their availability masks are jointly reassigned to different participants, independently of the labels.
Donor rows are reused in shuffled complete cycles followed by a remainder.
Decoded controls fit feature probes to the reassigned targets, and each control fits its own task probe.
This preserves row coverage, but participant-level row-count imbalance can cause donor rows to be reused or omitted and therefore can change the empirical feature distribution.

\noindent\runinheading{Control sensitivity.}
We audited donor-row reuse and omission for the fixed 33-relation analysis.
For 19 relations, every fitting and prediction partition already used each donor row exactly once, preserving the complete set of feature-family vectors and availability masks.
For the remaining 14 relations, we additionally tested a without-replacement, no-self reassignment whenever such a mapping was feasible in every partition.
This was feasible for four relations, and their joint Raw/Decoded decisions were unchanged, including the DeepBeat AF morphology/vascular-timing relation, which showed the largest Kolmogorov--Smirnov distance in the final evaluation partitions under the original control.
One individual pathway, decoded features for WESAD valence, changed from negative to positive without changing its joint relation decision.
For the other ten relations, at least one partition contained a participant with more rows than all other participants combined, making the same one-to-one, no-self reassignment infeasible without changing the partition or donor constraint.
This analysis is therefore a partial sensitivity test of donor-row reuse and omission rather than a replacement control for all 33 relations.

\noindent\runinheading{Inference.}
For each relation and pathway, the effect is observed task utility minus the median utility of its 32 controls.
Utility is $-\mathrm{MAE}$, AUROC, or macro-F1, as appropriate.
We use 1,000 paired participant/case bootstrap draws, shared across pathways and families within a task; each draw recomputes the control median while fitted models remain fixed.
Age uses participant-averaged predictions, and macro-F1 uses the fixed class set.
Binary-task draws containing only one class are excluded, with at least 500 valid draws required.
A pathway is positive for a relation when the one-sided 95\% lower bound of its effect exceeds zero.
These relation-wise comparisons are unadjusted.
Both denotes positivity in each pathway.
Table~\ref{tab:physiological-evidence}B summarizes the fixed relation set, and Table~\ref{tab:bridge-relation-matrix}A reports all task--family combinations.

\input{main4/tables/rae_ppg_bridge_evidence_map_main4}

\noindent\runinheading{Sensitivity to source interpretation.}
The fixed relation set and original labels predate the source re-audit.
In an alternative classification, the 12 marked cells in Table~\ref{tab:bridge-relation-matrix}A change as follows: seven R to I, two I to C, and three C to U.
The first two changes reflect differences in source measurements, endpoints, or reported findings; the C-to-U changes remove contextual mappings without a corresponding family measurement in the cited sources.
The resulting R/I subset contains 31 relations.
A separate post-audit check excludes the MESA and DREAMT apnea/hypopnea--BIV relations, whose cited source was revised during the re-audit.
Table~\ref{tab:bridge-relation-matrix}B reports these subsets and their combination using the same predictions and positivity decisions; the fixed 33-relation set remains the main analysis.

\input{main4/tables/rae_ppg_bridge_relation_matrix_main4}

% Output every Appendix C table before Appendix D begins.
\FloatBarrier

\section{Ablations of the retain-and-extend design}
\label{app:mechanistic-protocol}

\subsection{Matched conditions and evaluation}
\label{app:ablation-conditions}

The four-arm ablation compares nested-read/protected-write and fully shared encoders, each with retained-target supervision ON or OFF. A (RAE-PPG) and D use the nested/protected organization with supervision ON and OFF, respectively; B and C use the fully shared organization with supervision ON and OFF. Fully shared encoders read and update all groups at every stage, so the organization contrast changes read access, update protection, and active capacity together. Data, native targets and their weights, initialization, stage order, optimization schedules, and update budgets are matched. Disabling retained-target supervision removes the retained-loss terms without rescaling the native losses (Appendices~\ref{app:loss}--\ref{app:schedule}).

\subsection{Final transfer and preservation during training}
\label{app:ablation-transfer}

\input{main4/tables/rae_ppg_ablation_terminal_main4}

The four final encoders are evaluated on the 18 downstream tasks using the fixed TRAIN/\allowbreak VAL+\allowbreak TEST partitions and linear-probe procedure in Appendix~\ref{app:linear-evaluation}. Table~\ref{tab:ablation-terminal} reports all scores and confidence intervals; Table~\ref{tab:ablation-design-main} summarizes mean ranks.

\noindent\runinheading{Shorter-input representations and predictions.}
Let S, M, and L denote checkpoints after the 10~s, 30~s, and 240~s pretraining stages. We keep the input examples, participants, labels, and inner folds fixed while comparing these checkpoints. Seven tasks use 10~s inputs for S$\to$M and S$\to$L comparisons, and five tasks use 30~s inputs for M$\to$L, giving 19 comparisons across 12 tasks. Within each of A and D, representations and predictions are identical across all 19 comparisons. Both fully shared conditions have seven positive and twelve negative changes in downstream utility (Table~\ref{tab:stage-preservation}).

\input{main4/tables/rae_ppg_stage_preservation_main4}

\noindent\runinheading{Earlier-feature decoding across stages.}
We use the separate held-out trajectory panel and grouped feature-probe evaluation in Appendix~\ref{app:physiological-protocol}. The nine features introduced at 10~s are decoded from the same 10~s inputs at S, M, and L; the two features introduced at 30~s are decoded from the same 30~s inputs at M and L. Input examples, normalized targets, validity masks, aligned readouts, and grouped folds are matched across checkpoints and conditions. We use the existing, independently fitted feature probes for each checkpoint. Probe fitting, regularization selection, and feature-equal gain aggregation follow Appendix~\ref{app:physiological-protocol}. We report $\Delta G=G_{\mathrm{later}}-G_{\mathrm{earlier}}$, where $G$ is the mean gain over the corresponding feature set. Paired changes and supervision contrasts are computed using 5,000 shared, dataset-stratified participant/case bootstrap draws with the checkpoints, probes, and folds fixed. Table~\ref{tab:fixed-input-preservation} reports unadjusted 95\% percentile intervals.

Within each protected condition, the corresponding $87{,}362\times384$ readout matrices are bit-identical across the compared checkpoints, giving zero change in gain. For the three introduction-to-later comparisons, every feature-wise interval in both fully shared conditions is below zero. Retained-target supervision reduces the cumulative gain loss relative to the introduction checkpoint. Table~\ref{tab:fixed-input-preservation} also reports the secondary M$\to$L comparison for the 10~s features.

\input{main4/tables/rae_ppg_fixed_input_preservation_main4}

% Start D.3 after the preceding table continuations have been placed at the page top.
\newpage

\subsection{Feature decodability after pretraining}
\label{app:fourarm-feature-accessibility}

We compare the four final checkpoints on the held-out trajectories, feature targets, validity masks, and grouped folds used in Appendix~\ref{app:physiological-protocol}. Earlier features are decoded from longer inputs using readouts aligned to the centered earlier interval; newly introduced features use the mean of all 64 tokens at their introduction duration. Feature probes are fitted separately for each condition. For RAE-PPG, the representation exports and probe fits are separate from those used for Table~\ref{tab:physiological-evidence}A. Feature-probe selection, target normalization, and gain aggregation follow Appendix~\ref{app:physiological-protocol}. The same 5,000 dataset-stratified participant/case bootstrap draws provide absolute-gain intervals and paired contrasts, with the fitted checkpoints, probes, and folds fixed. Ablation contrasts are unadjusted.

\input{main4/tables/rae_ppg_ablation_features_main4_format}

Retained-target supervision increases the point estimates for all nine 10~s features decoded at 30~s and both 30~s features decoded at 240~s in the nested/protected encoder. The corresponding counts in the fully shared encoder are seven of nine and two of two. Table~\ref{tab:ablation-feature-accessibility}A reports absolute gains, and Table~\ref{tab:ablation-feature-accessibility}B reports paired supervision and organization contrasts.

\input{main4/tables/rae_ppg_ablation_paired_main4}

\subsection{Prior-stage reuse}
\label{app:shorter-group-reuse}

\noindent\runinheading{Intervention and stage initialization.}
We test reuse of the inherited pathway as a whole using a separate matched pair of continuations at each transition. The 30~s pair starts from the 10~s parent checkpoint and runs for 340,309 updates; the 240~s pair starts independently from the 30~s parent checkpoint and runs for 125,551 updates. LEARNED retains all previously learned parameters used in the next stage's forward pass. FULL-PRIOR INITIAL restores this entire pathway to values from a fixed untrained reference model, including the shared local feature extractor, token projection and normalization layers, and earlier attention/FFN groups (Appendix~\ref{app:allocation}). In both conditions, these parameters remain active but frozen, and retained-target supervision is disabled. Parameters and prediction heads assigned to the next stage have identical initial values in the two conditions. Data membership, sample order, new targets, normalizers, optimizer ownership, and schedules are also matched.

\noindent\runinheading{Independent feature probes.}
Evaluation uses the pretraining VAL panel: 4,508 recordings from 1,020 participants/cases for the 30~s continuation and 4,359 recordings from 1,022 groups for the 240~s continuation, across MIMIC-III and VitalDB. Fresh ridge probes use the mean of all 64 output tokens to predict each newly introduced feature. Each target is the average of the available native feature values after TRAIN normalization. Feature-specific validity masks and five outer/four inner participant/case-grouped folds are matched across conditions. Scaling and probe fitting give each recording equal weight. Regularization $\alpha$ is selected from $\{10^{-2},10^{-1},\ldots,10^4\}$ using validation Huber error balanced across participants/cases and datasets.

\noindent\runinheading{Aggregation and uncertainty.}
For each feature and condition, probe Huber error is averaged within participants/cases, then equally across groups within each dataset, and finally equally across MIMIC-III and VitalDB.
Dividing by the correspondingly aggregated zero-prediction error gives normalized error, and subtracting it from one gives feature gain.
Let $G$ be the equal-weight mean gain over the two features introduced at 30~s or the five introduced at 240~s.
With INITIAL denoting FULL-PRIOR INITIAL, we report $\Delta G=G_{\mathrm{LEARNED}}-G_{\mathrm{INITIAL}}$ and
\[
\mathrm{Error\ reduction}\,(\%)
=100\frac{G_{\mathrm{LEARNED}}-G_{\mathrm{INITIAL}}}{1-G_{\mathrm{INITIAL}}}.
\]
Here, $1-G$ is the feature-average error after normalization by each feature's zero-prediction error.
Thus, the percentage measures how much of the INITIAL condition's remaining normalized error is removed by LEARNED.
Paired 95\% intervals use 5,000 shared, dataset-stratified participant/case bootstrap draws, with the continuation checkpoints, fitted probes, and folds fixed.
Percentage intervals require at least 1,000 finite draws with $1-G_{\mathrm{INITIAL}}>0$.
Results are reported without multiplicity adjustment (Table~\ref{tab:shorter-group-reuse}).

LEARNED reduces normalized probe error at both transitions, with all seven feature-level intervals favoring reuse.

\input{main4/tables/rae_ppg_shorter_group_reuse_main4}

%% file: main4/tables/rae_ppg_evaluation_ledger_main4.tex
% Local H-table spacing: keep its native 12pt outer gap; avoid adding a second caption-top gap.
\begingroup
\setlength{\intextsep}{12pt}
\begin{table}[H]
\setlength{\abovecaptionskip}{0pt}
\topcaption{Evaluation protocols. Folds and bootstrap resampling group participants or cases, and paired comparisons use shared draws. Task-probe search grids and selection metrics are specified in Appendix~\ref{app:linear-evaluation}. Intervals condition on the fitted encoders and probes.}
\label{tab:evaluation-ledger}
\centering\fontsize{8}{9}\selectfont
\setlength{\tabcolsep}{4pt}\renewcommand{\arraystretch}{1}
\begin{tabularx}{\linewidth}{@{}>{\raggedright\arraybackslash}p{.22\linewidth}>{\raggedright\arraybackslash}p{.40\linewidth}>{\raggedright\arraybackslash}X@{}}
\toprule
Analysis & Fitting and evaluation & Draws / inference \\
\midrule
Primary transfer; terminal ablations & Four inner TRAIN folds; refit TRAIN, evaluate VAL+TEST & 1,000; score CIs and descriptive ranks \\
Cross-duration transfer & Same TRAIN/VAL+TEST roles and window centers & 1,000; Holm-8 Retention, Holm-11 Extend \\
RAE-PPG cross-duration OOF & Five outer / four inner folds & Same inference as fixed-partition cross-duration evaluation \\
LOSO & One held-out participant; four inner folds & 1,000; unadjusted contrasts \\
Feature decoding; feature ablations & Five outer / four inner folds; group-weighted fits; Huber selection, $\alpha\in\{10^{-1},\ldots,10^6\}$ & 5,000, dataset-stratified; retention/extension: Holm-4, BH-18; ablations: unadjusted \\
Feature-to-task Bridge & Nested TRAIN feature-probe cross-fitting ($\alpha=1$) and task-probe selection; VAL+TEST evaluation & 1,000; 32 reassignment controls; relation-wise lower bounds \\
Downstream stage preservation & Fixed inputs, participants and probe folds across checkpoints & 1,000; unadjusted contrasts \\
Prior-stage reuse & VAL panel; five outer / four inner folds; row-weighted fits; balanced Huber selection, $\alpha\in\{10^{-2},\ldots,10^4\}$ & 5,000, dataset-stratified; unadjusted contrasts \\
\bottomrule
\end{tabularx}
\end{table}
\endgroup

%% file: main4/tables/rae_ppg_downstream_cohorts_main4_format.tex
% CONFIRMED RESULT: TRAIN-only result plan; cohorts are model-independent and
% unchanged from main3 (PTT uses the saved split-21 counts).
\begin{table}[H]
\topcaption{Primary downstream tasks and evaluation cohorts. Counts are eligible examples (participants); final evaluation uses VAL+TEST.}
\label{tab:downstream-cohorts}
\begingroup
\centering
\scriptsize
\setlength{\tabcolsep}{3pt}
\renewcommand{\arraystretch}{1}
\begin{tabularx}{\textwidth}{@{}lXrlrr@{}}
\toprule
Dataset & Task(s) & Input (s) & Metric & TRAIN & Final evaluation \\
\midrule
PPG-DaLiA & Heart rate & 10 & MAE & 316 (9) & 216 (6) \\
 & Activity & 30 & macro-F1 & 311 (9) & 207 (6) \\
BUT PPG & Heart rate & 10 & MAE & 2,492 (30) & 1,396 (20) \\
DeepBeat & Atrial fibrillation & 10 & AUROC & 7,990 (119) & 21,617 (30) \\
 & Signal quality & 10 & macro-F1 & 7,990 (119) & 21,617 (30) \\
PTT-PPG & SBP; DBP & 10 & MAE & 84 (14) & 48 (8) \\
WESAD & Stress & 30 & AUROC & 651 (9) & 435 (6) \\
 & Arousal; valence & 30 & AUROC & 82 (9) & 55 (6) \\
MESA & Sleep stage & 240 & macro-F1 & 29,616 (1,234) & 19,728 (822) \\
 & Apnea/hypopnea burden & 240 & MAE & 29,616 (1,234) & 19,728 (822) \\
 & Arousal burden & 240 & MAE & 29,616 (1,234) & 19,728 (822) \\
 & SpO$_2$ desaturation & 240 & MAE & 29,616 (1,234) & 19,728 (822) \\
SDB & AHI-code classification & 240 & AUROC & 87 (87) & 59 (59) \\
DREAMT & Age & 10 & MAE & 1,416 (59) & 960 (40) \\
 & Sleep stage & 30 & macro-F1 & 1,416 (59) & 960 (40) \\
 & Apnea/hypopnea & 240 & AUROC & 1,416 (59) & 960 (40) \\
\bottomrule
\end{tabularx}
\par\smallskip
\noindent\parbox{\textwidth}{\scriptsize
Examples are input windows except for SDB, which uses one pooled example per participant. DREAMT age is scored after participant-level prediction averaging. Tasks can share participants and windows, so their counts should not be summed.}
\par\endgroup

%% file: main4/tables/rae_ppg_downstream_labels_main4.tex
\topcaption{Acquisition, window selection, and labels. Sampling rates refer to source signals before the model-specific processing in Appendix~\ref{app:foundation-interfaces}; cohort sizes and partitions are given in Table~\ref{tab:downstream-cohorts}.}
\label{tab:downstream-labels}
\centering\fontsize{8}{9}\selectfont
\setlength{\tabcolsep}{4pt}\renewcommand{\arraystretch}{1}
\begin{tabularx}{\linewidth}{@{}>{\raggedright\arraybackslash}p{.12\linewidth}>{\raggedright\arraybackslash}p{.21\linewidth}>{\raggedright\arraybackslash}X@{}}
\toprule
Dataset & Acquisition / channel & Window selection and labels \\
\midrule
DaLiA & Daily activities; wrist BVP, 64~Hz & Centered 10~s/30~s crops from finite, non-flat, non-overlapping 240~s blocks. HR is the central official 8~s ECG annotation on its 2~s grid. Activity is the modal label within the 30~s crop, with at least 80\% label purity; transition code 0 is retained.\\

BUT PPG & Smartphone PPG (Xiaomi Mi9, Huawei P20 Pro), finger or ear, 30~Hz; v2.0.0 & All 3,888 released 10~s records from 50 participants: 1,926 finger records (30 participants) and 1,962 ear records (20 participants). The red channel is used in RGB files; 48 legacy files use the flattened 300-sample layout. WFDB samples retain the declared regular grid. HR comes from the record-level ECG reference; both quality categories are retained.\\
DeepBeat & Released wearable PPG, 32~Hz & The central 10~s is cropped from each released, min--max-scaled 25~s segment. AF labels are binary; quality labels are noise, acceptable, or excellent. Validation records from the eight participants also present in TEST are removed. TRAIN and validation are randomly capped at 8,000 and 4,000 segments without using labels; TEST is not subsampled. Non-finite or flat crops are excluded.\\
PTT-PPG & Exercise recordings; \texttt{pleth\_2}, 500~Hz & First and last 10~s of each recording are paired with its start/end cuff SBP and DBP. Repeated trials remain with the same participant.\\
WESAD & Laboratory protocol; wrist BVP, 64~Hz & Stress is TSST versus baseline/amusement; meditation is excluded. Stress inputs are contiguous, non-overlapping 30~s crops inside one condition. Arousal/valence use the central 30~s of non-overlapping 240~s windows from all five conditions, including meditation, and inherit that condition's SAM rating (1--9), with class 1 for ratings $\le5$ and class 0 for ratings $>5$.\\
MESA & Overnight PSG; Pleth, 256~Hz & 24 eligible, evenly spaced 240~s windows per participant, centered on scored 30~s epochs. Stages map to wake/light/deep/REM (N1+N2 light; N3+N4 deep). Each event burden is the duration of the union of annotated intervals clipped to the 240~s window, divided by 240~s, giving a value in $[0,1]$.\\
SDB & Pediatric overnight phone-oximeter recordings, 62.5~Hz & Up to 128 evenly spaced finite, non-flat, non-overlapping 240~s windows per participant. Participant-level classification of the released AHI codes: 0 versus 1--3 (93 versus 53 children). The release does not specify their clinical AHI cutoffs. Window embeddings are averaged within participants before probing.\\
DREAMT & Wrist BVP aligned with sleep annotations, 64~Hz; v2.2.0 & 24 evenly spaced valid 240~s parent windows per participant. Sleep uses the central scored 30~s epoch (W/N1/N2/N3/REM). A/H presence is any positive overlap with obstructive apnea, central apnea or hypopnea. Age predictions are averaged within participants.\\
\bottomrule
\end{tabularx}
\end{table}

%% file: main4/tables/rae_ppg_foundation_interfaces_main4_format.tex
% CONFIRMED RESULT: released loaders, comparator adapters and transfer18 plan dimensions.
\begin{table}[!htb]
\topcaption{Model-specific interfaces for downstream evaluation. Sampling rates and dimensions refer to model inputs and probe representations. BP and LP denote band-pass and low-pass filtering. The DeepBeat-specific RAE-PPG preprocessing is described in Appendix~\ref{app:foundation-interfaces}.}
\label{tab:foundation-interfaces}
\centering
\scriptsize
\setlength{\tabcolsep}{2pt}
\renewcommand{\arraystretch}{1}
\begin{tabularx}{\textwidth}{@{}>{\raggedright\arraybackslash}p{48pt}p{24pt}>{\raggedright\arraybackslash}X>{\raggedright\arraybackslash}p{49pt}>{\raggedright\arraybackslash}p{85pt}r@{}}
\toprule
Model & $f_s$ (Hz) & Preprocessing used & Input handling & Representation & Dim. \\
\midrule
PaPaGei-P & 125 & z-score; 0.5--12 Hz BP; resample & 10~s tiles & 512-D embedding; tile mean + SD & 512/1,024 \\
PaPaGei-S & 125 & z-score; 0.5--12 Hz BP; resample & 10~s tiles & 512-D embedding; tile mean + SD & 512/1,024 \\
Pulse-PPG & 50 & Window z-score; no filter; resample & Full variable-length input & Global max & 512 \\
SIGMA-PPG & 50 & 0.5--8 Hz BP; resample; min--max $[-1,1]$ & 2~s patches; native positions & Mean of non-class patch tokens; final normalization & 360 \\
AnyPPG & 125 & 0.5--8 Hz BP; resample; window z-score & Full variable-length input & Internal global mean & 512 \\
RAE-PPG & 64 & 12 Hz LP; resample; masked root z-score & 10/30/240~s; 64 tokens & Signal-token mean & 384 \\
\bottomrule
\end{tabularx}
\par\smallskip
\noindent\parbox{\textwidth}{\scriptsize PaPaGei uses a 512-dimensional embedding for 10~s inputs. For longer inputs, the mean and population standard deviation of embeddings from complete, non-overlapping 10~s tiles are concatenated, yielding 1,024 dimensions.}
\end{table}

%% file: main4/tables/rae_ppg_baseline_artifacts_main4.tex
\begin{table}[!htb]
\topcaption{Comparison checkpoint files, source revisions and SHA-256
prefixes.}
\label{tab:baseline-artifacts}
\centering\fontsize{8}{9}\selectfont
\setlength{\tabcolsep}{4pt}\renewcommand{\arraystretch}{1}
\begin{tabular}{@{}llll@{}}
\toprule
Model & Checkpoint file & Source commit & SHA-256 prefix \\
\midrule
PaPaGei-P & \texttt{papagei\_p.pt} & \texttt{0c537da} & \texttt{3a6850961af5} \\
PaPaGei-S & \texttt{papagei\_s.pt} & \texttt{0c537da} & \texttt{79d68671e51b} \\
Pulse-PPG & \texttt{checkpoint\_best.pkl} & \texttt{716eaf9} & \texttt{485ade5033b3} \\
SIGMA-PPG & \shortstack[l]{\texttt{sigma\_student\_strict\_}\\\texttt{b185072e.pt}} & \texttt{b1cc8d9} & \texttt{3fe728902855} \\
AnyPPG & \texttt{anyppg\_ckpt.pth} & \texttt{661b877} & \texttt{99b9bb0a3c2b} \\
\bottomrule
\end{tabular}
\end{table}

%% file: main4/tables/rae_ppg_loso_wearable_main4.tex
% CONFIRMED RESULT: raeppg2_r1_primary_loso_20260917_v1/result.json.
% Local H-table spacing: keep its native 12pt outer gap; avoid adding a second caption-top gap.
\begingroup
\setlength{\intextsep}{12pt}
\begin{table}[H]
\setlength{\abovecaptionskip}{0pt}
\topcaption{Primary-duration LOSO evaluation. $\Delta U$ is RAE-PPG minus the observed-best external model, with utility defined as $-\mathrm{MAE}$, AUROC, or macro-F1. Brackets give paired, unadjusted 95\% participant-bootstrap intervals. Bold marks the better point score.}
\label{tab:loso-wearable-best}
\centering
\fontsize{8}{9}\selectfont
\setlength{\tabcolsep}{3pt}
\renewcommand{\arraystretch}{1}
\begin{tabularx}{\textwidth}{@{}>{\raggedright\arraybackslash}p{78pt}>{\centering\arraybackslash}p{41pt}>{\centering\arraybackslash}p{85pt}>{\centering\arraybackslash}X@{}}
\toprule
Task / duration & RAE-PPG & Best external model & $\Delta U$ [95\% CI] \\
\midrule
DaLiA heart rate\newline{\scriptsize 10~s; MAE $\downarrow$} & 8.6138 & AnyPPG\newline \textbf{7.3293} & $-1.2844$ [-2.0133, -0.4897] \\
DaLiA activity\newline{\scriptsize 30~s; macro-F1 $\uparrow$} & \textbf{0.3937} & AnyPPG\newline 0.3397 & $+0.0540$ [-0.0021, +0.1194] \\
WESAD stress\newline{\scriptsize 30~s; AUROC $\uparrow$} & \textbf{0.9484} & AnyPPG\newline 0.9407 & $+0.0077$ [-0.0080, +0.0242] \\
WESAD arousal\newline{\scriptsize 30~s; AUROC $\uparrow$} & \textbf{0.8016} & AnyPPG\newline 0.7781 & $+0.0235$ [-0.1299, +0.1607] \\
WESAD valence\newline{\scriptsize 30~s; AUROC $\uparrow$} & \textbf{0.7095} & PaPaGei-S\newline 0.6549 & $+0.0546$ [-0.0998, +0.2477] \\
PTT SBP\newline{\scriptsize 10~s; MAE $\downarrow$} & \textbf{9.4724} & PaPaGei-P\newline 9.4932 & $+0.0207$ [-0.6675, +0.6648] \\
PTT DBP\newline{\scriptsize 10~s; MAE $\downarrow$} & 7.2528 & Pulse-PPG\newline \textbf{6.9416} & $-0.3112$ [-0.9697, +0.2514] \\
\bottomrule
\end{tabularx}
\end{table}
\endgroup

%% file: main4/tables/rae_ppg_bootstrap_ranks_main4.tex
% CONFIRMED RESULT: raeppg2_r1_paired_bootstrap_ranks_20260918_v1/result.json.
\begin{table}[H]
\topcaption{Rank stability across the 18 downstream tasks.
Bootstrap summaries use the 1,000 shared participant/case resamples within each task.}
\label{tab:bootstrap-ranks}
\centering
\setlength{\parskip}{0pt}
\small
\setlength{\tabcolsep}{6pt}
\renewcommand{\arraystretch}{1}
\begin{tabular}{@{}lrrrr@{}}
\toprule
& \multicolumn{2}{c}{Observed scores} & \multicolumn{2}{c}{Bootstrap mean} \\
\cmidrule(lr){2-3}\cmidrule(l){4-5}
Model & Firsts /18 & Rank $\downarrow$ & Firsts /18 & Rank $\downarrow$ \\
\midrule
RAE-PPG & 12 & 1.444 & 11.217 & 1.656 \\
AnyPPG & 5 & 2.389 & 4.647 & 2.480 \\
SIGMA-PPG & 0 & 3.722 & 0.697 & 3.795 \\
Pulse-PPG & 0 & 3.944 & 0.528 & 3.958 \\
PaPaGei-P & 1 & 4.167 & 0.572 & 3.997 \\
PaPaGei-S & 0 & 5.333 & 0.371 & 5.114 \\
\bottomrule
\end{tabular}
\par\smallskip
\noindent\parbox{\textwidth}{\scriptsize Firsts count tasks with the best score;
joint firsts count for each tied model, so totals may exceed 18.
Ranks use average ranks for ties, with rank 1 best.}
\end{table}

%% file: main4/tables/rae_ppg_seed_robustness_main4.tex
% CONFIRMED RESULT: raeppg2_r1_seeds_eval_20260920_v2/result.json (verification PASS).
% Display-only rendering by main4_rebuild_20260917/render_seed_robustness.py; no new evaluation.
\begin{table}[H]
\topcaption{Sensitivity to the pretraining seed. Primary denotes the run used for the main results; five additional runs vary only the random seed.}
\label{tab:seed-robustness}
\centering
\setlength{\parskip}{0pt}
\fontsize{7.5}{9}\selectfont
\setlength{\tabcolsep}{2.2pt}
\renewcommand{\arraystretch}{1}
\begin{tabular}{@{}llrrrcc@{}}
\toprule
& & & \multicolumn{2}{c}{Five additional runs} & Rank range & Runs \\
\cmidrule(lr){4-5}
Downstream task & Metric & Primary & Mean $\pm$ SD & Range & (six runs) & ahead /6 \\
\midrule
DaLiA: HR\textsuperscript{10} & MAE $\downarrow$ & 8.652 & 8.460 $\pm$ 0.262 & 8.152--8.864 & 2 & 0 \\
BUT PPG: HR\textsuperscript{10} & MAE $\downarrow$ & 10.514 & 10.518 $\pm$ 0.112 & 10.441--10.716 & 2--4 & 0 \\
PTT-PPG: SBP\textsuperscript{10} & MAE $\downarrow$ & 9.529 & 9.629 $\pm$ 0.064 & 9.561--9.709 & 1 & 6 \\
PTT-PPG: DBP\textsuperscript{10} & MAE $\downarrow$ & 5.449 & 5.367 $\pm$ 0.055 & 5.283--5.433 & 1 & 6 \\
DREAMT: Age\textsuperscript{10} & MAE $\downarrow$ & 11.833 & 11.909 $\pm$ 0.061 & 11.813--11.966 & 4--5 & 0 \\
MESA: A/H\textsuperscript{240} & MAE $\downarrow$ & 0.0834 & 0.0842 $\pm$ 0.0002 & 0.0839--0.0843 & 1 & 6 \\
MESA: Arousals\textsuperscript{240} & MAE $\downarrow$ & 0.0432 & 0.0437 $\pm$ 0.0002 & 0.0436--0.0440 & 1--2 & 4 \\
MESA: SpO$_2$\textsuperscript{240} & MAE $\downarrow$ & 0.1768 & 0.1805 $\pm$ 0.0015 & 0.1787--0.1821 & 1 & 6 \\
\cmidrule(lr){1-7}
DeepBeat: AF\textsuperscript{10} & AUROC $\uparrow$ & 0.777 & 0.765 $\pm$ 0.005 & 0.758--0.771 & 2 & 0 \\
DeepBeat: Quality\textsuperscript{10} & F1 $\uparrow$ & 0.672 & 0.682 $\pm$ 0.010 & 0.675--0.700 & 1 & 6 \\
WESAD: Stress\textsuperscript{30} & AUROC $\uparrow$ & 0.951 & 0.952 $\pm$ 0.008 & 0.939--0.960 & 1 & 6 \\
WESAD: Arousal\textsuperscript{30} & AUROC $\uparrow$ & 0.857 & 0.817 $\pm$ 0.007 & 0.810--0.828 & 1 & 6 \\
WESAD: Valence\textsuperscript{30} & AUROC $\uparrow$ & 0.697 & 0.729 $\pm$ 0.009 & 0.719--0.739 & 1 & 6 \\
DaLiA: Activity\textsuperscript{30} & F1 $\uparrow$ & 0.408 & 0.379 $\pm$ 0.022 & 0.353--0.413 & 1 & 6 \\
DREAMT: Sleep\textsuperscript{30} & F1 $\uparrow$ & 0.269 & 0.262 $\pm$ 0.017 & 0.248--0.288 & 1--3 & 1 \\
MESA: Sleep\textsuperscript{240} & F1 $\uparrow$ & 0.474 & 0.467 $\pm$ 0.004 & 0.463--0.472 & 2 & 0 \\
SDB: AHI code\textsuperscript{240} & AUROC $\uparrow$ & 0.678 & 0.695 $\pm$ 0.047 & 0.628--0.758 & 1 & 6 \\
DREAMT: A/H\textsuperscript{240} & AUROC $\uparrow$ & 0.522 & 0.505 $\pm$ 0.020 & 0.474--0.527 & 1--6 & 2 \\
\midrule
Best observed score, tasks /18 & & 12 & \multicolumn{2}{c}{10--12} & & \\
Mean rank $\downarrow$ & & 1.44 & \multicolumn{2}{c}{1.56--1.72} & & \\
\bottomrule
\end{tabular}
\par\smallskip
\noindent\parbox{\textwidth}{\scriptsize Superscripts: input seconds. Mean, sample SD, and score range summarize the five additional runs. Rank range and runs ahead include the primary run, for six runs in total. Ranks compare each RAE-PPG run with the five fixed baseline scores in Table~\ref{tab:transfer-all}; runs ahead counts strict improvements over the best baseline. F1 is macro-F1, and task abbreviations follow Table~\ref{tab:transfer-all}.}
\end{table}

%% file: main4/tables/rae_ppg_cross_duration_ci_main4.tex
% CONFIRMED RESULT: raeppg2_canonical_export_20260920_v1/fixed/result.json; verified canonical predictions.
\begin{table}[!htb]
  \fontsize{8}{9}\selectfont
  \setlength{\tabcolsep}{2pt}
  \topcaption{Fixed-partition cross-duration evaluation of RAE-PPG. Probes are selected and fitted on TRAIN and evaluated on VAL+TEST. Brackets give 95\% participant/case-cluster bootstrap intervals.}
  \label{tab:cross-duration-ci}
\centering
  \newcommand{\durationci}[3]{\mbox{\scriptsize #1\,[#2--#3]}}
  \renewcommand{\arraystretch}{1}
  \renewcommand{\tabularxcolumn}[1]{m{#1}}
  \begin{tabularx}{\textwidth}{@{}>{\raggedright\arraybackslash}X*{3}{>{\centering\arraybackslash}m{75pt}}*{2}{>{\centering\arraybackslash}m{29pt}}@{}}
    \toprule
    Task & 10~s & 30~s & 240~s & {\scriptsize Retention} & Extend \\
    \midrule
    DaLiA heart rate\newline{\scriptsize MAE $\downarrow$, ref. 10~s} &
      \durationci{8.6522}{6.3069}{11.1771} & \durationci{7.0623}{5.2881}{8.7911} & \durationci{8.8074}{7.8910}{9.8989} & $\times$ & -- \\
    DaLiA activity\newline{\scriptsize F1 $\uparrow$, ref. 30~s} &
      \durationci{0.3097}{0.2425}{0.3547} & \durationci{0.4078}{0.3115}{0.5070} & \durationci{0.4581}{0.4109}{0.4975} & $\times$ & $\times$ \\
    WESAD stress\newline{\scriptsize AUROC $\uparrow$, ref. 30~s} &
      \durationci{0.8436}{0.7286}{0.9640} & \durationci{0.8821}{0.7522}{0.9750} & \durationci{0.9410}{0.8083}{1.0000} & $\times$ & $\times$ \\
    WESAD arousal\newline{\scriptsize AUROC $\uparrow$, ref. 30~s} &
      \durationci{0.7527}{0.6300}{0.8685} & \durationci{0.8571}{0.7583}{0.9593} & \durationci{0.7875}{0.6763}{0.8964} & $\times$ & $\checkmark$ \\
    WESAD valence\newline{\scriptsize AUROC $\uparrow$, ref. 30~s} &
      \durationci{0.6907}{0.6469}{0.7590} & \durationci{0.6973}{0.6023}{0.8406} & \durationci{0.6080}{0.4887}{0.7766} & $\times$ & $\times$ \\
    MESA sleep staging\newline{\scriptsize F1 $\uparrow$, ref. 30~s} &
      \durationci{0.3607}{0.3556}{0.3656} & \durationci{0.3821}{0.3758}{0.3889} & \durationci{0.4735}{0.4640}{0.4833} & $\checkmark$ & $\checkmark$ \\
    MESA A/H burden\newline{\scriptsize MAE $\downarrow$, ref. 240~s} &
      \durationci{0.0930}{0.0906}{0.0956} & \durationci{0.0912}{0.0889}{0.0937} & \durationci{0.0834}{0.0813}{0.0858} & -- & $\checkmark$ \\
    MESA arousal-event burden\newline{\scriptsize MAE $\downarrow$, ref. 240~s} &
      \durationci{0.0521}{0.0510}{0.0533} & \durationci{0.0505}{0.0494}{0.0516} & \durationci{0.0432}{0.0422}{0.0442} & -- & $\checkmark$ \\
    \bottomrule
  \end{tabularx}
\end{table}

Score aggregation and the 1,000 shared participant/case-cluster bootstrap draws follow Appendix~\ref{app:linear-evaluation}.
Utility is $U=-\mathrm{MAE}$ for regression and AUROC or macro-F1 for classification.
Retention compares a 10~s reference with both longer durations, or a 30~s reference with 240~s.
Its task-level $p$-value is the maximum of the component $p$-values.
A task meets Retention when every paired two-sided 95\% interval has a lower bound above the negative task-specific margin and the Holm-adjusted task-level $p$-value is at most $.05$.
Extend compares the immediately preceding duration with the reference and requires the lower endpoint of its paired two-sided 95\% interval to exceed zero, together with Holm-adjusted $p\leq.05$.
Null-centered one-sided tests use a plus-one correction, with separate Holm families for the eight Retention and eleven Extend tests~\citep{holm1979multiple}.
Table~\ref{tab:cross-duration-ci} reports all scores, intervals, and decisions.

% CONFIRMED RESULT: raeppg2_canonical_export_20260920_v1/fixed/result.json; verified canonical predictions.
\begin{table}[!htb]
  \addtocounter{table}{-1}
  \renewcommand{\theHtable}{\arabic{table}.continued}
  \fontsize{8}{9}\selectfont
  \setlength{\tabcolsep}{2pt}
  \topcaption{Fixed-partition cross-duration evaluation of RAE-PPG (continued).}
\centering
  \newcommand{\durationci}[3]{\mbox{\scriptsize #1\,[#2--#3]}}
  \renewcommand{\arraystretch}{1}
  \renewcommand{\tabularxcolumn}[1]{m{#1}}
  \begin{tabularx}{\textwidth}{@{}>{\raggedright\arraybackslash}X*{3}{>{\centering\arraybackslash}m{75pt}}*{2}{>{\centering\arraybackslash}m{29pt}}@{}}
    \toprule
    Task & 10~s & 30~s & 240~s & {\scriptsize Retention} & Extend \\
    \midrule
    MESA SpO$_2$ desaturation burden\newline{\scriptsize MAE $\downarrow$, ref. 240~s} &
      \durationci{0.2258}{0.2228}{0.2291} & \durationci{0.2166}{0.2137}{0.2195} & \durationci{0.1768}{0.1739}{0.1797} & -- & $\checkmark$ \\
    SDB AHI-code classification\newline{\scriptsize AUROC $\uparrow$, ref. 240~s} &
      \durationci{0.4286}{0.2802}{0.6050} & \durationci{0.5000}{0.3483}{0.6486} & \durationci{0.6779}{0.5250}{0.8184} & -- & $\times$ \\
    DREAMT age\newline{\scriptsize MAE $\downarrow$, ref. 10~s} &
      \durationci{11.8329}{9.1310}{14.7515} & \durationci{11.1262}{8.6348}{13.9619} & \durationci{10.3018}{7.5973}{13.2041} & $\times$ & -- \\
    DREAMT sleep staging\newline{\scriptsize F1 $\uparrow$, ref. 30~s} &
      \durationci{0.2643}{0.2327}{0.2999} & \durationci{0.2687}{0.2361}{0.3071} & \durationci{0.3014}{0.2705}{0.3322} & $\checkmark$ & $\times$ \\
    DREAMT A/H detection\newline{\scriptsize AUROC $\uparrow$, ref. 240~s} &
      \durationci{0.4958}{0.4282}{0.5627} & \durationci{0.4500}{0.3684}{0.5383} & \durationci{0.5225}{0.4355}{0.6143} & -- & $\checkmark$ \\
    \midrule
    \multicolumn{4}{@{}l}{Confirmed tasks / applicable tasks} & 2/8 & 6/11 \\
    \bottomrule
  \end{tabularx}
  \par\smallskip
  \noindent\parbox{\textwidth}{\scriptsize Reference is the prespecified duration for this analysis, which can differ from the primary reporting duration. $\checkmark$: both the paired-interval and Holm-adjusted test criteria are met; $\times$: these criteria are not met; --: not applicable.}
\end{table}

%% file: main4/tables/rae_ppg_cross_duration_oof_supp_main4.tex
% CONFIRMED RESULT: raeppg2_canonical_export_20260920_v1/oof/result.json; verified canonical predictions.
\begin{table}[!htb]
  \fontsize{8}{9}\selectfont
  \setlength{\tabcolsep}{2pt}
  \topcaption{Grouped out-of-fold cross-duration evaluation of RAE-PPG, using five outer and four inner participant-grouped folds. Brackets give 95\% participant/case-cluster bootstrap intervals; Retention and Extend follow the criteria in Appendix~\ref{app:cross-duration-scores}.}
  \label{tab:oof-cross-duration-ci}
\centering
  \newcommand{\oofdurationci}[3]{\mbox{\scriptsize #1\,[#2--#3]}}
  \renewcommand{\arraystretch}{1}
  \renewcommand{\tabularxcolumn}[1]{m{#1}}
  \begin{tabularx}{\textwidth}{@{}>{\raggedright\arraybackslash}X*{3}{>{\centering\arraybackslash}m{75pt}}*{2}{>{\centering\arraybackslash}m{29pt}}@{}}
    \toprule
    Task & 10~s & 30~s & 240~s & {\scriptsize Retention} & Extend \\
    \midrule
    DaLiA heart rate\newline{\scriptsize MAE $\downarrow$, ref. 10~s} &
      \oofdurationci{8.4987}{7.1158}{10.1447} & \oofdurationci{7.3716}{6.2626}{8.6637} & \oofdurationci{8.9185}{7.8384}{10.3123} & $\times$ & -- \\
    DaLiA activity\newline{\scriptsize F1 $\uparrow$, ref. 30~s} &
      \oofdurationci{0.3148}{0.2824}{0.3469} & \oofdurationci{0.3795}{0.3291}{0.4273} & \oofdurationci{0.4651}{0.4070}{0.5196} & $\checkmark$ & $\checkmark$ \\
    WESAD stress\newline{\scriptsize AUROC $\uparrow$, ref. 30~s} &
      \oofdurationci{0.7792}{0.6938}{0.8729} & \oofdurationci{0.9446}{0.9071}{0.9762} & \oofdurationci{0.8742}{0.7992}{0.9582} & $\times$ & $\checkmark$ \\
    WESAD arousal\newline{\scriptsize AUROC $\uparrow$, ref. 30~s} &
      \oofdurationci{0.7430}{0.6458}{0.8500} & \oofdurationci{0.7556}{0.6825}{0.8368} & \oofdurationci{0.7398}{0.6140}{0.8644} & $\times$ & $\times$ \\
    \bottomrule
  \end{tabularx}
\end{table}

\begin{table}[!htb]
  \addtocounter{table}{-1}
  \renewcommand{\theHtable}{\arabic{table}.continued}
  \fontsize{8}{9}\selectfont
  \setlength{\tabcolsep}{2pt}
  \topcaption{Grouped out-of-fold cross-duration evaluation of RAE-PPG (continued).}
\centering
  \newcommand{\oofdurationci}[3]{\mbox{\scriptsize #1\,[#2--#3]}}
  \renewcommand{\arraystretch}{1}
  \renewcommand{\tabularxcolumn}[1]{m{#1}}
  \begin{tabularx}{\textwidth}{@{}>{\raggedright\arraybackslash}X*{3}{>{\centering\arraybackslash}m{75pt}}*{2}{>{\centering\arraybackslash}m{29pt}}@{}}
    \toprule
    Task & 10~s & 30~s & 240~s & {\scriptsize Retention} & Extend \\
    \midrule
    WESAD valence\newline{\scriptsize AUROC $\uparrow$, ref. 30~s} &
      \oofdurationci{0.6757}{0.6011}{0.7900} & \oofdurationci{0.6792}{0.5729}{0.8272} & \oofdurationci{0.6414}{0.5312}{0.7686} & $\times$ & $\times$ \\
    MESA sleep staging\newline{\scriptsize F1 $\uparrow$, ref. 30~s} &
      \oofdurationci{0.3594}{0.3563}{0.3627} & \oofdurationci{0.3789}{0.3754}{0.3830} & \oofdurationci{0.4688}{0.4633}{0.4749} & $\checkmark$ & $\checkmark$ \\
    MESA A/H burden\newline{\scriptsize MAE $\downarrow$, ref. 240~s} &
      \oofdurationci{0.0934}{0.0920}{0.0949} & \oofdurationci{0.0920}{0.0906}{0.0934} & \oofdurationci{0.0840}{0.0827}{0.0854} & -- & $\checkmark$ \\
    MESA arousal-event burden\newline{\scriptsize MAE $\downarrow$, ref. 240~s} &
      \oofdurationci{0.0519}{0.0512}{0.0526} & \oofdurationci{0.0504}{0.0498}{0.0511} & \oofdurationci{0.0436}{0.0430}{0.0443} & -- & $\checkmark$ \\
    MESA SpO$_2$ desaturation burden\newline{\scriptsize MAE $\downarrow$, ref. 240~s} &
      \oofdurationci{0.2255}{0.2235}{0.2275} & \oofdurationci{0.2164}{0.2144}{0.2184} & \oofdurationci{0.1779}{0.1760}{0.1798} & -- & $\checkmark$ \\
    SDB AHI-code classification\newline{\scriptsize AUROC $\uparrow$, ref. 240~s} &
      \oofdurationci{0.5023}{0.4121}{0.5992} & \oofdurationci{0.5769}{0.4737}{0.6762} & \oofdurationci{0.6730}{0.5847}{0.7647} & -- & $\times$ \\
    DREAMT age\newline{\scriptsize MAE $\downarrow$, ref. 10~s} &
      \oofdurationci{11.1645}{9.6883}{12.8526} & \oofdurationci{10.8672}{9.3503}{12.5800} & \oofdurationci{10.8876}{9.4015}{12.4720} & $\times$ & -- \\
    DREAMT sleep staging\newline{\scriptsize F1 $\uparrow$, ref. 30~s} &
      \oofdurationci{0.2276}{0.2114}{0.2430} & \oofdurationci{0.2749}{0.2525}{0.2965} & \oofdurationci{0.3109}{0.2851}{0.3341} & $\checkmark$ & $\checkmark$ \\
    DREAMT A/H detection\newline{\scriptsize AUROC $\uparrow$, ref. 240~s} &
      \oofdurationci{0.4889}{0.4513}{0.5238} & \oofdurationci{0.5226}{0.4794}{0.5646} & \oofdurationci{0.5931}{0.5452}{0.6344} & -- & $\checkmark$ \\
    \midrule
    \multicolumn{4}{@{}l}{Confirmed tasks / applicable tasks} & 3/8 & 8/11 \\
    \bottomrule
  \end{tabularx}
  \par\smallskip
  \noindent\parbox{\textwidth}{\scriptsize Reference is the prespecified duration for this analysis, which can differ from the primary reporting duration. $\checkmark$: both the paired-interval and Holm-adjusted test criteria are met; $\times$: these criteria are not met; --: not applicable.}
\end{table}

%% file: main4/tables/rae_ppg_physiology_all_main4.tex
% CONFIRMED RESULT: original 45 feature/view values transposed without recomputation.
\begin{table}[!htb]
\topcaption{Decoding of the sixteen signal features from the final frozen encoder. Column headings give input duration, and entries report gain [95\% CI]. Readouts are specified by panel or row.}
\label{tab:physiology-all}
\centering
\setlength{\parskip}{0pt}
\scriptsize
\setlength{\tabcolsep}{3pt}
\renewcommand{\arraystretch}{1}
\begin{tabularx}{\textwidth}{@{}>{\raggedright\arraybackslash}Xr*{3}{>{\centering\arraybackslash}p{75pt}}@{}}
\toprule
Signal feature / readout & $N$ & 10~s & 30~s & 240~s \\
\midrule
\multicolumn{5}{@{}l}{\textit{A. Features introduced at 10~s; centered 10~s readout}} \\
Slope transit time & 85,150 & 0.930\, [0.915, 0.942] & 0.935\, [0.922, 0.945] & 0.782\, [0.758, 0.805] \\
APG $b/a$ & 84,187 & 0.767\, [0.720, 0.812] & 0.775\, [0.727, 0.819] & 0.670\, [0.624, 0.714] \\
APG $d/a$ & 77,251 & 0.643\, [0.586, 0.697] & 0.627\, [0.572, 0.680] & 0.450\, [0.394, 0.500] \\
Onset-to-APG-$b$ & 84,187 & 0.866\, [0.850, 0.881] & 0.859\, [0.842, 0.875] & 0.727\, [0.707, 0.746] \\
Normalized crest time & 85,153 & 0.864\, [0.841, 0.885] & 0.855\, [0.833, 0.875] & 0.628\, [0.597, 0.657] \\
Normalized PW50 & 85,148 & 0.905\, [0.884, 0.923] & 0.901\, [0.877, 0.920] & 0.771\, [0.744, 0.795] \\
Split-area ratio & 83,614 & 0.694\, [0.655, 0.732] & 0.668\, [0.627, 0.707] & 0.299\, [0.263, 0.335] \\
Peak-interval rate & 85,432 & 0.942\, [0.928, 0.954] & 0.947\, [0.935, 0.958] & 0.864\, [0.848, 0.879] \\
Spectral purity & 87,302 & 0.933\, [0.924, 0.942] & 0.904\, [0.890, 0.917] & 0.607\, [0.577, 0.635] \\
\midrule
\multicolumn{5}{@{}l}{\textit{B. Features introduced at 30~s; readout specified below}} \\
Normalized reflected delay\newline 30~s readout & 79,108 & -- & 0.904\, [0.896, 0.912] & 0.677\, [0.654, 0.700] \\
\hspace{1em}token mean &  & 0.646\, [0.618, 0.672] & 0.904\, [0.896, 0.912] & -- \\
PRV-RMSSD\newline 30~s readout & 75,734 & -- & 0.892\, [0.877, 0.905] & 0.655\, [0.623, 0.685] \\
\hspace{1em}token mean &  & 0.577\, [0.541, 0.610] & 0.892\, [0.877, 0.905] & -- \\
\midrule
\multicolumn{5}{@{}l}{\textit{C. Features introduced at 240~s; token mean}} \\
PPGV LF fraction & 86,626 & -- & 0.194\, [0.159, 0.228] & 0.544\, [0.515, 0.572] \\
Pulse-interval SD & 82,027 & -- & 0.555\, [0.511, 0.596] & 0.761\, [0.709, 0.805] \\
Mean pulse rate & 87,255 & -- & 0.910\, [0.895, 0.923] & 0.938\, [0.926, 0.947] \\
PRV-pNN50 & 84,794 & -- & 0.787\, [0.763, 0.808] & 0.779\, [0.751, 0.805] \\
Uncentered ACF index & 87,274 & -- & 0.451\, [0.407, 0.493] & 0.755\, [0.718, 0.789] \\
\bottomrule
\end{tabularx}
\par\smallskip
\noindent\parbox{\textwidth}{\scriptsize An interval readout uses overlap-weighted pooling over the centered interval; token mean averages all 64 signal tokens. $N$ counts available trajectories and is shared across a feature's views. Intervals describe individual-view gains; paired comparisons are reported in Appendix~\ref{app:physiological-protocol}. Dashes indicate views not reported in this table.}
\end{table}

%% file: main4/tables/rae_ppg_bridge_evidence_map_main4.tex
% CONFIRMED RESULT: 33 original relations grouped into 26 source-summary rows.
\begin{table}[H]
\centering\scriptsize
\setlength{\tabcolsep}{3pt}\renewcommand{\arraystretch}{1}
\caption{Literature sources for the fixed 33 Bridge relations. Tasks sharing the same source and finding are grouped within each feature family. Asterisks mark the 22 core anchors. Source measurements and findings, including mixed or nonsignificant results, are distinct from the Bridge outcomes reported in Table~\ref{tab:bridge-relation-matrix}A.}
\label{tab:bridge-evidence-map}
\begin{tabularx}{\linewidth}{@{}>{\raggedright\arraybackslash}p{.16\linewidth}>{\raggedright\arraybackslash}p{.35\linewidth}>{\raggedright\arraybackslash}X@{}}
\toprule
Task(s) & Source measurement and setting & Reported finding \\
\midrule
\multicolumn{3}{@{}l}{\textit{Rate}} \\
DaLiA HR$^{*}$; BUT PPG HR$^{*}$ & Smartphone PPG HR compared with simultaneous 10~s ECG recordings in a clinic subset~\citep{avram2019heartrate}. & HR was compared with ECG in that acquisition setting; the DaLiA and BUT PPG settings differ. \\
PTT SBP; PTT DBP & Resting HR and BP measured by the same automated BP monitor after seated rest~\citep{liu2010heartratebp}. & Adjusted population-level associations of HR with SBP and DBP; not a PPG-based prediction experiment. \\
DeepBeat AF$^{*}$ & PPG from AF and sinus-rhythm groups; 120~s segments from 10~min recordings~\citep{basza2023afmorphology}. & Univariate HR difference; the groups also differed in age and clinical characteristics. \\
DREAMT age & Smartphone PPG HR in a large cohort~\citep{avram2019heartrate}. & HR was associated with age; the source used smartphone rather than wrist PPG. \\
WESAD stress$^{*}$ & 30~s fingertip-PPG intervals during Stroop and rest~\citep{tsai2025stress}. & Mean RR interval decreased during Stroop phases. \\
WESAD arousal$^{*}$ & ECG HR over 60~s portions of emotion blocks~\citep{brouwer2013emotion}. & HR was associated with arousal; the source signal was ECG. \\
WESAD valence$^{*}$ & ECG HR over 60~s portions of emotion blocks~\citep{brouwer2013emotion}. & HR was associated with valence; the source signal was ECG. \\
DaLiA activity & Daily-life wrist-PPG pulse time compared with a 1~min activity index~\citep{moscato2022dailyppg}. & Pulse time was associated with the activity index, which differs from the present activity labels. \\
DREAMT sleep$^{*}$; MESA sleep$^{*}$ & Fingertip-PPG HR by sleep stage, with labels assigned to 30~s epochs~\citep{yilmaz2023sleepppg}. & HR differed across sleep stages; the present input durations are fixed evaluation settings. \\
MESA A/H burden; DREAMT A/H presence & ECG R--R-derived HR around respiratory events~\citep{habarubio2005arousal}. & Event-locked HR changes; the present window-level burden and presence endpoints differ. \\
MESA arousal burden$^{*}$ & ECG HR time courses~\citep{habarubio2005arousal} and PPG power-weighted mean frequency around arousals~\citep{luukinen2024arousal}. & HR time courses differed by arousal grade, but maximal HR rise and $\Delta$HR did not differ significantly. PPG spectral-mean-frequency responses differed by arousal cause; the burden endpoint differs. \\
SDB AHI code$^{*}$ & Overnight summaries of 2~min pulse-interval windows in source-defined pediatric SDB groups (AHI $\geq5$)~\citep{garde2014sdb}. & Mean-interval contrast: $p=.04$ versus the source criterion $p<.01$. Source groups are distinct from the released-code endpoint in Appendix~\ref{app:downstream-details}. \\
\multicolumn{3}{@{}l}{\textit{Morphology / vascular timing}} \\
PTT SBP$^{*}$ & One-minute means of PPG area-ratio and timing indices versus cuff BP across rest, exercise, and recovery~\citep{li2014ppgbp}. & SBP associations; multivariable models also included ECG--PPG pulse transit time. \\
PTT DBP$^{*}$ & Single-index fits using PPG-derived features~\citep{li2014ppgbp}. & DBP was calculated as predicted SBP minus predicted pulse pressure; multivariable models also included ECG--PPG pulse transit time. \\
DeepBeat AF$^{*}$ & PPG waveform descriptors in AF and sinus-rhythm groups; 120~s segments from 10~min recordings~\citep{basza2023afmorphology}. & Differences in selected features, including $b/a$ and crest time; groups differed in age and clinical characteristics, and crest time was not pulse-duration normalized. \\
DREAMT age$^{*}$ & Fingertip APG in 600 adults~\citep{takazawa1998apg}. & $b/a$ increased and $d/a$ decreased with age; cross-sectional, not wrist. \\
WESAD stress$^{*}$ & Arm-PPG timing across Stroop and rest phases~\citep{celka2020stress}. & Crest-time significance varied by contrast: $p=.07$ for baseline versus the first Stroop phase; later contrasts were significant. \\
WESAD arousal & Repeated 4~min smartphone PPG recordings and self-reported arousal~\citep{liu2026mentalhealthppg}. & Crest time met nominal significance with cluster-robust standard errors, but not with participant-level aggregation or the Bonferroni threshold of $.0056$. \\
DaLiA activity$^{*}$ & Daily-life wrist PPG compared with a 1~min activity index~\citep{moscato2022dailyppg}. & Pulse time and systolic--diastolic delay were associated with the activity index; rise time was not. \\
\bottomrule
\end{tabularx}
\end{table}

\newpage % Continue at a complete panel/row boundary.
\begin{table}[H]
\centering\scriptsize
\setlength{\tabcolsep}{3pt}\renewcommand{\arraystretch}{1}
\addtocounter{table}{-1}
\renewcommand{\theHtable}{\arabic{table}.continued}
\caption{Literature sources for the fixed Bridge relations (continued).}
\begin{tabularx}{\linewidth}{@{}>{\raggedright\arraybackslash}p{.16\linewidth}>{\raggedright\arraybackslash}p{.35\linewidth}>{\raggedright\arraybackslash}X@{}}
\toprule
Task(s) & Source measurement and setting & Reported finding \\
\midrule
\multicolumn{3}{@{}l}{\textit{Morphology / vascular timing (continued)}} \\
DREAMT sleep; MESA sleep & Fingertip-PPG timing and shape by sleep stage~\citep{yilmaz2023sleepppg}. & Stage differences in normalized timing and shape; the present input durations are fixed evaluation settings. \\
\multicolumn{3}{@{}l}{\textit{Beat-interval variability}} \\
WESAD stress$^{*}$ & 30~s fingertip-PPG RMSSD during Stroop and rest~\citep{tsai2025stress}. & Sex-specific RMSSD responses were described; pooled adjacent-phase differences were not marked significant. \\
WESAD arousal$^{*}$ & ECG normalized RMSSD over 60~s; the source specified at least 1~min for HRV analysis~\citep{brouwer2013emotion}. & RMSSD was associated with arousal but not valence; the source signal was ECG. \\
DaLiA activity & Five-minute wrist-PPG variability at 06:00--07:00 compared with mean daily steps over the preceding 90 days~\citep{natarajan2020ppgvariability}. & Activity associations were quantified for HF and LF power; analogous findings for other HRV metrics were stated without data. Habitual steps differ from contemporaneous activity labels. \\
DREAMT sleep$^{*}$; MESA sleep$^{*}$ & A model trained with ECG-derived HRV and movement features and evaluated using wrist-PPG-derived HRV and movement~\citep{fonseca2017ppgsleep}. & Sleep-stage prediction used HRV and movement jointly; 30~s denotes label epochs rather than every feature's analysis interval. \\
MESA A/H burden$^{*}$; DREAMT A/H presence$^{*}$ & Whole-night wrist-PPG pulse-interval variability compared with sleep-apnea indices~\citep{hayano2020apnea}. & Conventional indices were not significantly associated with AHI; whole-night severity differs from the present 240~s burden and presence endpoints. \\
SDB AHI code & Overnight means and across-window dispersions of 2~min PRV values~\citep{garde2014sdb}. & Mean RMSSD and SDNN showed no significant group differences. Their dispersions met $p<.01$ under Mann--Whitney tests, but not log-transformed t-tests; these differ from the averaged Bridge features. \\
\bottomrule
\end{tabularx}
\end{table}

%% file: main4/tables/rae_ppg_bridge_relation_matrix_main4.tex
\begin{table}[H]
\centering\scriptsize
\setlength{\tabcolsep}{3pt}\renewcommand{\arraystretch}{1}
\caption{Bridge results for all task--feature-family combinations (A) and source-sensitivity subsets (B).}
\label{tab:bridge-relation-matrix}
\par{\raggedright \noindent\textbf{A. Task--feature-family results}\par}
\begin{tabularx}{\linewidth}{@{}Xrccccc@{}}
\toprule Task & Seconds & Rate & Morph. & BIV & Wave. & SQ \\
\midrule
    DaLiA HR & 10 & $R^{*}\;+/+$ & $U\;+/+$ & -- & -- & $U\;-/-$ \\
    BUT PPG HR & 10 & $R^{*}\;-/-$ & $U\;+/+$ & -- & -- & $U\;-/-$ \\
    PTT SBP & 10 & $I^{\ddagger}\;-/-$ & $R^{*}\;+/-$ & -- & -- & $U\;-/-$ \\
    PTT DBP & 10 & $I^{\ddagger}\;-/-$ & $I^{*}\;+/+$ & -- & -- & $U\;-/-$ \\
    DeepBeat AF & 10 & $R^{*}\;-/-$ & $R^{*\ddagger}\;+/+$ & -- & -- & $U\;+/+$ \\
    DeepBeat quality & 10 & $U\;+/+$ & $C\;+/+$ & -- & -- & $C\;+/+$ \\
    DREAMT age & 10 & $R\;-/-$ & $R^{*}\;+/+$ & -- & -- & $U\;-/-$ \\
    WESAD stress & 30 & $R^{*}\;+/+$ & $R^{*\ddagger}\;+/+$ & $I^{*}\;-/-$ & -- & $U\;+/+$ \\
    WESAD arousal & 30 & $I^{*}\;+/-$ & $R^{\ddagger}\;+/+$ & $I^{*}\;-/-$ & -- & $U\;+/+$ \\
    WESAD valence & 30 & $I^{*}\;-/-$ & $C\;+/+$ & $U\;-/-$ & -- & $U\;+/+$ \\
    DaLiA activity & 30 & $I\;+/+$ & $R^{*\ddagger}\;+/+$ & $I\;-/-$ & -- & $C\;+/+$ \\
    DREAMT sleep staging & 30 & $R^{*}\;+/+$ & $R\;+/+$ & $R^{*\ddagger}\;-/+$ & -- & $U\;+/+$ \\
    MESA sleep staging & 240 & $R^{*}\;+/+$ & $R\;+/+$ & $R^{*\ddagger}\;+/+$ & $C\;+/+$ & $U\;+/+$ \\
    MESA apnea/hypopnea burden & 240 & $I\;+/+$ & $C\;+/+$ & $I^{*}\;+/+$ & $C\;+/+$ & $U\;+/+$ \\
    MESA arousal burden & 240 & $I^{*}\;+/+$ & $C\;+/+$ & $C^{\ddagger}\;+/+$ & $C^{\ddagger}\;+/+$ & $U\;+/+$ \\
    MESA desaturation burden & 240 & $C\;+/+$ & $C\;+/+$ & $C^{\ddagger}\;+/+$ & $C\;+/+$ & $U\;+/+$ \\
    SDB AHI-code classification (binary; participant-level) & 240 & $R^{*\ddagger}\;+/+$ & $C\;-/-$ & $I\;-/-$ & $C\;-/-$ & $U\;-/-$ \\
    DREAMT apnea/hypopnea presence & 240 & $I\;-/-$ & $C\;-/-$ & $I^{*}\;-/-$ & $C\;-/-$ & $U\;-/-$ \\
\bottomrule\end{tabularx}
\par\smallskip
\noindent\parbox{\textwidth}{\scriptsize R: related measurement; I: indirect support; C: contextual; U: unmapped in the cited sources. Labels are the original classifications. $^{*}$ marks core anchors; $\ddagger$ marks re-audit changes (R to I, I to C, or C to U). Sign pairs give PPG/Decoded decisions: $+$ meets the one-sided 95\% lower-bound criterion, $-$ does not; dashes denote duration-ineligible families. Decisions are relation-wise and unadjusted. Morph.: morphology/vascular timing; BIV: beat-interval variability; Wave.: waveform statistics; SQ: signal quality.}
\end{table}

\newpage % Continue at a complete panel/row boundary.
\begin{table}[H]
\centering\scriptsize
\setlength{\tabcolsep}{3pt}\renewcommand{\arraystretch}{1}
\addtocounter{table}{-1}
\renewcommand{\theHtable}{\arabic{table}.continued}
\caption{Bridge results for all task--feature-family combinations and source-sensitivity subsets (continued).}
\par\medskip
{\raggedright \noindent\textbf{B. Source-sensitivity subsets}\par}
\begin{tabularx}{\textwidth}{@{}Xrrrr@{}}
\toprule
Analysis set & Relations & PPG positive & Decoded positive & Both positive \\
\midrule
Fixed analysis set & 33 & 20 & 19 & 18 \\
Alternative R/I subset & 31 & 20 & 19 & 18 \\
Excluding the two A/H--BIV relations & 31 & 19 & 18 & 17 \\
Both restrictions & 29 & 19 & 18 & 17 \\
\bottomrule
\end{tabularx}
\par\smallskip
\noindent\parbox{\textwidth}{\scriptsize Alternative R/I applies the marked source reclassifications before selecting R/I relations. The two A/H--BIV relations concern MESA burden and DREAMT presence. Both positive counts the intersection of pathway-wise decisions; predictions and decision thresholds are unchanged across subsets.}
\end{table}

%% file: main4/tables/rae_ppg_ablation_terminal_main4.tex
% CONFIRMED RESULT: all 72 cells from raeppg2_r1_p4_complete_20260916_v2.
\begin{table}[!ht]
  \fontsize{8}{9}\selectfont
  \setlength{\tabcolsep}{2pt}
  \renewcommand{\arraystretch}{1}
  \topcaption{Final downstream performance of the four ablation conditions. Each score is followed by its 95\% participant/case-cluster bootstrap interval. Bold marks the best point score, including ties; F1 denotes macro-F1. Condition definitions are given in Appendix~\ref{app:ablation-conditions}.}
  \label{tab:ablation-terminal}
\centering
  \newcommand{\ablationci}[3]{\shortstack{#1\\{\scriptsize [#2, #3]}}}
  \renewcommand{\tabularxcolumn}[1]{m{#1}}
  \begin{tabularx}{\textwidth}{@{}>{\raggedright\arraybackslash}X*{4}{>{\centering\arraybackslash}m{58pt}}@{}}
    \toprule
    & \multicolumn{2}{c}{Nested/protected} & \multicolumn{2}{c}{Fully shared} \\
    \cmidrule(lr){2-3}\cmidrule(l){4-5}
    Downstream task & A: ON & D: OFF & B: ON & C: OFF \\
    \midrule
    PPG-DaLiA heart rate\newline{\scriptsize 10~s, MAE $\downarrow$} &
      \ablationci{\textbf{8.6522}}{6.3069}{11.1771} & \ablationci{\textbf{8.6522}}{6.3069}{11.1771} & \ablationci{9.8918}{7.9631}{11.9085} & \ablationci{10.8341}{8.8499}{12.8399} \\
    BUT PPG ECG-referenced heart rate\newline{\scriptsize 10~s, MAE $\downarrow$} &
      \ablationci{\textbf{10.5144}}{8.2084}{13.3728} & \ablationci{\textbf{10.5144}}{8.2084}{13.3728} & \ablationci{10.8212}{8.0354}{14.1561} & \ablationci{10.7846}{8.0600}{14.0935} \\
    DeepBeat atrial fibrillation\newline{\scriptsize 10~s, AUROC $\uparrow$} &
      \ablationci{\textbf{0.7774}}{0.7164}{0.8087} & \ablationci{\textbf{0.7774}}{0.7164}{0.8087} & \ablationci{0.7040}{0.6700}{0.7336} & \ablationci{0.6895}{0.6464}{0.7170} \\
    DeepBeat signal quality\newline{\scriptsize 10~s, F1 $\uparrow$} &
      \ablationci{\textbf{0.6718}}{0.6360}{0.6903} & \ablationci{\textbf{0.6718}}{0.6360}{0.6903} & \ablationci{0.6601}{0.6158}{0.6916} & \ablationci{0.6647}{0.6297}{0.6882} \\
    PTT-PPG systolic blood pressure\newline{\scriptsize 10~s, MAE $\downarrow$} &
      \ablationci{\textbf{9.5291}}{5.2400}{14.8673} & \ablationci{\textbf{9.5291}}{5.2400}{14.8673} & \ablationci{9.8373}{5.5890}{15.0733} & \ablationci{10.0162}{5.5610}{15.2647} \\
    PTT-PPG diastolic blood pressure\newline{\scriptsize 10~s, MAE $\downarrow$} &
      \ablationci{\textbf{5.4489}}{3.7838}{7.2084} & \ablationci{\textbf{5.4489}}{3.7838}{7.2084} & \ablationci{5.5652}{4.1228}{7.1169} & \ablationci{5.6967}{4.1493}{7.3704} \\
    WESAD stress\newline{\scriptsize 30~s, AUROC $\uparrow$} &
      \ablationci{\textbf{0.9514}}{0.9183}{0.9794} & \ablationci{0.9298}{0.8776}{0.9667} & \ablationci{0.9042}{0.8076}{0.9694} & \ablationci{0.9017}{0.8152}{0.9682} \\
    WESAD arousal\newline{\scriptsize 30~s, AUROC $\uparrow$} &
      \ablationci{\textbf{0.8571}}{0.7583}{0.9593} & \ablationci{0.8425}{0.7617}{0.9238} & \ablationci{0.7033}{0.5655}{0.8298} & \ablationci{0.8022}{0.6591}{0.9325} \\
    WESAD valence\newline{\scriptsize 30~s, AUROC $\uparrow$} &
      \ablationci{0.6973}{0.6023}{0.8406} & \ablationci{\textbf{0.7413}}{0.6577}{0.8252} & \ablationci{0.6227}{0.5031}{0.7736} & \ablationci{0.6493}{0.5679}{0.7619} \\
    PPG-DaLiA activity\newline{\scriptsize 30~s, F1 $\uparrow$} &
      \ablationci{\textbf{0.4078}}{0.3115}{0.5070} & \ablationci{0.3405}{0.2841}{0.3858} & \ablationci{0.3650}{0.3024}{0.4216} & \ablationci{0.3434}{0.2986}{0.3773} \\
    MESA sleep staging\newline{\scriptsize 240~s, F1 $\uparrow$} &
      \ablationci{0.4735}{0.4640}{0.4833} & \ablationci{0.4727}{0.4649}{0.4806} & \ablationci{\textbf{0.4967}}{0.4878}{0.5057} & \ablationci{0.4934}{0.4844}{0.5019} \\
    MESA apnea/hypopnea burden\newline{\scriptsize 240~s, MAE $\downarrow$} &
      \ablationci{0.0834}{0.0813}{0.0858} & \ablationci{0.0840}{0.0818}{0.0864} & \ablationci{\textbf{0.0824}}{0.0802}{0.0847} & \ablationci{0.0829}{0.0808}{0.0852} \\
    MESA arousal-event burden\newline{\scriptsize 240~s, MAE $\downarrow$} &
      \ablationci{0.0432}{0.0422}{0.0442} & \ablationci{0.0438}{0.0428}{0.0448} & \ablationci{\textbf{0.0420}}{0.0411}{0.0430} & \ablationci{0.0428}{0.0418}{0.0438} \\
    MESA SpO$_2$-desaturation burden\newline{\scriptsize 240~s, MAE $\downarrow$} &
      \ablationci{0.1768}{0.1739}{0.1797} & \ablationci{0.1779}{0.1752}{0.1806} & \ablationci{\textbf{0.1711}}{0.1684}{0.1740} & \ablationci{0.1713}{0.1686}{0.1741} \\
    SDB AHI-code classification\newline{\scriptsize 240~s, AUROC $\uparrow$} &
      \ablationci{\textbf{0.6779}}{0.5250}{0.8184} & \ablationci{0.6253}{0.4694}{0.7692} & \ablationci{0.6291}{0.4717}{0.7857} & \ablationci{0.6115}{0.4615}{0.7624} \\
    DREAMT age\newline{\scriptsize 10~s, MAE $\downarrow$} &
      \ablationci{11.8329}{9.1310}{14.7515} & \ablationci{11.8329}{9.1310}{14.7515} & \ablationci{\textbf{10.9565}}{8.5759}{13.7628} & \ablationci{11.0956}{8.7766}{13.7448} \\
    DREAMT sleep staging\newline{\scriptsize 30~s, F1 $\uparrow$} &
      \ablationci{0.2687}{0.2361}{0.3071} & \ablationci{0.2574}{0.2344}{0.2797} & \ablationci{0.2441}{0.2183}{0.2679} & \ablationci{\textbf{0.2720}}{0.2347}{0.3028} \\
    DREAMT apnea/hypopnea detection\newline{\scriptsize 240~s, AUROC $\uparrow$} &
      \ablationci{0.5225}{0.4355}{0.6143} & \ablationci{0.5269}{0.4393}{0.6115} & \ablationci{0.5548}{0.4861}{0.6333} & \ablationci{\textbf{0.5700}}{0.4898}{0.6527} \\
    \bottomrule
  \end{tabularx}
\end{table}

%% file: main4/tables/rae_ppg_stage_preservation_main4.tex
% CONFIRMED RESULT: raeppg2_r1_p4_complete_20260916_v2; all 76 comparisons.
\begin{table}[!ht]
\topcaption{Changes in downstream utility across pretraining checkpoints for fixed input windows. Entries are later-minus-earlier differences with paired, unadjusted 95\% participant-cluster intervals. Utility is $-\mathrm{MAE}$, AUROC, or macro-F1, so positive differences indicate improvement.}
\label{tab:stage-preservation}
\centering
\scriptsize
\setlength{\tabcolsep}{3pt}
\renewcommand{\arraystretch}{1}
\providecommand{\stageci}[3]{\shortstack{#1\\{\scriptsize[#2, #3]}}}
\begin{tabularx}{\textwidth}{@{}Xc>{\centering\arraybackslash}p{55pt}*{2}{>{\centering\arraybackslash}p{94pt}}@{}}
\toprule
Task / input & Checkpoints & A and D (each) & B: ON & C: OFF \\
\midrule
DaLiA HR / 10~s & S$\to$M & 0 & \stageci{-0.6653}{-2.4852}{+1.5717} & \stageci{-1.4716}{-2.7188}{+0.2765} \\
DaLiA HR / 10~s & S$\to$L & 0 & \stageci{-0.8051}{-2.0708}{+1.4527} & \stageci{-1.7474}{-3.0385}{+0.4282} \\
BUT PPG HR / 10~s & S$\to$M & 0 & \stageci{+0.0412}{-0.3957}{+0.4515} & \stageci{-0.1614}{-0.4752}{+0.0820} \\
BUT PPG HR / 10~s & S$\to$L & 0 & \stageci{-0.1816}{-0.5692}{+0.1106} & \stageci{-0.1450}{-0.4798}{+0.1277} \\
DeepBeat AF / 10~s & S$\to$M & 0 & \stageci{-0.0568}{-0.0797}{-0.0207} & \stageci{-0.0356}{-0.0562}{-0.0238} \\
DeepBeat AF / 10~s & S$\to$L & 0 & \stageci{-0.0654}{-0.0903}{-0.0197} & \stageci{-0.0799}{-0.0970}{-0.0431} \\
DeepBeat quality / 10~s & S$\to$M & 0 & \stageci{+0.0135}{-0.0053}{+0.0295} & \stageci{+0.0019}{-0.0189}{+0.0247} \\
DeepBeat quality / 10~s & S$\to$L & 0 & \stageci{-0.0055}{-0.0297}{+0.0232} & \stageci{-0.0009}{-0.0133}{+0.0144} \\
\bottomrule
\end{tabularx}
\end{table}

\begin{table}[!ht]
\addtocounter{table}{-1}
\renewcommand{\theHtable}{\arabic{table}.continued}
\topcaption{Changes in downstream utility across pretraining checkpoints for fixed input windows (continued).}
\centering
\scriptsize
\setlength{\tabcolsep}{3pt}
\renewcommand{\arraystretch}{1}
\providecommand{\stageci}[3]{\shortstack{#1\\{\scriptsize[#2, #3]}}}
\begin{tabularx}{\textwidth}{@{}Xc>{\centering\arraybackslash}p{55pt}*{2}{>{\centering\arraybackslash}p{94pt}}@{}}
\toprule
Task / input & Checkpoints & A and D (each) & B: ON & C: OFF \\
\midrule
PTT SBP / 10~s & S$\to$M & 0 & \stageci{-0.0806}{-0.5095}{+0.3675} & \stageci{-0.2695}{-0.7449}{+0.1594} \\
PTT SBP / 10~s & S$\to$L & 0 & \stageci{-0.2082}{-0.8874}{+0.4442} & \stageci{-0.3870}{-1.0548}{+0.3268} \\
PTT DBP / 10~s & S$\to$M & 0 & \stageci{+0.4238}{-0.7027}{+1.3793} & \stageci{+0.2218}{-0.9236}{+1.2553} \\
PTT DBP / 10~s & S$\to$L & 0 & \stageci{+0.2420}{-0.7670}{+1.1346} & \stageci{+0.1105}{-0.9515}{+1.0945} \\
DREAMT age / 10~s & S$\to$M & 0 & \stageci{+0.7183}{-0.0397}{+1.6900} & \stageci{+0.5945}{-0.3183}{+1.6819} \\
DREAMT age / 10~s & S$\to$L & 0 & \stageci{+1.2587}{+0.3049}{+2.7789} & \stageci{+1.1196}{+0.1184}{+2.6454} \\
WESAD stress / 30~s & M$\to$L & 0 & \stageci{-0.0558}{-0.1346}{-0.0035} & \stageci{-0.0355}{-0.0850}{+0.0002} \\
WESAD arousal / 30~s & M$\to$L & 0 & \stageci{-0.1172}{-0.1749}{-0.0460} & \stageci{-0.0659}{-0.1920}{+0.0190} \\
WESAD valence / 30~s & M$\to$L & 0 & \stageci{-0.1027}{-0.1786}{+0.0037} & \stageci{-0.0707}{-0.1427}{+0.0041} \\
DaLiA activity / 30~s & M$\to$L & 0 & \stageci{+0.0133}{-0.0537}{+0.0694} & \stageci{+0.0170}{-0.0809}{+0.0883} \\
DREAMT sleep stage / 30~s & M$\to$L & 0 & \stageci{-0.0431}{-0.0999}{+0.0054} & \stageci{+0.0035}{-0.0380}{+0.0379} \\
\bottomrule
\end{tabularx}
\par\smallskip
\noindent\parbox{\textwidth}{\scriptsize S, M, and L denote checkpoints after the 10~s, 30~s, and 240~s stages. Within each of A and D, all changes are exactly zero with intervals [0, 0], reflecting identical representations and predictions across the compared checkpoints. Intervals use 1,000 shared participant-bootstrap draws with the checkpoint-specific task probes fixed.}
\end{table}

%% file: main4/tables/rae_ppg_fixed_input_preservation_main4.tex
% CONFIRMED RESULT: raeppg2_r1_representation_forgetting_20260918_v1/aggregate_results.json
% (sha256 f0cd0147f40222e9...); rendered by render_fixed_input_preservation.py.
\begin{table}[!ht]
\topcaption{Changes in earlier-feature decoding across pretraining checkpoints at fixed input duration. B and C report later-minus-earlier feature-average gain changes; the final column reports the paired difference between those stage changes. Entries include unadjusted 95\% percentile intervals.}
\label{tab:fixed-input-preservation}
\centering
\scriptsize
\setlength{\tabcolsep}{3pt}
\renewcommand{\arraystretch}{1}
\begin{tabularx}{\textwidth}{@{}>{\raggedright\arraybackslash}X>{\raggedright\arraybackslash}p{56pt}>{\centering\arraybackslash}p{48pt}*{3}{>{\centering\arraybackslash}p{72pt}}@{}}
\toprule
Fixed input / feature set & Checkpoints & A and D (each) & B: ON & C: OFF & \shortstack{Stage-change\\difference: B$-$C} \\
\midrule
10~s / nine 10~s features & S$\to$M & 0 & \shortstack{$-$0.1324\\{[$-$0.1401, $-$0.1248]}} & \shortstack{$-$0.1637\\{[$-$0.1732, $-$0.1542]}} & \shortstack{+0.0313\\{[+0.0252, +0.0372]}} \\
10~s / nine 10~s features & S$\to$L & 0 & \shortstack{$-$0.1660\\{[$-$0.1747, $-$0.1575]}} & \shortstack{$-$0.1866\\{[$-$0.1965, $-$0.1769]}} & \shortstack{+0.0205\\{[+0.0140, +0.0271]}} \\
30~s / two 30~s features & M$\to$L & 0 & \shortstack{$-$0.3128\\{[$-$0.3327, $-$0.2931]}} & \shortstack{$-$0.3807\\{[$-$0.4036, $-$0.3583]}} & \shortstack{+0.0679\\{[+0.0519, +0.0840]}} \\
10~s / nine 10~s features & M$\to$L (secondary) & 0 & \shortstack{$-$0.0336\\{[$-$0.0392, $-$0.0282]}} & \shortstack{$-$0.0228\\{[$-$0.0297, $-$0.0161]}} & \shortstack{-0.0107\\{[-0.0197, -0.0019]}} \\
\bottomrule
\end{tabularx}
\par\smallskip
\noindent\parbox{\textwidth}{\scriptsize The A and D entries are zero within each condition, with intervals [0, 0]. B$-$C is $(G_{\mathrm{later},B}-G_{\mathrm{earlier},B})-(G_{\mathrm{later},C}-G_{\mathrm{earlier},C})$; positive values indicate a more favorable stage change with retained-target supervision. Intervals use the shared draws described in Appendix~\ref{app:ablation-transfer}. Readouts are interval-aligned, unlike the mean-pooled native-feature readouts in Table~\ref{tab:ablation-design-main}.}
\end{table}

%% file: main4/tables/rae_ppg_ablation_features_main4_format.tex
% CONFIRMED RESULT; source: raeppg2_r1_p4_complete_20260916_v2/result.json.
% Consolidated20absolute plus15paired cells, unchanged stored P4 values.
\begin{table}[!htb]
  \topcaption{Feature decodability after pretraining. Panel A reports feature-average gain [95\% CI] for each condition; Panel B reports paired gain differences for the same feature sets and readouts. Feature introduction / input gives the targets\textquotesingle{} introduction duration and the encoder input duration.}
  \label{tab:ablation-feature-accessibility}
\centering
  \scriptsize
  \setlength{\tabcolsep}{2pt}
  \renewcommand{\arraystretch}{1}
  {\raggedright \noindent\textbf{A. Feature decoding gains}\par}
  \begin{tabularx}{\columnwidth}{@{}Xc*{4}{>{\centering\arraybackslash}p{0.19\columnwidth}}@{}}
    \toprule
    & & \multicolumn{2}{c}{Nested/protected} & \multicolumn{2}{c}{Fully shared} \\
    \cmidrule(lr){3-4}\cmidrule(l){5-6}
    Feature introduction / input & $n$ & A: ON & D: OFF & B: ON & C: OFF \\
    \midrule
    \multicolumn{6}{@{}l}{\textit{Earlier features in longer inputs: aligned readout}} \\
    10~s / 30~s & 9 & \shortstack{0.8303\\{[0.8149, 0.8458]}} & \shortstack{0.7097\\{[0.6944, 0.7248]}} & \shortstack{0.6634\\{[0.6487, 0.6784]}} & \shortstack{0.6348\\{[0.6206, 0.6485]}} \\
    10~s / 240~s & 9 & \shortstack{0.6441\\{[0.6284, 0.6591]}} & \shortstack{0.5866\\{[0.5710, 0.6017]}} & \shortstack{0.6647\\{[0.6497, 0.6790]}} & \shortstack{0.6104\\{[0.5973, 0.6234]}} \\
    30~s / 240~s & 2 & \shortstack{0.6661\\{[0.6443, 0.6871]}} & \shortstack{0.5341\\{[0.5104, 0.5574]}} & \shortstack{0.7470\\{[0.7304, 0.7624]}} & \shortstack{0.5297\\{[0.5069, 0.5517]}} \\
    \midrule
    \multicolumn{6}{@{}l}{\textit{New features at introduction duration: token mean}} \\
    30~s / 30~s & 2 & \shortstack{0.8984\\{[0.8901, 0.9060]}} & \shortstack{0.8924\\{[0.8841, 0.9001]}} & \shortstack{0.6057\\{[0.5855, 0.6252]}} & \shortstack{0.5334\\{[0.5095, 0.5561]}} \\
    240~s / 240~s & 5 & \shortstack{0.7553\\{[0.7392, 0.7712]}} & \shortstack{0.7488\\{[0.7333, 0.7640]}} & \shortstack{0.8015\\{[0.7859, 0.8163]}} & \shortstack{0.7872\\{[0.7711, 0.8023]}} \\
    \bottomrule
  \end{tabularx}
\end{table}

%% file: main4/tables/rae_ppg_ablation_paired_main4.tex
\begin{table}[!htb]
  \addtocounter{table}{-1}
  \renewcommand{\theHtable}{\arabic{table}.continued}
  \topcaption{Feature decodability after pretraining (continued).}
  \centering\scriptsize
% CONFIRMED RESULT: original15 P4 paired differences, now Panel B of the common table.
{\raggedright \noindent\textbf{B. Paired differences in decoding gain}\par}
\setlength{\parskip}{0pt}
\scriptsize
\setlength{\tabcolsep}{3pt}
\renewcommand{\arraystretch}{1}
\begin{tabularx}{\textwidth}{@{}Xc*{3}{>{\centering\arraybackslash}p{95pt}}@{}}
\toprule
Feature introduction / input & $n$ & A$-$D & B$-$C & A$-$B \\
\midrule
    \multicolumn{5}{@{}l}{\textit{Earlier features in longer inputs: aligned readout}} \\
    10~s / 30~s & 9 & \shortstack{+0.1205\\{[+0.1130, +0.1285]}} & \shortstack{+0.0287\\{[+0.0224, +0.0351]}} & \shortstack{+0.1668\\{[+0.1579, +0.1760]}} \\
    10~s / 240~s & 9 & \shortstack{+0.0575\\{[+0.0504, +0.0645]}} & \shortstack{+0.0544\\{[+0.0470, +0.0620]}} & \shortstack{-0.0207\\{[-0.0277, -0.0137]}} \\
    30~s / 240~s & 2 & \shortstack{+0.1320\\{[+0.1180, +0.1459]}} & \shortstack{+0.2173\\{[+0.2004, +0.2352]}} & \shortstack{-0.0809\\{[-0.0960, -0.0669]}} \\
    \midrule
    \multicolumn{5}{@{}l}{\textit{New features at introduction duration: token mean}} \\
    30~s / 30~s & 2 & \shortstack{+0.0060\\{[+0.0011, +0.0110]}} & \shortstack{+0.0723\\{[+0.0574, +0.0879]}} & \shortstack{+0.2927\\{[+0.2730, +0.3127]}} \\
    240~s / 240~s & 5 & \shortstack{+0.0065\\{[+0.0017, +0.0113]}} & \shortstack{+0.0144\\{[+0.0079, +0.0207]}} & \shortstack{-0.0462\\{[-0.0527, -0.0398]}} \\
\bottomrule
\end{tabularx}
  \par\smallskip
  \noindent\parbox{\textwidth}{\scriptsize $n$ counts targets. A$-$D and B$-$C compare supervision ON versus OFF within an organization; A$-$B compares organizations with supervision ON. Positive differences favor the first condition. Paired contrasts are computed before rounding using the shared draws in Appendix~\ref{app:fourarm-feature-accessibility}. Multiplying $\Delta$gain by 100 gives percentage points of baseline error removed.}
\end{table}

%% file: main4/tables/rae_ppg_shorter_group_reuse_main4.tex
% CONFIRMED RESULT: raeppg2_full_prior_fresh_val_20260916_v1/{M,L}/result.json.
% Combined-dataset feature-specific and native_macro rows; original observations and CIs unchanged.
% 2026-09-19 (user instruction): former main-text Table 6 (rae_ppg_reuse_features_main4.tex) merged here;
% aggregate gain intervals of the former summary table are kept.
\begin{table}[!htb]
\topcaption{Decoding newly introduced features after matched training continuations. Error reduction is relative to FULL-PRIOR INITIAL, as defined in Appendix~\ref{app:shorter-group-reuse}. Brackets give unadjusted 95\% participant/case-cluster bootstrap intervals; error-reduction intervals are paired.}
\label{tab:shorter-group-reuse}
\label{tab:shorter-group-reuse-summary}
\centering
\scriptsize
\setlength{\tabcolsep}{3pt}
\renewcommand{\arraystretch}{1}
\begin{tabularx}{\textwidth}{@{}lX*{3}{>{\centering\arraybackslash}p{0.19\textwidth}}@{}}
\toprule
Input & Newly introduced feature & FULL-PRIOR INITIAL gain & LEARNED gain & Error reduction (\%) \\
\midrule
30~s & Normalized reflected delay & 0.7346 & 0.8449 & 41.57 [34.42, 47.88] \\
 & PRV-RMSSD & 0.8396 & 0.8771 & 23.41 [16.56, 29.74] \\
 & \textbf{Aggregate: two features} & \textbf{0.7871} [0.7670, 0.8061] & \textbf{0.8610} [0.8472, 0.8731] & \textbf{34.72} [29.23, 39.77] \\
\midrule
240~s & PRV-pNN50 & 0.6294 & 0.7712 & 38.25 [32.35, 43.85] \\
 & Uncentered ACF index & 0.5973 & 0.7654 & 41.74 [35.43, 48.01] \\
 & PPGV LF fraction & 0.3998 & 0.5472 & 24.55 [20.22, 28.63] \\
 & Pulse-interval SD & 0.6682 & 0.7815 & 34.14 [27.53, 41.01] \\
 & Mean pulse rate & 0.8585 & 0.9389 & 56.80 [49.49, 62.87] \\
 & \textbf{Aggregate: five features} & \textbf{0.6306} [0.6128, 0.6475] & \textbf{0.7608} [0.7455, 0.7754] & \textbf{35.24} [32.39, 38.09] \\
\bottomrule
\end{tabularx}
\par\smallskip
\noindent\parbox{\textwidth}{\scriptsize Aggregate gains are equal-weight means over the two new features at 30~s or the five at 240~s. Aggregate error reductions are calculated from the aggregate normalized errors, not by averaging the feature-level percentages.}
\end{table}

%% file: main4/sections/appendix_matched_retrain_main4.tex
\FloatBarrier
\section{Comparison-model retraining}
\label{app:matched-retrain}

To assess sensitivity to pretraining-data differences, we retrained the PPG encoders of PaPaGei-P/S, Pulse-PPG, SIGMA-PPG, and AnyPPG from scratch using eligible source recordings in the RAE-PPG TRAIN partition.
Table~\ref{tab:matched-retrain-setup} specifies the model-specific data subsets, training procedures, and optimization budgets, adapted from the published methods and released implementations~\citep{pillai2025papagei,saha2025pulseppg,guo2026sigmappg,nie2025anyppg}.
AnyPPG retained a frozen, pretrained ECG teacher. The RAE-PPG checkpoint was unchanged.

\noindent\runinheading{Retraining data.}
Source signals are read at their native sampling rates at the offsets of RAE-PPG TRAIN windows with the corresponding input duration.
Each model then applies its own admission and preprocessing rules.
No RAE-PPG pretraining VAL or TEST recordings, or MESA recordings, are used in these retraining runs.
No downstream score was used to tune the retraining settings.

\noindent\runinheading{Evaluation.}
The retrained encoders are frozen and evaluated using the model-specific interfaces in Appendix~\ref{app:foundation-interfaces} and the participants, examples, and shared bootstrap draws in Appendix~\ref{app:linear-evaluation}.
Probes are selected and fitted separately for each encoder on TRAIN using that appendix's procedure.
Table~\ref{tab:matched-retrain-transfer} uses the first run for each PaPaGei model and the checkpoint specified in Table~\ref{tab:matched-retrain-setup} for each other model; each checkpoint is used across all 18 tasks.
Re-extracting representations from the released checkpoints through this pipeline reproduced the stored features within 0.15\% of the largest feature magnitude in each task.

\noindent\runinheading{Results.}
With all five comparison models replaced by their retrained versions, RAE-PPG achieves the best point score on 12 of 18 tasks and the lowest mean rank, 1.556 (Table~\ref{tab:matched-retrain-transfer}A).
Table~\ref{tab:matched-retrain-transfer}B summarizes comparisons of RAE-PPG with each retrained model and of each retrained model with its released checkpoint.

\input{main4/tables/rae_ppg_matched_retrain_setup_main4}

\input{main4/tables/rae_ppg_matched_retrain_transfer_main4}

\noindent\runinheading{Additional runs.}
Across the five individual runs of each PaPaGei model, RAE-PPG has the better point score on 13--14 tasks against PaPaGei-P and on 17--18 against PaPaGei-S.
Two additional five-epoch AnyPPG runs use the released warm-up length and the paper's warm-up fraction, respectively.
The first outperforms the released AnyPPG checkpoint on five tasks, with one numerical tie; the second outperforms it on two tasks.
Neither run has a paired 95\% interval entirely favoring retraining.

%% file: main4/tables/rae_ppg_matched_retrain_setup_main4.tex
% CONFIRMED RESULT: original training/extraction receipts and saved AnyPPG checkpoint audited in Revision87.
\begin{table}[H]
\topcaption{Model-specific retraining procedures. Data are drawn from the RAE-PPG pretraining TRAIN partition; counts refer to the admitted pools before the recipe-specific validation split where one is used.}
\label{tab:matched-retrain-setup}
\centering
\setlength{\parskip}{0pt}
\fontsize{8}{9}\selectfont
\setlength{\tabcolsep}{3pt}
\renewcommand{\arraystretch}{1}
\begin{tabularx}{\textwidth}{@{}>{\raggedright\arraybackslash}p{43pt}>{\raggedright\arraybackslash}X>{\raggedright\arraybackslash}p{87pt}>{\raggedright\arraybackslash}p{110pt}@{}}
\toprule
Model & Preprocessing and objective & Admitted data & Optimization and checkpoint \\
\midrule
PaPaGei-P/S & 10~s, 125~Hz; 0.5--12~Hz Chebyshev-II filtering and 50~ms smoothing; segment $z$-score; flat-line screening of the $z$-scored segment and morphology-range admission. S: sVRI-bin contrastive loss with IPA/SQI regression. P: same-participant positives. & 11{,}350{,}415 windows; 10{,}126 participants/cases. & Adam, $10^{-4}$; global batch 1{,}024; 15{,}000 updates. Five runs per model, differing only in random-number streams. \\
Pulse-PPG & 240~s, 50~Hz; no filtering; per-participant $z$-score. Learned motif-based distance (MotifDist), then relative contrastive training (RelCon). & 1{,}337{,}278 windows; 2\% of participants/cases held out for validation-loss checkpoint selection. & MotifDist: 20 epochs, batch 16, $10^{-3}$. RelCon: 6 epochs, batch 64, $10^{-4}$.$^{a}$ Lowest-validation-loss checkpoint at each stage. \\
SIGMA-PPG & 240~s, 50~Hz; 0.5--8~Hz second-order Butterworth filtering; min--max scaling; quality scores over 2~s patches. Vector-quantized tokenizer, then teacher-guided masked modeling. & 1{,}337{,}261 windows; released 80/20 file split within eligible TRAIN data. & Tokenizer: 100 epochs. Masked modeling: 80 epochs. Global batch 4{,}096; masked-modeling gradient accumulation 2. \\
AnyPPG & 10~s PPG--ECG pairs. PPG: 0.5--8~Hz band-pass, 125~Hz. Lead-II ECG: 0.5~Hz high-pass, 50~Hz notch, 500~Hz. Flat-segment and ECG-quality screening; segment $z$-score. Contrastive alignment with a frozen, pretrained single-lead ECGFounder teacher~\citep{li2025ecgfounder}. & 8{,}205{,}102 pairs, MIMIC-III only; 2\% of participants/cases held out for validation-loss checkpoint selection. & AdamW, $5\times10^{-4}$; 2{,}560 pairs per process, two processes, accumulation 1. Run: 24 epochs, 37{,}848 optimizer updates. Lowest-validation-loss saved checkpoint: 33{,}116 updates.$^{b}$ \\
\bottomrule
\end{tabularx}
\par\smallskip
\noindent\parbox{\textwidth}{\scriptsize
$^{a}$RelCon batch size 64 and 6 epochs follow the paper; the released configuration lists 16 and 20.
The released loader's amplitude-clipping step is omitted because its threshold is unspecified.
$^{b}$The scheduler counter advances twice per optimizer update; the counts above are optimizer updates, not scheduler positions.
Selection minimizes validation loss over all 24 saved checkpoints using fixed batches.
The VitalDB retraining cache contains PPG only; Appendix~\ref{app:targets} obtains ECG separately for selected VitalDB cases.}
\end{table}

%% file: main4/tables/rae_ppg_matched_retrain_transfer_main4.tex
% CONFIRMED RESULT: AnyPPG uses the validation-selected 33116-update checkpoint; other model scores unchanged.
\begin{table}[H]
\topcaption{Downstream performance after comparison-model retraining. Panel A compares the unchanged RAE-PPG checkpoint with the five retrained checkpoints; bold and underline mark the best and second-best point scores. Panel B summarizes paired comparisons. Released-checkpoint scores are reported in Table~\ref{tab:transfer-all}.}
\label{tab:matched-retrain-transfer}
\centering
\setlength{\parskip}{0pt}
\fontsize{8}{9}\selectfont
\setlength{\tabcolsep}{3pt}
\renewcommand{\arraystretch}{1}
\begin{tabularx}{\textwidth}{@{}>{\raggedright\arraybackslash}X*{6}{>{\centering\arraybackslash}p{43pt}}@{}}
\toprule
\multicolumn{7}{@{}l}{\textbf{A. RAE-PPG and the five retrained models}} \\
Downstream task & RAE-PPG & PaPaGei-P & PaPaGei-S & Pulse-PPG & SIGMA-PPG & AnyPPG \\
\midrule
\multicolumn{7}{@{}l}{\textit{Regression: MAE $\downarrow$}} \\
DaLiA: HR\textsuperscript{10} & \bestscore{8.6522} & \secondbest{8.7955}$^{\blacktriangle}$ & 17.6340$^{\blacktriangledown}$ & 14.4006$^{\blacktriangledown}$ & 14.2939 & 9.0165$^{\blacktriangledown}$ \\
BUT PPG: HR\textsuperscript{10} & 10.5144 & \secondbest{10.1444}$^{\blacktriangle}$ & 10.8978 & 10.9398$^{\blacktriangledown}$ & 10.6272 & \bestscore{10.1215}$^{\blacktriangledown}$ \\
PTT-PPG: SBP\textsuperscript{10} & \bestscore{9.5291} & \secondbest{9.6366} & 10.2725 & 10.3561$^{\blacktriangledown}$ & 10.4055 & 10.2046 \\
PTT-PPG: DBP\textsuperscript{10} & \bestscore{5.4489} & \secondbest{5.5092} & 5.8239 & 6.0458$^{\blacktriangledown}$ & 6.0637 & 5.9476 \\
DREAMT: Age\textsuperscript{10} & 11.8329 & \bestscore{10.4513} & 13.3999$^{\blacktriangledown}$ & 12.0621$^{\blacktriangledown}$ & 11.9785 & \secondbest{11.5162} \\
MESA: A/H\textsuperscript{240} & \bestscore{0.0834} & 0.0907$^{\blacktriangledown}$ & 0.0887$^{\blacktriangle}$ & 0.0895 & \secondbest{0.0880}$^{\blacktriangle}$ & 0.0905$^{\blacktriangledown}$ \\
MESA: Arousals\textsuperscript{240} & \bestscore{0.0432} & 0.0476$^{\blacktriangledown}$ & 0.0471$^{\blacktriangle}$ & 0.0483$^{\blacktriangledown}$ & \secondbest{0.0455}$^{\blacktriangle}$ & 0.0480$^{\blacktriangledown}$ \\
MESA: SpO$_2$\textsuperscript{240} & \bestscore{0.1768} & 0.2086$^{\blacktriangledown}$ & 0.2033$^{\blacktriangle}$ & \secondbest{0.2007} & 0.2082 & 0.2151$^{\blacktriangledown}$ \\
\bottomrule
\end{tabularx}
\end{table}

\newpage % Continue at a complete panel/row boundary.
\begin{table}[H]
\addtocounter{table}{-1}
\renewcommand{\theHtable}{\arabic{table}.taskscontinued}
\topcaption{Downstream performance after comparison-model retraining (continued).}
\centering
\setlength{\parskip}{0pt}
\fontsize{8}{9}\selectfont
\setlength{\tabcolsep}{3pt}
\renewcommand{\arraystretch}{1}
\begin{tabularx}{\textwidth}{@{}>{\raggedright\arraybackslash}X*{6}{>{\centering\arraybackslash}p{43pt}}@{}}
\toprule
\multicolumn{7}{@{}l}{\textbf{A. RAE-PPG and the five retrained models}} \\
Downstream task & RAE-PPG & PaPaGei-P & PaPaGei-S & Pulse-PPG & SIGMA-PPG & AnyPPG \\
\midrule
\multicolumn{7}{@{}l}{\textit{Classification: AUROC $\uparrow$ ($\dagger$ macro-F1 $\uparrow$)}} \\
DeepBeat: AF\textsuperscript{10} & \secondbest{0.7774} & 0.7750$^{\blacktriangle}$ & 0.6541 & 0.6397 & 0.7275 & \bestscore{0.8197}$^{\blacktriangledown}$ \\
DeepBeat: Quality\textsuperscript{10\ensuremath{\dagger}} & \bestscore{0.6718} & \secondbest{0.6639}$^{\blacktriangle}$ & 0.5986 & 0.6360$^{\blacktriangledown}$ & 0.6139$^{\blacktriangledown}$ & 0.6366 \\
WESAD: Stress\textsuperscript{30} & \bestscore{0.9514} & \secondbest{0.9499}$^{\blacktriangle}$ & 0.8288 & 0.7983 & 0.8778 & 0.9035$^{\blacktriangledown}$ \\
WESAD: Arousal\textsuperscript{30} & \bestscore{0.8571} & \secondbest{0.7875} & 0.6978 & 0.6886 & 0.7289 & 0.7637 \\
WESAD: Valence\textsuperscript{30} & \secondbest{0.6973} & \bestscore{0.7120}$^{\blacktriangle}$ & 0.6493 & 0.5440 & 0.6573 & 0.6640 \\
DaLiA: Activity\textsuperscript{30\ensuremath{\dagger}} & \bestscore{0.4078} & \secondbest{0.3807}$^{\blacktriangle}$ & 0.2265$^{\blacktriangle}$ & 0.1689$^{\blacktriangledown}$ & 0.2831 & 0.3119 \\
DREAMT: Sleep\textsuperscript{30\ensuremath{\dagger}} & \secondbest{0.2687} & 0.2508 & 0.2131 & 0.2400 & \bestscore{0.2776} & 0.2612 \\
MESA: Sleep\textsuperscript{240\ensuremath{\dagger}} & \bestscore{0.4735} & 0.4428$^{\blacktriangle}$ & 0.4374 & 0.4243 & \secondbest{0.4436} & 0.4043$^{\blacktriangledown}$ \\
SDB: AHI code\textsuperscript{240} & \bestscore{0.6779} & 0.5213 & 0.4649 & 0.4424 & \secondbest{0.5395} & 0.5263 \\
DREAMT: A/H\textsuperscript{240} & 0.5225 & 0.5237 & 0.4916 & \secondbest{0.5358} & \bestscore{0.5566} & 0.4975 \\
\bottomrule
\end{tabularx}
\end{table}

\begin{table}[H]
\addtocounter{table}{-1}
\renewcommand{\theHtable}{\arabic{table}.continued}
\topcaption{Downstream performance after comparison-model retraining (continued).}
\centering
\setlength{\parskip}{0pt}
\fontsize{8}{9}\selectfont
\setlength{\tabcolsep}{3pt}
\renewcommand{\arraystretch}{1}
\begin{tabularx}{\textwidth}{@{}>{\raggedright\arraybackslash}X*{6}{>{\centering\arraybackslash}p{43pt}}@{}}
\toprule
\multicolumn{7}{@{}l}{\textbf{A. Summary across the 18 tasks}} \\
Summary & RAE-PPG & PaPaGei-P & PaPaGei-S & Pulse-PPG & SIGMA-PPG & AnyPPG \\
\midrule
Best-score tasks /18 & \bestscore{12} & \secondbest{2} & 0 & 0 & \secondbest{2} & \secondbest{2} \\
Mean rank $\downarrow$ & \bestscore{1.556} & \secondbest{2.778} & 4.722 & 5.000 & 3.500 & 3.444 \\
\bottomrule
\end{tabularx}
\par\smallskip
\begin{tabularx}{\textwidth}{@{}>{\raggedright\arraybackslash}X*{4}{>{\centering\arraybackslash}p{72pt}}@{}}
\toprule
\multicolumn{5}{@{}l}{\textbf{B. Paired-comparison summaries}} \\
& \multicolumn{2}{c}{RAE-PPG vs. retrained} & \multicolumn{2}{c}{Retrained vs. released} \\
\cmidrule(lr){2-3}\cmidrule(lr){4-5}
Retrained model & RAE-PPG better (tasks /18) & CI favors RAE-PPG / retrained & Retrained better (tasks /18) & CI favors retrained / released \\
\midrule
PaPaGei-P & 14 & 5 / 1 & 14 & 8 / 3 \\
PaPaGei-S & 18 & 14 / 0 & 10 & 4 / 2 \\
Pulse-PPG & 17 & 13 / 0 & 3 & 0 / 8 \\
SIGMA-PPG & 16 & 11 / 0 & 8 & 2 / 1 \\
AnyPPG & 15 & 7 / 1 & 2 & 0 / 8 \\
\bottomrule
\end{tabularx}
\par\smallskip
\noindent\parbox{\textwidth}{\scriptsize Numeric superscripts on task names give input duration in seconds; $\dagger$ denotes macro-F1.
Ranks use unrounded scores from the six models in Panel A only.
$^{\blacktriangle}$/$^{\blacktriangledown}$ indicates that the paired 95\% interval for a retrained model versus its own released checkpoint lies entirely on the better/worse side of zero; no symbol means the interval includes zero.
Panel B counts tasks out of 18.
Intervals use 1,000 shared participant/case-bootstrap draws with fitted encoders and probes fixed, without multiplicity adjustment.
Paired intervals favor a retrained model over RAE-PPG for BUT PPG: HR (AnyPPG); DREAMT: Age (PaPaGei-P).}
\end{table}